\documentclass{article} % For LaTeX2e
\usepackage{amsthm}

\usepackage[preprint]{icml2026}
\usepackage{amsmath,amsfonts,bm}

\def\eqref#1{equation~\ref{#1}}
\def\1{\bm{1}}

\DeclareMathAlphabet{\mathsfit}{\encodingdefault}{\sfdefault}{m}{sl}
\SetMathAlphabet{\mathsfit}{bold}{\encodingdefault}{\sfdefault}{bx}{n}

\usepackage{hyperref}
\usepackage{url}
\usepackage[utf8]{inputenc} % allow utf-8 input
\usepackage[T1]{fontenc}    % use 8-bit T1 fonts
\usepackage{amsmath}
\usepackage{hyperref}       % hyperlinks
\usepackage{url}            % simple URL typesetting
\usepackage{booktabs}       % professional-quality tables
\usepackage{amsfonts}       % blackboard math symbols
\usepackage{nicefrac}       % compact symbols for 1/2, etc.
\usepackage{microtype}      % microtypography
\usepackage{multirow}       % table
\usepackage{xcolor}         % colors
\usepackage{graphicx}       % graph size
\usepackage{subfigure}
\usepackage{subcaption}

\usepackage{amssymb}
\usepackage{algorithm}
\usepackage{algorithmic}
\usepackage{wrapfig}
\usepackage{dirtytalk}

\newcommand{\draft}[1]{}

\begin{document}

\twocolumn[
\icmltitle{SD-MoE: Spectral Decomposition for Effective Expert Specialization}

% \begin{icmlauthorlist}
% \icmlauthor{Firstname1 Lastname1}{equal,yyy}
% \icmlauthor{Firstname2 Lastname2}{equal,yyy,comp}
% \end{icmlauthorlist}



\icmlsetsymbol{equal}{*}

\begin{icmlauthorlist}
\icmlauthor{Ruijun Huang}{fudan,thu2}
\icmlauthor{Fang Dong}{equal,fudan}
\icmlauthor{Xin Zhang}{equal,fudan}
\icmlauthor{Hengjie Cao}{fudan}
\icmlauthor{Zhendong Huang}{fudan}
\icmlauthor{Anrui Chen}{fudan}
\icmlauthor{Jixian Zhou}{fudan}
\icmlauthor{Mengyi Chen}{fudan}
\icmlauthor{Yifeng Yang}{fudan}
\icmlauthor{Mingzhi Dong}{bath}
\icmlauthor{Yujiang Wang}{ox}
\icmlauthor{Jinlong Hou}{sii}
\icmlauthor{Qin Lv}{boulder}
\icmlauthor{Robert P. Dick}{michigan}
\icmlauthor{Yuan Cheng}{sii}
\icmlauthor{Fan Yang}{fudanmi}
\icmlauthor{Tun Lu}{fudan}
\icmlauthor{Chun Zhang}{thu}
\icmlauthor{Li Shang}{fudan}
\end{icmlauthorlist}
  
\icmlaffiliation{fudan}{College of Computer Science and Artificial Intelligence, Fudan University, Shanghai, China}
\icmlaffiliation{bath}{University of Bath, Bath, United Kingdom}
\icmlaffiliation{ox}{Oxford Suzhou Centre for Advanced Research, Suzhou, China}
\icmlaffiliation{michigan}{Department of Electrical Engineering and Computer Science, University of Michigan}
\icmlaffiliation{sii}{Shanghai Innovation Institute, Shanghai, China}
\icmlaffiliation{boulder}{Department of Computer Science, University of Colorado Boulder, Colorado, USA}
\icmlaffiliation{thu}{Research Institute of Tsinghua University in Shenzhen, Shenzhen, China}
\icmlaffiliation{thu2}{Greater Bay Area National Center of Technology Innovation, Research Institute of Tsinghua University in Shenzhen, Shenzhen, China}
\icmlaffiliation{fudanmi}{School of Microelectronics, Fudan University, Shanghai, China}
% \icmlaffiliation{ul}{Bayes Business School, City St George's, University of London, London, UK}
% \icmlaffiliation{ul}{Bayes Business School, City St George's, University of London, London, UK}
% \icmlaffiliation{ul}{Bayes Business School, City St George's, University of London, London, UK}

\icmlcorrespondingauthor{Li Shang}{lishang@fudan.edu.cn}
%%% TODO：houjinglong //chengyuan

\icmlkeywords{Machine Learning, ICML}

\vskip 0.3in

]

\printAffiliationsAndNotice{\icmlEqualContribution} % otherwise use the standard text.

\begin{abstract}
Mixture-of-Experts (MoE) architectures scale Large Language Models via expert specialization induced by conditional computation. In practice, however, expert specialization often fails: some experts become functionally similar, while others functioning as de facto shared experts, limiting the effective capacity and model performance.
In this work, we analysis from a spectral perspective on parameter and gradient spaces, uncover that
(1) experts share highly overlapping dominant spectral components in their parameters,
(2) dominant gradient subspaces are strongly aligned across experts, driven by ubiquitous low-rank structure in human corpus, and
(3) gating mechanisms preferentially route inputs along these dominant directions, further limiting specialization.
To address this, we propose Spectral-Decoupled MoE (SD-MoE), which decomposes both parameter and gradient in the spectral space.
SD-MoE improves performance across downstream tasks, enables effective expert specialization,  incurring minimal additional computation, and can be seamlessly integrated into a wide range of existing MoE architectures, including Qwen and DeepSeek.
\end{abstract}

\section{Introduction}
\label{sec:introduction}

Mixture-of-Experts (MoE) scales Large Language Models (LLM) via expert specialization induced by conditional computation.
In practice, however, specialization in LLM MoEs often degrades: some experts are nearly interchangeable with minimal performance impact~\cite{zheng2025gateproparameterfreeexpertselection, zhang2025mixtureexpertslargelanguage}, while others are activated almost universally, effectively acting as shared experts~\cite{cai2025longtailed, guo2025advancing}. Despite its prevalence, the underlying causes of this failure of specialization remain poorly understood.

\begin{figure*}[h]
    \centering
    \subfigure[]{
        \includegraphics[width=.22\linewidth]{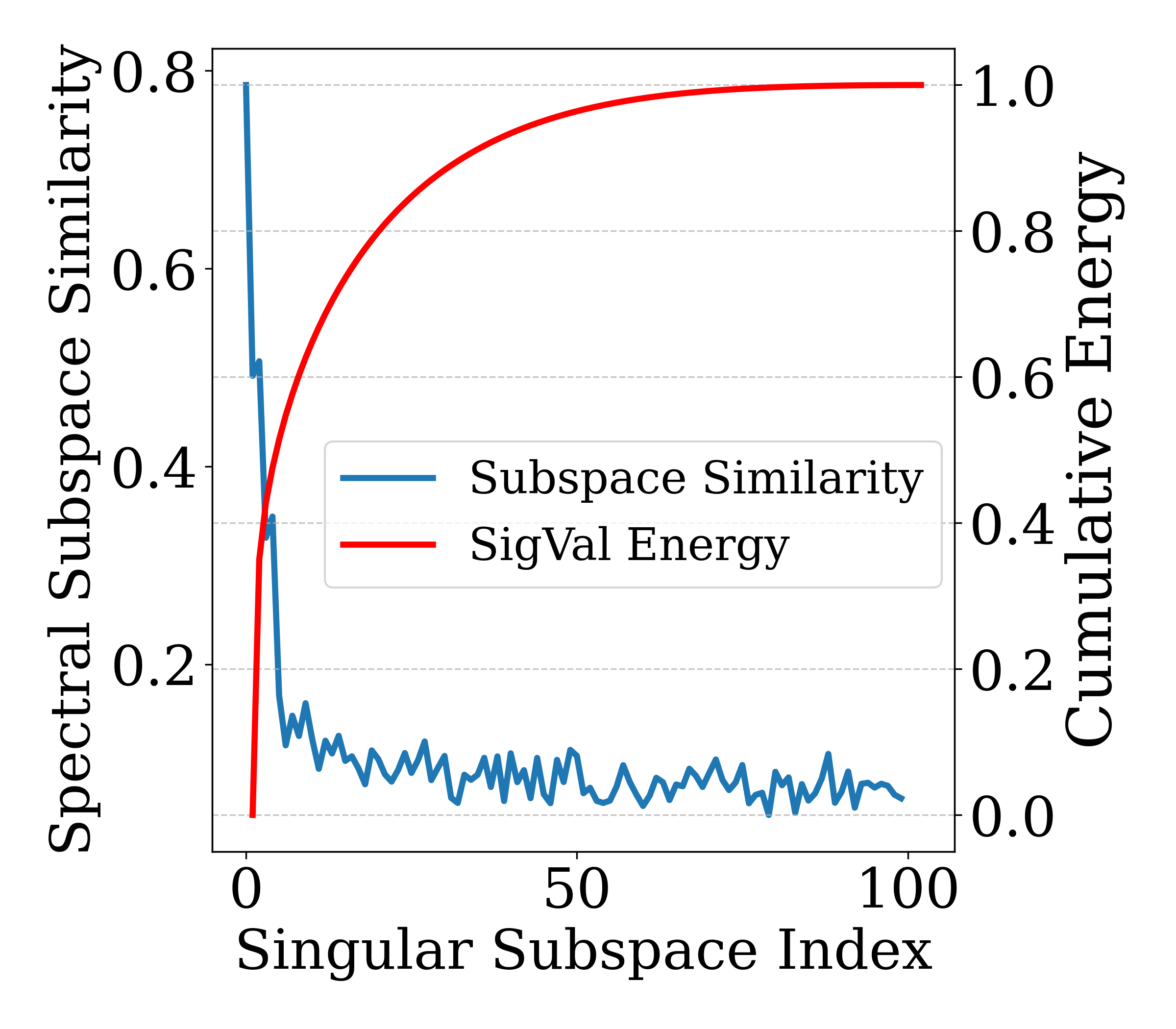}
        \label{fig:intro_expert_sig_vals}
    }
    \subfigure[]{
        \includegraphics[width=.22\linewidth]{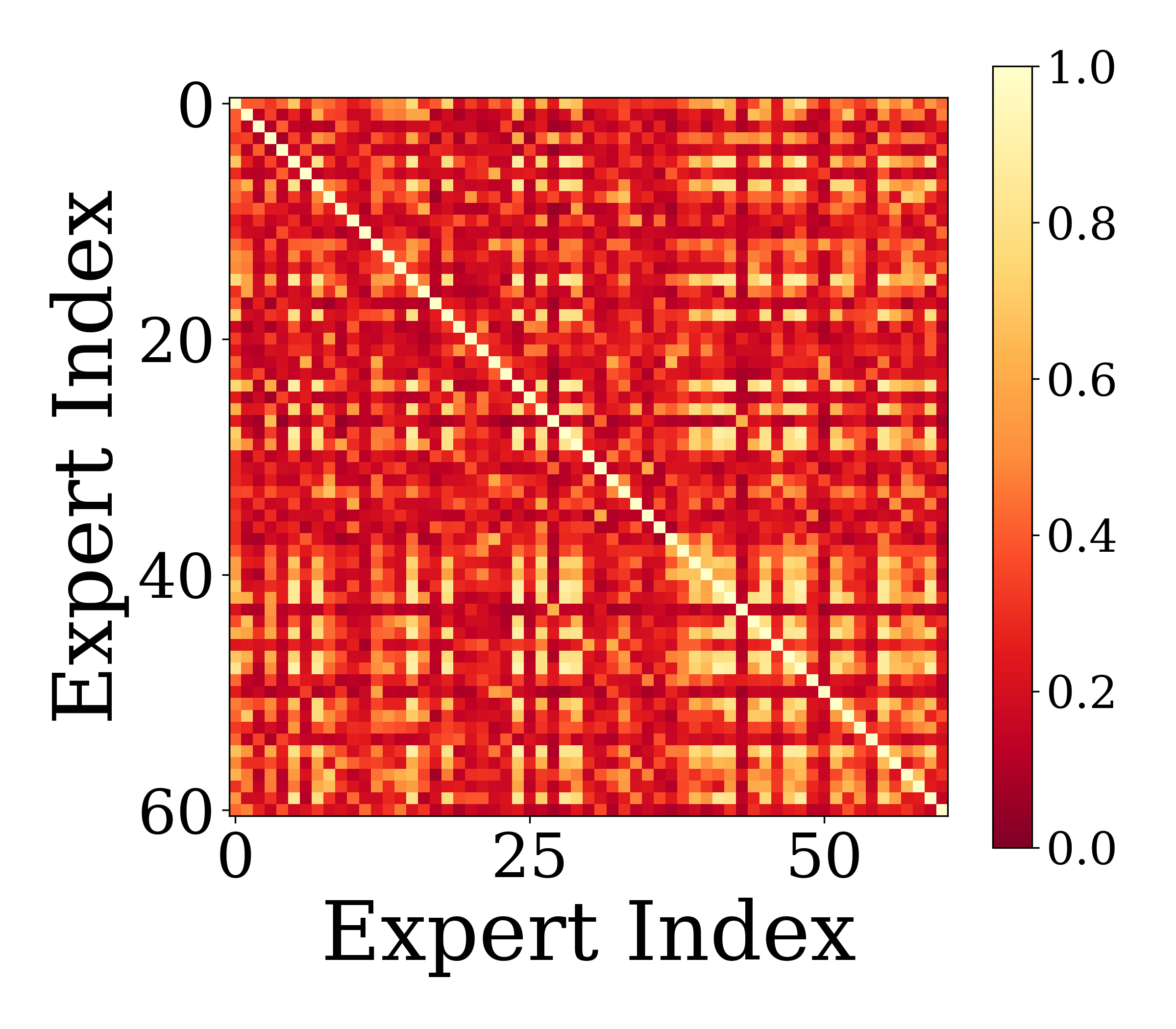}
        \label{fig:intro_param_spec_sim_qwen}
    }
    \subfigure[]{
        \includegraphics[width=.22\linewidth]{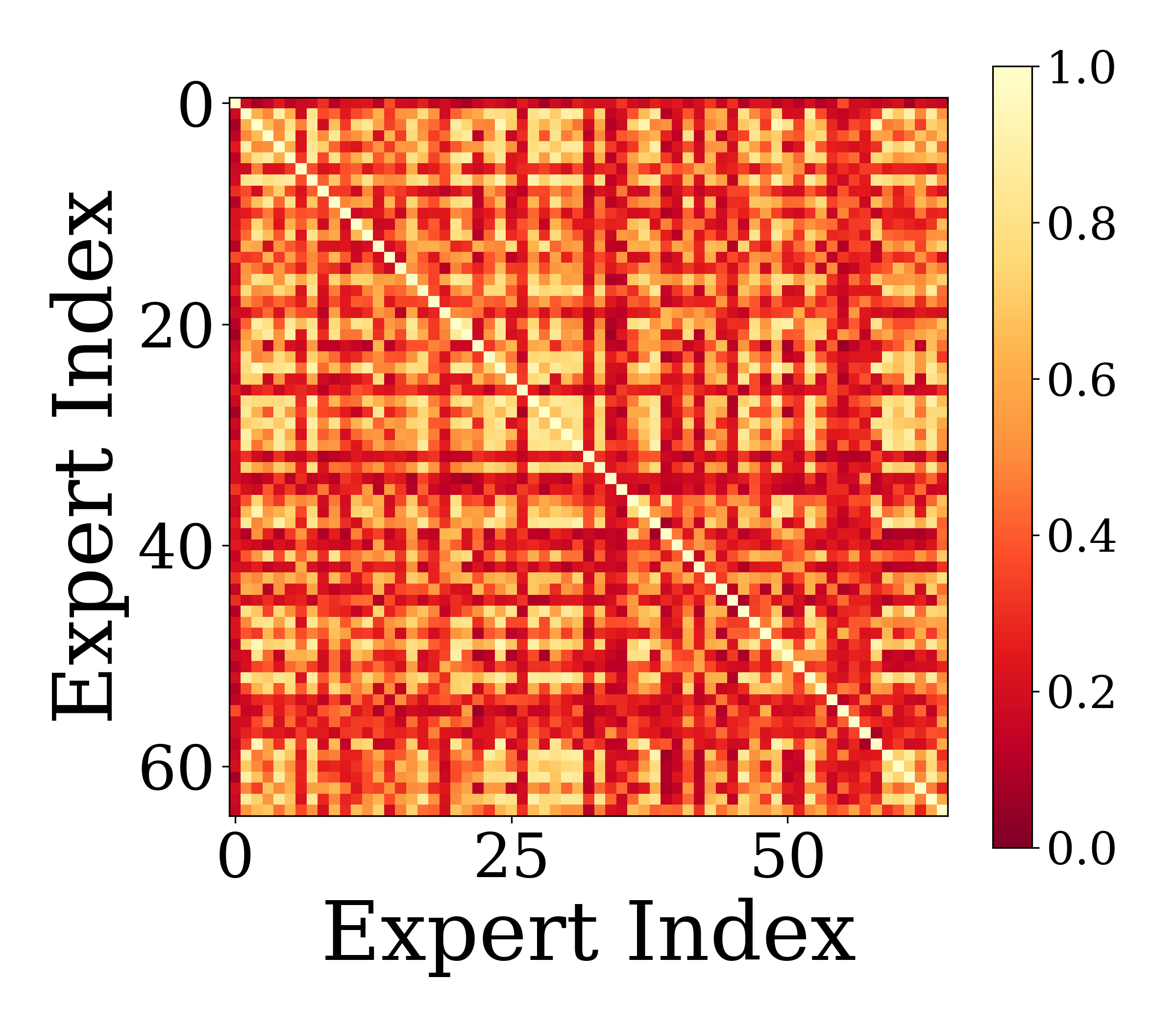}
        \label{fig:intro_param_spec_sim_ds}
    }
    \subfigure[]{
        \includegraphics[width=.22\linewidth]{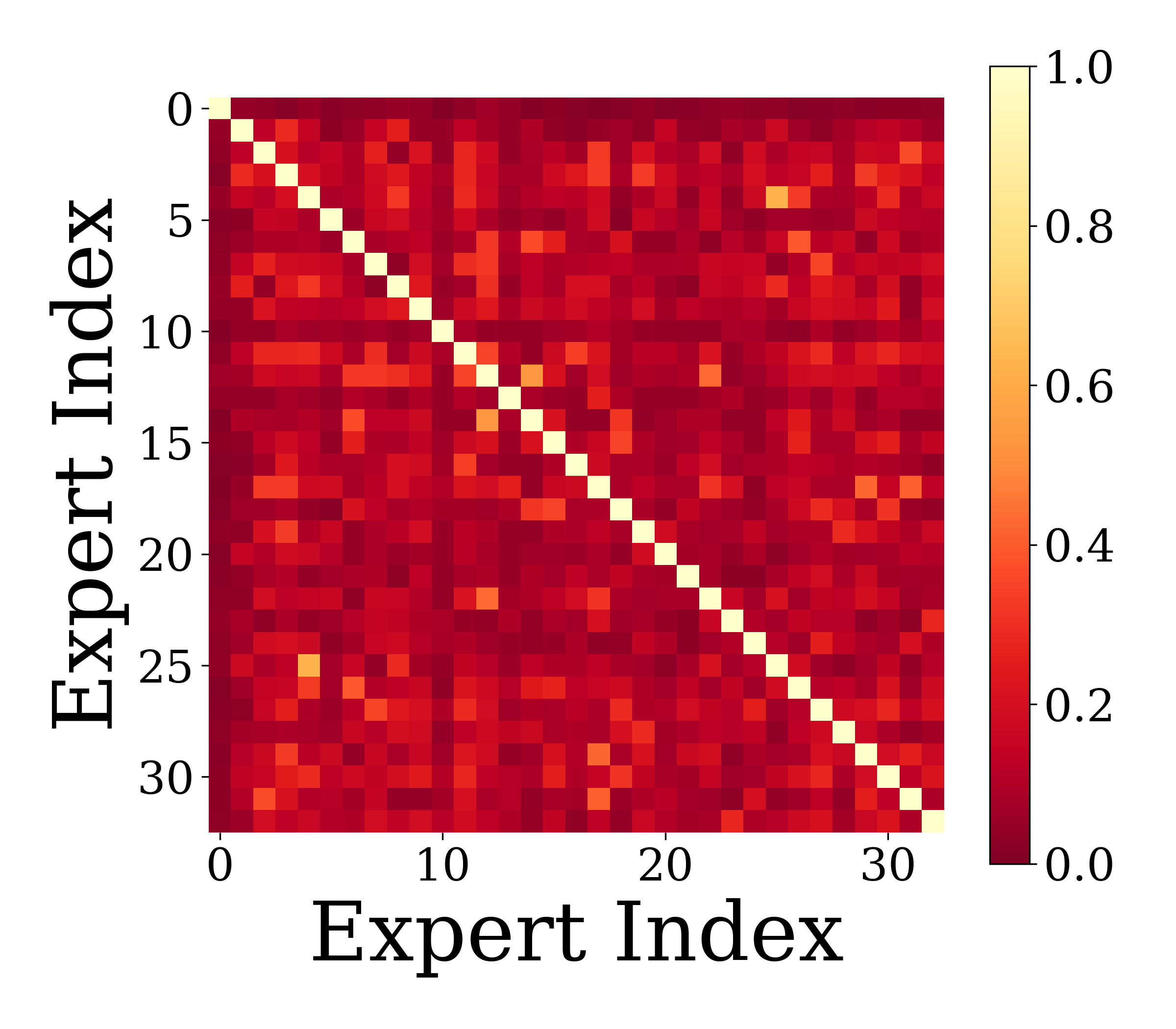}
        \label{fig:intro_param_spec_sim_our}
    }
    \caption{
        \textbf{(a)} The top 1\% spectral subspace in DeepSeek model is highly aligned (similarity 0.8) and carries $>$30\% of the energy~\cite{ethayarajh2019contextual, puccetti2022outlier, cao2025metis}, while the remaining 99\% tail is weakly aligned ($\sim$0.1).
        \textbf{(b--d)} Pairwise spectral similarity of expert parameters for \textbf{(b)} Qwen1.5-MoE~\cite{qwen_moe}, \textbf{(c)} DeepSeek-V2-Light~\cite{dai2024deepseekmoe}, and \textbf{(d)} SD-MoE. Existing MoE models exhibit strong overlap in the top 1\% spectral subspace (avg.~0.7; some $>$0.9), whereas SD-MoE reduces this similarity to $\sim$0.1.
        Supplementary results in more models and layers are in Appendix~\ref{append:more_results} Figure~\ref{fig:append.all_expert_sigvals}-\ref{fig:append.all_expert_prin_sim_ds}. 
    }
\label{fig:intro_param_spectrum}
\end{figure*}

This paper addresses this gap through a spectral analysis of both parameter and gradient spaces.
We formalize expert parameter overlap, uncover the training dynamics that give rise to limited specialization, and propose a new MoE architecture that enables effective specialization among experts. Our key findings are summarized as follows:

\noindent\textbf{Experts share highly aligned dominant spectral components, even when a common expert is explicitly used.}
As shown in Figure~\ref{fig:intro_param_spectrum}, a large fraction of spectral energy is concentrated in a small number of leading singular directions, and these dominant subspaces are highly aligned across experts with a similarity larger than 0.9.
Even in models that explicitly employ a shared (or “common”) expert, such as Qwen~\cite{qwen_moe} and DeepSeek~\cite{dai2024deepseekmoe}, the weight matrices of unique experts still exhibit strongly overlapping dominant spectral directions.

\begin{figure}[t]
    \centering
    \subfigure[]{
        \includegraphics[width=.45\linewidth]{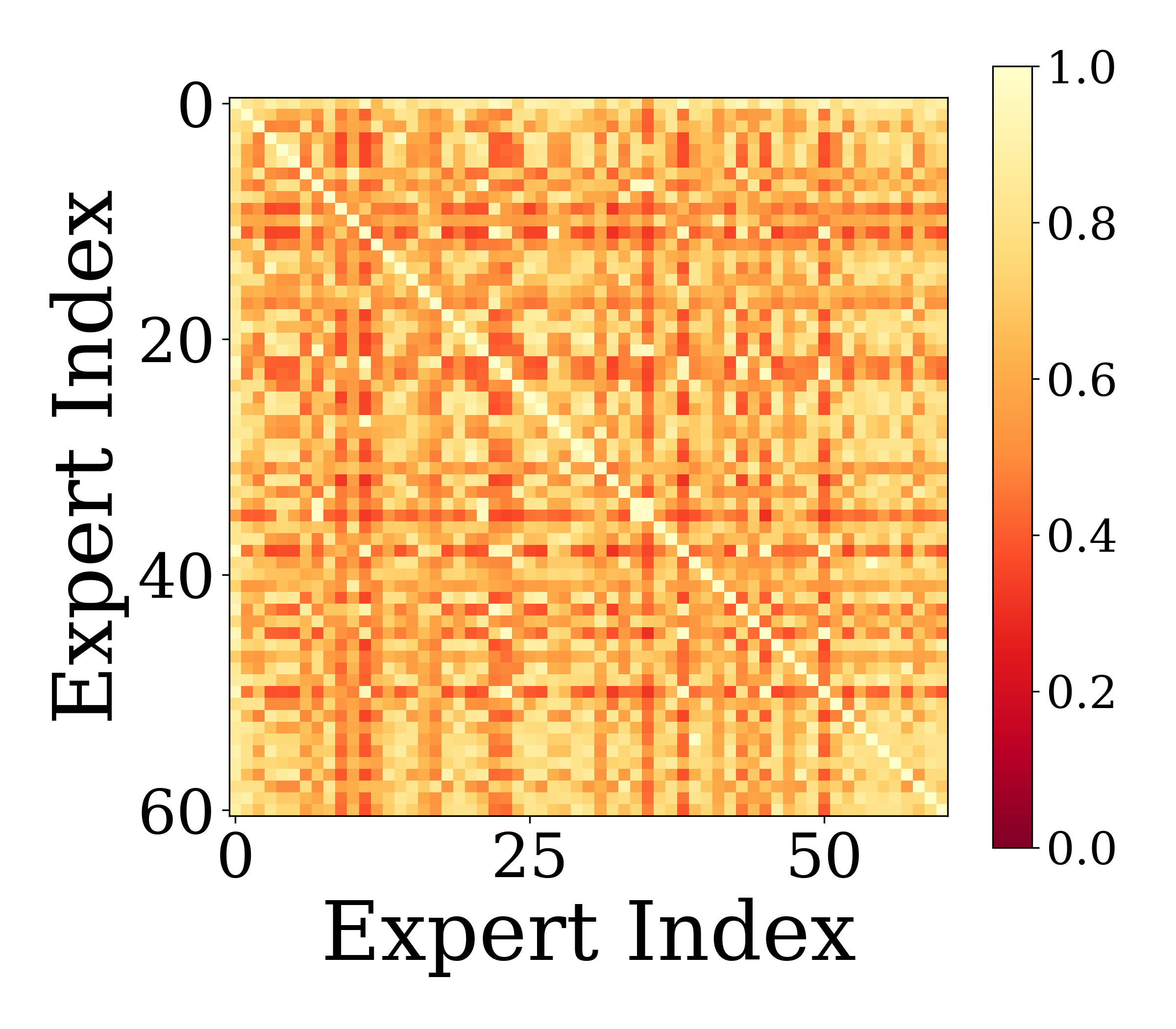}
        \label{fig:intro_grad_lowrank_spectrum_sim}
    }
    \subfigure[]{
        \includegraphics[width=.45\linewidth]{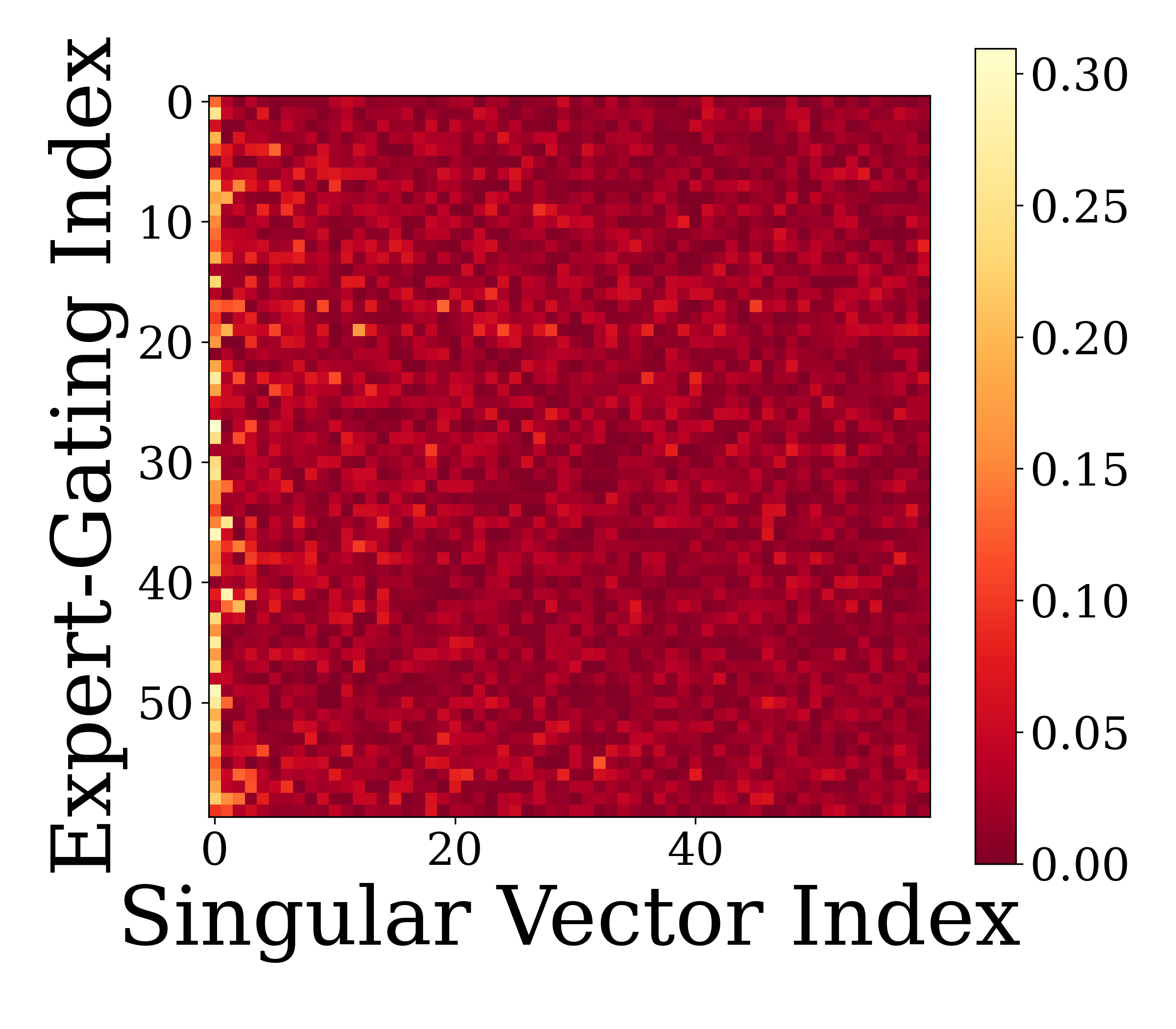}
        \label{fig:intro_gate_expert_aligment}
    }
    \caption{
        Analysis of Qwen1.5-MoE-A2.7B.
        \textbf{(a)} Pair-wise principal similarity of the dominant low-rank subspace from the gradient matrices of all experts. High values indicate near-identical spectral directions across experts, suggesting gradient similarity in their low-rank subspace.
        \textbf{(b)} The row vectors of the gating matrix exhibit high alignment with the leading singular directions of the expert weight matrices, indicating that the gating mechanism is dominated by common information.
    }
\label{fig:intro_grad_spectrum}
\end{figure}

\draft{
\noindent\textbf{The dominant gradient spectral spaces are highly aligned across experts, driving overlap in their parameter subspaces.}
As illustrated in Figure~\ref{fig:intro_grad_lowrank_spectrum_sim}, the dominant low-rank components of expert gradients are strongly aligned across experts, indicating a persistent shared update direction during training.
In contrast, the remaining long-tail directions exhibit substantially lower cross-expert similarity, suggesting that they capture more expert-specific variations.
As detailed in Section~\ref{subsec:shared_grad}, this alignment originates from ubiquitous structure in natural-language representations: input activations concentrate in a shared low-rank subspace across data samples.
During backpropagation, this shared activation subspace induces a common right-singular subspace in the gradient matrices of all experts, which in turn drives their dominant parameter updates to align.
}

\noindent\textbf{Dominant gradient subspaces are highly aligned across experts, driving parameter overlap.}
As illustrated in Figure~\ref{fig:intro_grad_lowrank_spectrum_sim}, the dominant low-rank components of expert gradients exhibit strong cross-expert alignment, reflecting a shared update direction during training.
In contrast, the remaining long-tail directions exhibit substantially lower cross-expert similarity, suggesting that they capture more expert-specific variations.
As detailed in Section~\ref{subsec:shared_grad}, this alignment arises from a shared low-rank structure in input representations, which induces a common right-singular subspace in expert gradients and consequently aligns their dominant parameter updates.

\noindent\textbf{Gating is dominated by leading spectral structure, resulting in non-specialized routing.}
As shown in Figure~\ref{fig:intro_gate_expert_aligment}, the row vectors of gating weight matrices exhibit strong alignment with the dominant spectral directions of expert weight matrices.
This indicates that the gating mechanism itself fails to fulfill its intended role of promoting expert specialization; instead, it routes inputs based on the same shared low-rank features, which undermines effective expert specialization.

These findings reveal that shared dominant spectral components induce parameter redundancy and gating bias, diminishing the effective capacity of existing MoE models and ultimately degrading downstream performance~\cite{zhou2022mixture, qiu2025demons, guo2025advancing}.

To address this spectral entanglement, we propose \textit{Spectral-Decoupled MoE (SD-MoE)}.
SD-MoE decomposes each expert matrix into a shared low-rank spectral subspace, and an orthogonal, expert-specific complement in the long-tail subspace.
During training, the gradient is similarly decomposed: the projection onto the shared subspace updates the common expert, while the residual updates corresponding unique experts.
SD-MoE provides three key advantages:

•	\textit{Improved performance:} The method yields on average 3\% gains on downstream tasks and 30\% training efficiency improvement.

•	\textit{Effective expert specialization:} SD-MoE reduces inter-expert similarity to below 0.1 (Figure~\ref{fig:intro_param_spec_sim_our}) and enabling more diverse semantic routing.

•	\textit{Low overhead and broad applicability:} SD-MoE introduces only ~5\% additional computational cost and can be integrated seamlessly into existing MoE architectures, including Qwen~\cite{qwen_moe} and DeepSeek~\cite{dai2024deepseekmoe}.

\section{Spectral Analysis for Expert Specialization}
\label{sec:analysis}
This section analyzes expert specialization failure through three lenses: parameter redundancy, gradient alignment, and gating bias in a shared dominant spectral subspace.

Our analysis centers on widely used open-source MoE models Qwen~\cite{qwen_moe} and DeepSeek~\cite{dai2024deepseekmoe}, which replace the standard FFN with $E$ routed experts. For each token representation $\mathbf{x}_t\in\mathbb{R}^{d}$, a router selects a small subset of experts $\mathcal{E}_t\subseteq\{1,\dots,E\}$ and aggregates their outputs as $\mathbf{h}_t=\sum_{i\in\mathcal{E}_t} g_{t,i}\,\mathrm{FFN}_i(\mathbf{x}_t)$, where $g_{t,i}\ge 0$ are weights over $i\in\mathcal{E}_t$ (typically normalized to sum to $1$). Each expert is a SwiGLU block: $\mathrm{FFN}_i(\mathbf{x})=\mathbf{W}^{(i)}_{D}\!\big(\sigma(\mathbf{W}^{(i)}_{G}\mathbf{x})\odot(\mathbf{W}^{(i)}_{U}\mathbf{x})\big)$, where $\odot$ is element-wise product.

\subsection{Spectral Overlap in Expert Parameters}
\label{subsec:param_sing}
\emph{Objective 1: Character expert specialization in parameter spectral space.}
We begin by examining specialization at the level of expert parameters.
In a MoE architecture, specialization implies that different experts implement \emph{distinct transformations}~\cite{guo2025advancing}.
We therefore characterize each expert’s transformation through its spectral structure, using the singular vectors obtained from singular value decomposition (SVD).

\emph{Dominant spectral subspaces and overlap metric.}
Formally, consider a linear projection in expert $i$ and write its transformation as $\mathbf{y}=\mathbf{W}^{(i)}\mathbf{x}$,
where $\mathbf{W}^{(i)}\in\mathbb{R}^{d_{\text{out}}\times d_{\text{in}}}$, $\mathbf{x}\in\mathbb{R}^{d_{\text{in}}}$, and $\mathbf{y}\in\mathbb{R}^{d_{\text{out}}}$.
With singular value decomposition,
\begin{equation}
\mathbf{W}^{(i)}=\mathbf{U}^{(i)}\boldsymbol{\Sigma}^{(i)}\mathbf{V}^{(i)\top},
\end{equation}
the columns of $\mathbf{V}^{(i)}$ form orthonormal directions in the input feature space,
while the columns of $\mathbf{U}^{(i)}$ form orthonormal directions in the output feature space.

SVD provides singular directions for both the input and output spaces, so we must choose the appropriate basis for comparing experts.
In SwiGLU, the up- and gate-projections operate in the \emph{input} feature space, whereas the down-projection mixes directions in the \emph{output} feature space.
Accordingly, we define the comparison basis for each expert matrix as
$$
\mathbf{B}^{(i)} :=
\begin{cases}
\mathbf{V}^{(i)}, & \text{up\_proj / gate\_proj (input feature space)},\\
\mathbf{U}^{(i)}, & \text{down\_proj (output feature space)}.
\end{cases}
$$

For any spectral index interval $[s:e]$, we define the corresponding subspace of expert $i$ as
\[
\mathcal{S}^{(i)}_{[s:e]} := \mathrm{span}\!\left(\mathbf{B}^{(i)}_{s:e}\right),
\]
where $\mathbf{B}^{(i)}_{s:e}$ denotes columns $s,\dots,e$ of $\mathbf{B}^{(i)}$.

To quantify specialization, we measure the overlap between expert subspaces using principal subspace similarity:
\begin{equation}
\mathrm{sim}\!\left(\mathcal{S}^{(i)}_{[s:e]}, \mathcal{S}^{(j)}_{[s:e]}\right)
:= \lambda_{\max}\!\left(
\mathbf{B}^{(j)\top}_{s:e}\mathbf{B}^{(i)}_{s:e}
\right),
\label{equa:principal_sim}
\end{equation}
where $\lambda_{\max}(\cdot)$ denotes the largest singular value.
Higher values indicate stronger alignment between expert subspaces and thus weaker specialization.

\emph{Finding 1: Dominant spectral subspaces are highly aligned across expert parameters.}
Applying Eq.~\eqref{equa:principal_sim} across consecutive $1\%$ spectral intervals (and reporting the corresponding energy CDF),
we observe a sharp head--tail contrast in inter-expert similarity (Fig.~\ref{fig:intro_param_spectrum}):
the first $1\%$ interval attains very high similarity $\approx 0.8$,
whereas the remaining intervals are much lower, roughly $\approx 0.1$ on average.
Together with the strongly concave energy CDF, this suggests that a small leading subspace both carries most of the transformation energy and dominates inter-expert similarity.
Therefore, in the rest of this paper we refer to the dominant-subspace as the top $1\%$ directions unless otherwise specified.

\emph{Objective 2: Analyze the functional role of shared dominant spectral subspaces.}
The overlap metric captures similarity in expert parameters, but does not reveal the \emph{functional role} of the shared dominant subspace.
To understand what these shared parameter directions represent, we further analyze how they are activated by real language inputs.

For any input token $\mathbf{x} \in \mathbb{R}^{d_{\text{in}}}$, its projection onto the $m$-th singular direction is $\langle \mathbf{v}_m, \mathbf{x} \rangle$.
We say that $\mathbf{x}$ \emph{activates} the $m$-th direction if
$| \langle \mathbf{v}_m, \mathbf{x} \rangle | > \tau$,
where $\tau$ is a layer-specific threshold chosen to match the scale of activations in each layer.
All activation experiments are conducted using DCLM data~\cite{li2024dclm} on Qwen1.5-MoE~\cite{qwen_moe}.

\begin{figure}[h]
    \centering
    \subfigure[]{
        \includegraphics[width=.45\linewidth]{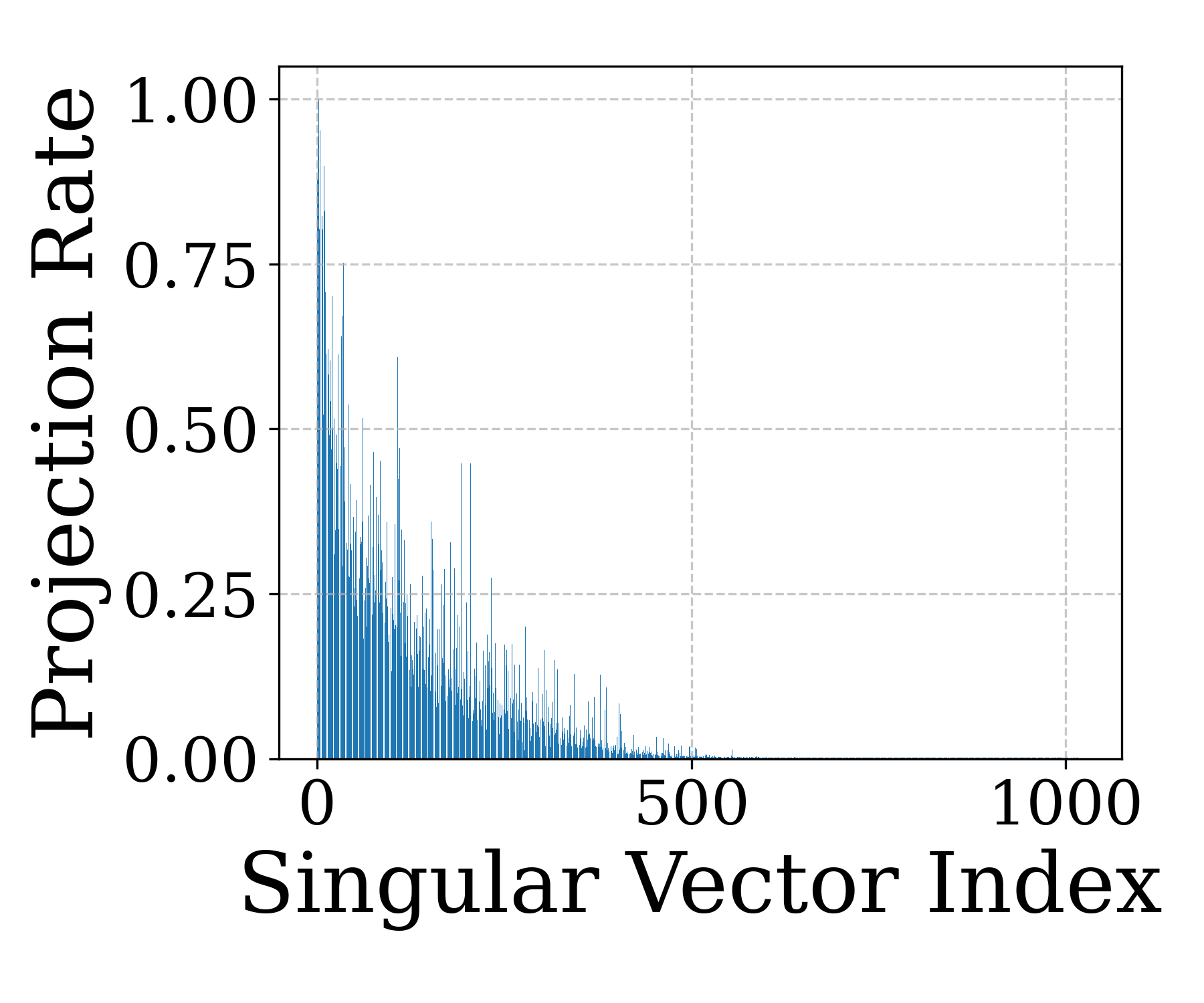}
        \label{fig:param_activation}
    }
    \subfigure[]{
        \includegraphics[width=.45\linewidth]{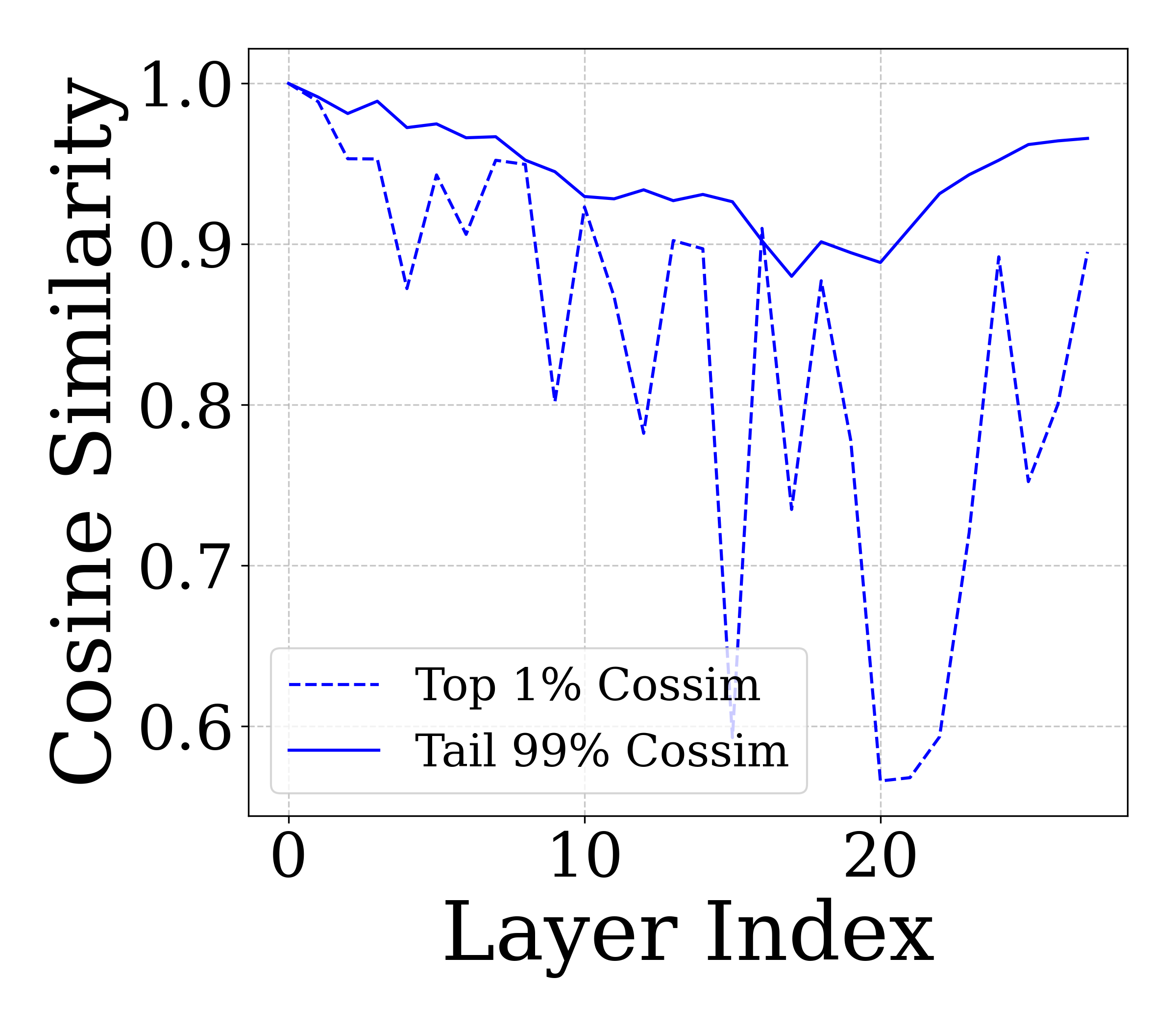}
        \label{fig:act_param_cossim}
    }
    \caption{
        \textbf{(a)} Activation ratio of each singular direction, defined as the fraction of tokens whose projections exceed the activation threshold.
        Leading singular directions are activated by nearly all tokens, indicating that they encode features consistently present across tokens and domains.
        \textbf{(b)} After shuffling input tokens, the activation projections onto the top $1\%$ singular directions changes larger than that for the tail $99\%$,
        indicating that the dominant subspace is highly sensitive to syntactic structure.
    }
    \label{fig:param_spectrum}
\end{figure}

As shown in Figure~\ref{fig:param_activation}, the activation ratios across singular directions exhibit a pronounced long-tailed distribution.
The leading singular directions are activated by nearly all input tokens, whereas the vast majority of directions in the spectral tail are rarely activated.
This indicates that the dominant subspace encodes broadly applicable features shared across inputs, rather than expert-specific behavior.

\emph{Finding 2: Dominant spectral subspace in parameters encodes mainly syntactic structures.}
To further characterize the information encoded by the dominant subspace, we shuffle input tokens to see how projection varies.
As shown in Figure~\ref{fig:act_param_cossim}, we compute the cosine similarity between activations projected onto the top $1\%$ and tail $99\%$ singular directions.
Shuffling causes a larger similarity drop for the dominant subspace, indicating that these directions are highly sensitive to syntactic structure.
This suggests that the shared dominant parameter directions encode general-purpose linguistic knowledge, such as syntax, rather than expert-specialized content.

\subsection{Shared Spectral Subspace in Expert Gradients}
\label{subsec:shared_grad}
\emph{Objective 3: Identify the training dynamics responsible for parameter overlap across experts.}
Having established that expert parameters exhibit highly aligned dominant subspaces, we now turn to the \emph{training dynamics} that give rise to this phenomenon.
We analyze the spectral structure of expert gradients to understand how such alignment emerges across experts.

\emph{Spectral geometry of expert gradients.}
Consider a same linear projection in expert $i$ $\mathbf{y} = \mathbf{W}^{(i)}\mathbf{x},$ (with $\mathbf{W}^{(i)}\in\mathbb{R}^{d_{\text{out}}\times d_{\text{in}}}$) as defined in Section~\ref{subsec:param_sing}, and let $\mathcal{B}_i$ denote the routed mini-batch for expert $i$ at a given training step.
The gradient of the loss with respect to $\mathbf{W}^{(i)}$ is

\begin{equation}
\mathbf{G}^{(i)} := \frac{\partial\mathcal{L}(\mathcal{B}_i)}{\partial\mathbf{W}^{(i)}}
=\sum_{t\in\mathcal{B}_i}\frac{\partial \ell_t}{\partial \mathbf{y}_t}\frac{\partial\mathbf{y}_t}{\partial\mathbf{W}^{(i)}}
= \sum_{t \in \mathcal{B}_i} \boldsymbol{\delta}_t \mathbf{x}_t^\top,
\label{equa:gradient_chain}
\end{equation}
where $\mathbf{x}_t \in \mathbb{R}^{d_{\text{in}}}$ is the input vector of token $t$,
$\mathbf{y}_t=\mathbf{W}^{(i)}\mathbf{x}_t$ is the token-level output of this linear map on token $t$,
and $\boldsymbol{\delta}_t := \frac{\partial\ell_t}{\partial\mathbf{y}_t}\in\mathbb{R}^{d_{\text{out}}}$ is the upstream error signal.
The singular value decomposition (SVD) of the gradient is thus
$ \mathbf{G}^{(i)} = \mathbf{U}_G^{(i)} \boldsymbol{\Sigma}_G^{(i)} (\mathbf{V}_G^{(i)})^\top $.

For concreteness, we focus on the input-side subspace and use the right singular vectors $\mathbf{V}_G^{(i)}$ in the following analysis.
The output-side case is completely analogous as is discussed in Section~\ref{subsec:param_sing}.
We therefore define the dominant gradient-update subspace as
$$
\mathcal{S}_G^{(i)} := \mathrm{span}\!\left( \mathbf{V}_{G,1:k}^{(i)} \right),\qquad 
\mathbf{V}_{G,1:k}^{(i)} := \mathbf{V}_G^{(i)}[:,1\!:\!k],
$$

with $k$ chosen to capture the top $1\%$ of singular directions (matching our parameter analysis).
Similar to Equation~\ref{equa:principal_sim}, cross-expert gradient alignment is also quantified via principal subspace similarity:
\begin{equation}
\mathrm{sim}\!\left( \mathcal{S}_G^{(i)}, \mathcal{S}_G^{(j)} \right)
:= \lambda_{\max}\!\left( (\mathbf{V}_{G,1:k}^{(j)})^\top \mathbf{V}_{G,1:k}^{(i)} \right).
\end{equation}

\begin{figure}[h]
    \centering
    \subfigure[]{
        \includegraphics[width=.45\linewidth]{figures/grad_spectrum_head.png}
        \label{fig:ana_grad_lowrank_spectrum_sim}
    }
    \subfigure[]{
        \includegraphics[width=.45\linewidth]{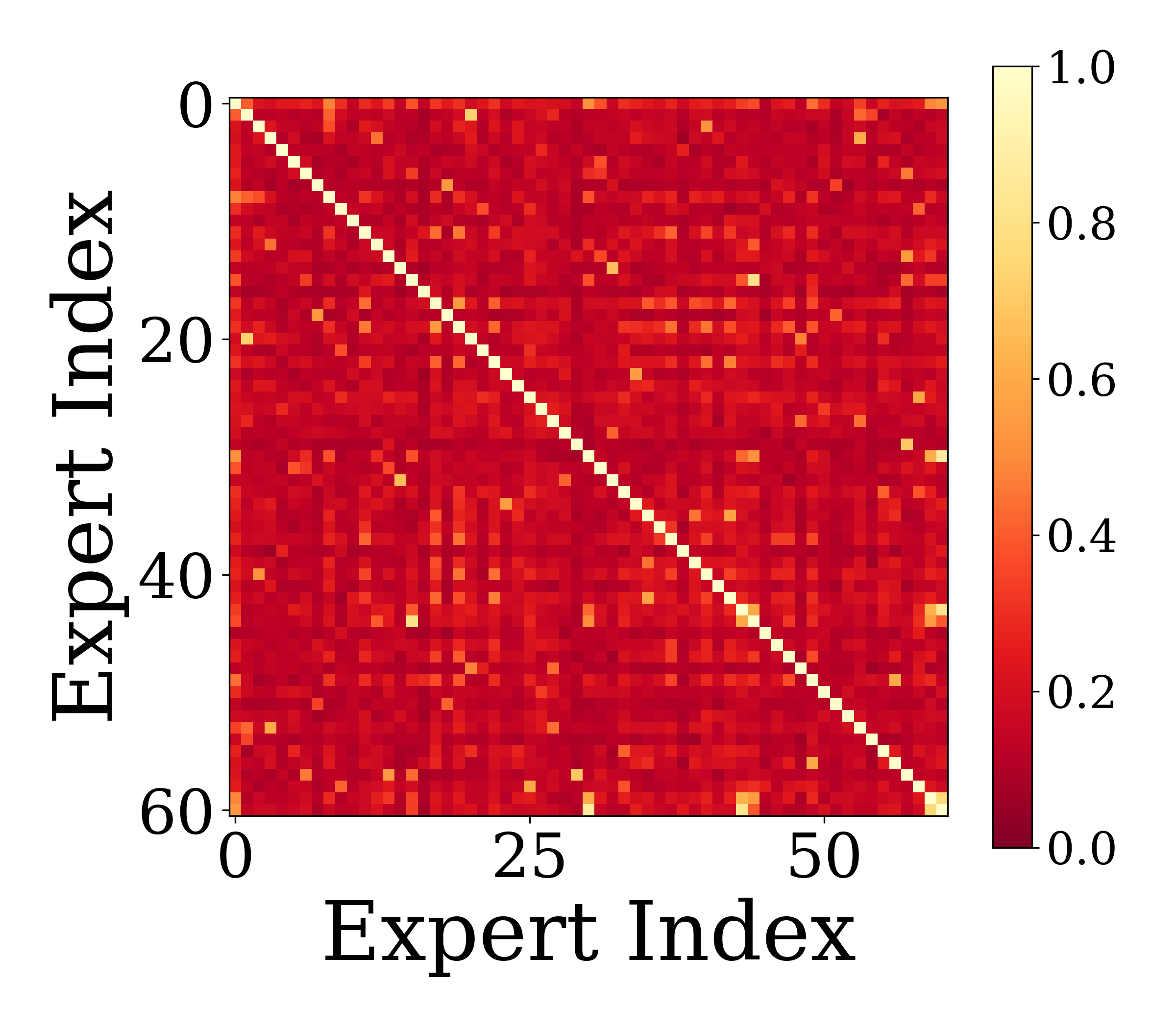}
        \label{fig:ana_grad_longtail_spectrum_sim}
    }
    \caption{
        Gradient spectral analysis of experts in Qwen1.5-MoE-A2.7B.
        Pairwise principal similarity of \textbf{(a)} the dominant low-rank (1\%) gradient subspace and 
        \textbf{(b)} the remaining long-tail subspace across all experts, including the common expert.
        High similarity in the low-rank subspace indicates shared gradient directions across experts, while tail similarity is substantially weaker.
        Additional results across models and layers are provided in Appendix~\ref{append:more_results} (Figures~\ref{fig:append.all_expert_grad_prin_sim} and~\ref{fig:append.all_expert_grad_tail_sim}).
    }
    \label{fig:ana_grad_spectrum}
\end{figure}

\emph{Finding 3: Dominant gradient subspaces are highly aligned across experts.}
Figure~\ref{fig:ana_grad_spectrum} reveals a striking pattern: the dominant $1\%$ gradient subspaces exhibit high alignment across all expert pairs (Figure~\ref{fig:ana_grad_lowrank_spectrum_sim}), despite receiving disjoint sets of tokens through gating.
In contrast, the long-tail $99\%$ subspace shows significantly weak alignment (Figure~\ref{fig:ana_grad_longtail_spectrum_sim}).
This indicates that the \emph{primary direction of parameter updates} is effectively shared among all experts.

\emph{Objective 4: Explain the origin of cross-expert alignment in dominant gradient subspaces.}Motivated by Finding~3, we seek to explain why the input-side dominant gradient-update subspace
$\mathcal{S}_G^{(i)}=\mathrm{span}(\mathbf{V}^{(i)}_{G,1:k})$ remains highly similar across experts
despite receiving disjoint routed tokens.
We investigate a representation-driven explanation: whether input activations exhibit a shared dominant subspace $C$
that persists across diverse samples and routed mini-batches, and whether this shared structure can induce
a common component in each expert gradient through $\frac{\partial\ell}{\partial\mathbf{W}}=\boldsymbol{\delta}\mathbf{x}^\top$.

\begin{figure}[h]
    \centering
    \includegraphics[width=.45\linewidth]{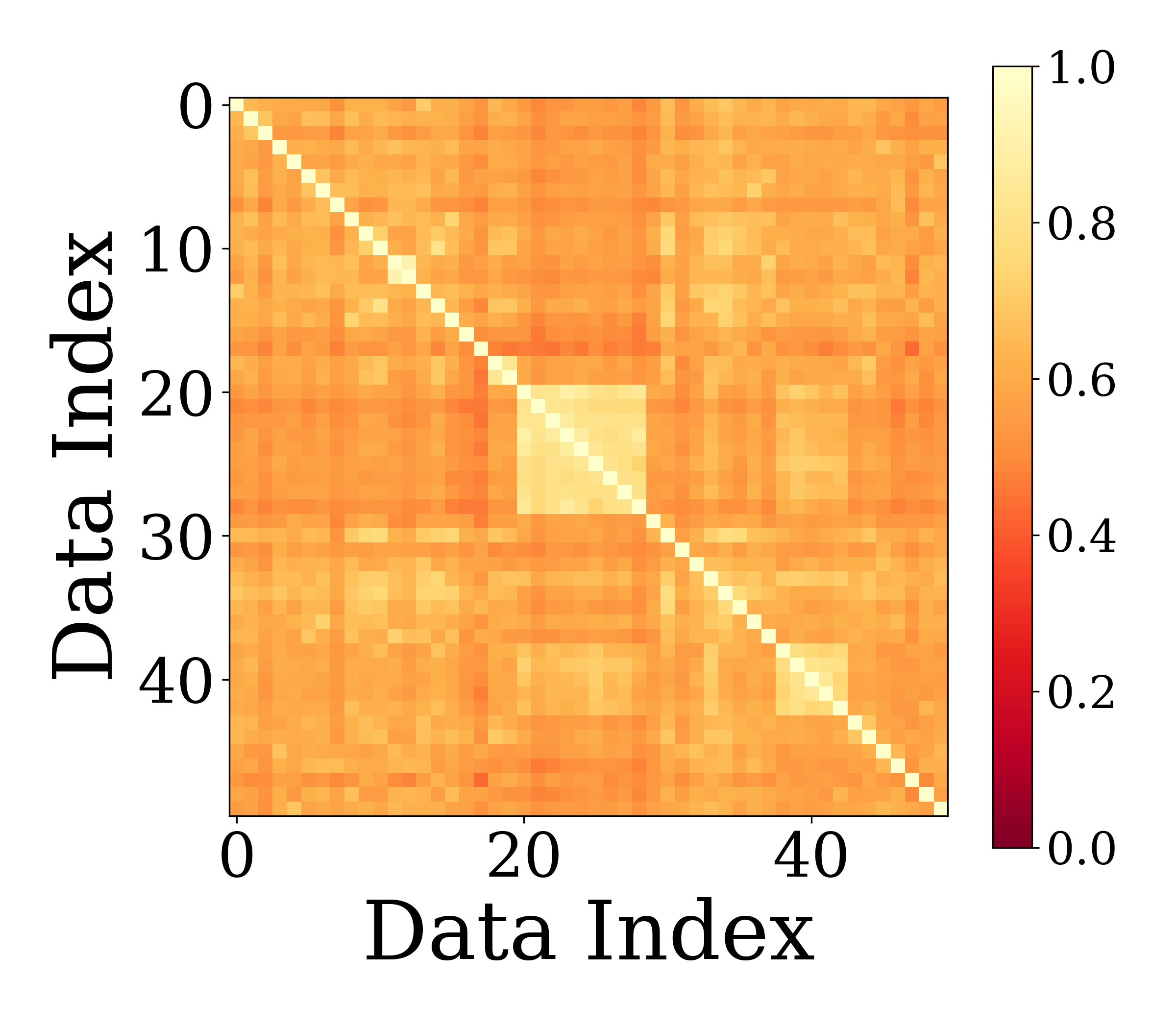}
    \caption{
        Pairwise principal subspace similarity of the top $1\%$ activation spectral directions across data samples, confirming the presence of a shared low-rank input subspace. Additional results across models and layers are provided in Appendix~\ref{append:more_results} (Figure~\ref{fig:append.all_data_act_alignment}).
    }
    \label{fig:activation_similarity}
\end{figure}

\emph{Finding 4: Shared low-rank structure in input representations drives gradient alignment.}
To validate our hypothesis, we compute activation tensors (shaped [sequence length, dimension]) from randomly sampled sequences and compute the principal similarity in their top $1\%$ spectral subspace.
As shown in Figure~\ref{fig:activation_similarity}, these dominant subspaces exhibit high pairwise cosine similarity across diverse data samples.
This confirms the existence of a shared low-rank structure in the input representation space.

We then outline the mathematical mechanism by which shared input structure drives gradient alignment.
For the same linear projection $\mathbf{y}=\mathbf{W}^{(i)}\mathbf{x}$
, the per-token gradient is $\frac{\partial\ell}{\partial\mathbf{W}} = \boldsymbol{\delta} \mathbf{x}^\top$, which is similar to Equation~\ref{equa:gradient_chain}.
Let $\mathbf{P}_C$ denote the orthogonal projector onto $C$, and define $\mathbf{P}_{C^\perp}:=\mathbf{I}-\mathbf{P}_C$.
Decomposing each input into a shared dominant subspace and a long-tailed subspace $\mathbf{x}_t = \mathbf{P}_C \mathbf{x}_t + \mathbf{P}_{C^\perp} \mathbf{x}_t$, the batch gradient becomes
\begin{equation}
    \mathbf{G}^{(i)} = \underbrace{\sum_{t \in \mathcal{B}_i} \boldsymbol{\delta}_t (\mathbf{P}_C \mathbf{x}_t)^\top}_{\mathbf{G}_C^{(i)}}
    + \underbrace{\sum_{t \in \mathcal{B}_i} \boldsymbol{\delta}_t (\mathbf{P}_{C^\perp} \mathbf{x}_t)^\top}_{\mathbf{R}^{(i)}}.    
\end{equation}
The term $\mathbf{G}_C^{(i)}$ presented in \emph{every} expert’s update acts entirely within the shared subspace $C$.
When $C$ captures most of the input energy (as observed), $\mathbf{G}_C^{(i)}$ dominates the spectral norm of $\mathbf{G}^{(i)}$,
forcing the leading right singular vectors of $\mathbf{G}^{(i)}$ to concentrate near $C$.
A more detailed analysis under mild dominance conditions is provided in Appendix~\ref{app:routing_to_grad}.

\subsection{Spectral Bias in Gating Decisions}
\label{sec:ana.gating}
\emph{Objective 5: Examine the impact of spectral dominance on gating behavior.}
Having shown that both activations and expert parameters have a shared dominant subspace, we now examine how the \emph{gating mechanism} is influenced by this spectral dominance.

\emph{Gating as directional matching.}
In standard Mixture-of-Experts (MoE) layers, the router computes a distribution over experts via
$$\mathbf{s} = \mathrm{softmax}\!\left( \mathbf{W}_{gate} \mathbf{x} \right)$$, 
where $\mathbf{x} \in \mathbb{R}^{d_{\text{in}}}$ is the input token representation and  $\mathbf{W}_{gate} \in \mathbb{R}^{E \times d_{\text{in}}}$ is a learnable gating matrix, where $E$ is the number of experts.
The  $ i $-th gating score is therefore
$$s_i \propto \exp\!\big( \mathbf{w}_i \mathbf{x} \big)$$,
where $\mathbf{w}_i$ denotes the  $ i $ -th \emph{row} of  $ \mathbf{W}_{gate} $ .
This reveals that gating is fundamentally a process of \emph{directional matching}: each expert is associated with a direction  $ \mathbf{w}_i $  in the input space, and tokens are assigned based on their inner product with these expert-specific vectors.

Consequently, we examine how these gating directions  $ \{\mathbf{w}_i\}_{i=1}^E $ relate to the spectral subspaces.

\begin{figure}[h]
    \centering
    \subfigure[]{
        \includegraphics[width=.45\linewidth]{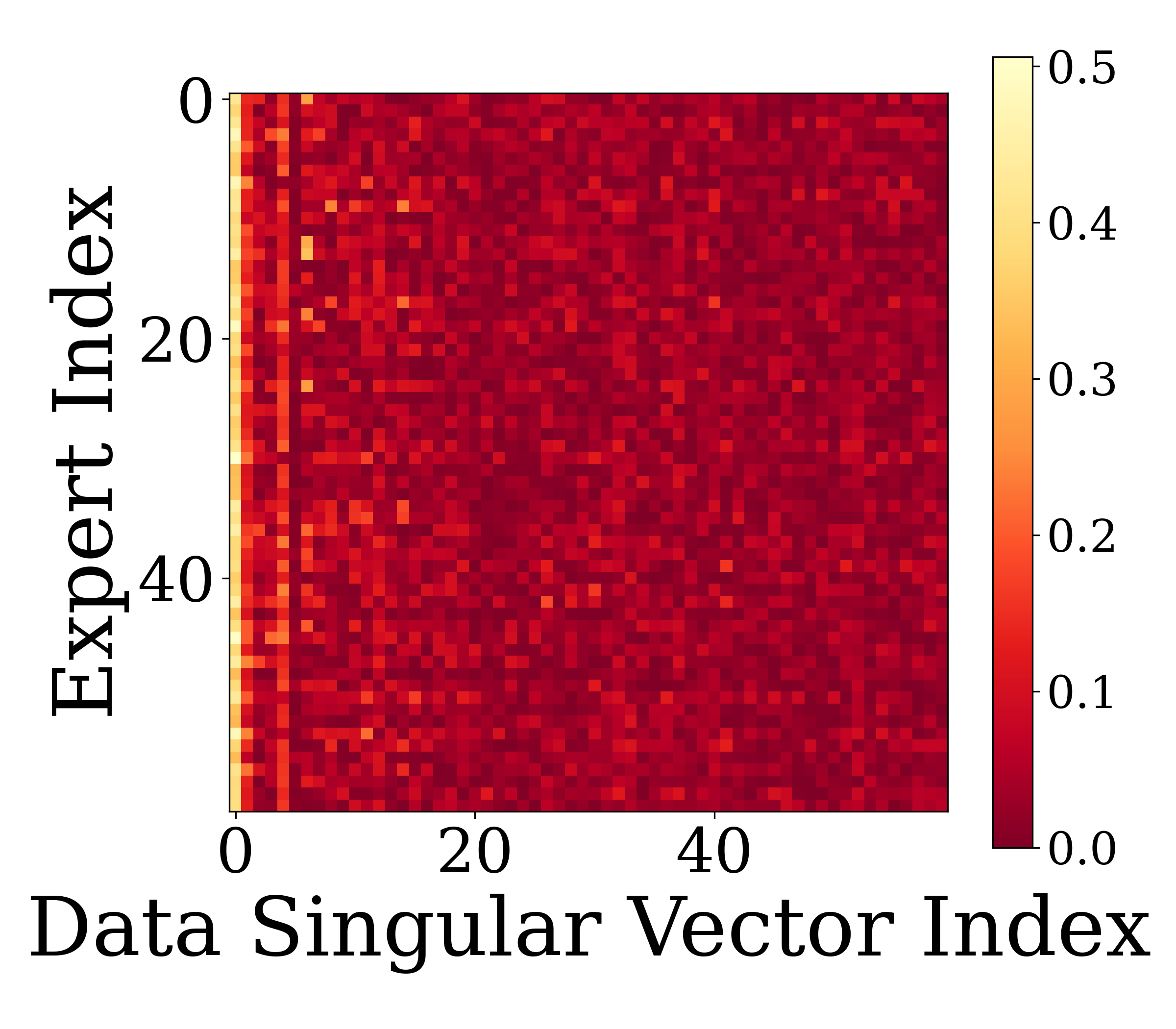}
        \label{fig:gate_activation_alignment}
    }
    \subfigure[]{
        \includegraphics[width=.45\linewidth]{figures/qwen_public_gating_expert_similarity_deep.png}
        \label{fig:gate_expert_aligment_deep}
    }
    \caption{
        Gate matrix analysis in Qwen1.5-MoE~\cite{qwen_moe}. The alignment of row vectors of the gating matrix with \textbf{(a)} the leading spectral directions of the data activation \textbf{(b)} the leading singular directions of the expert parameters. Gating mechanism is dominated by common information. Supplementary results in more models and layers are in Appendix~\ref{append:more_results} Figure~\ref{fig:append.all_gate_act_alignment} and \ref{fig:append.all_gate_param_spectral_alignment_qwen}.
    }
\label{fig:gate_expert_aligment}
\end{figure}

\emph{Finding 5: Gating aligns with shared spectral subspaces, fails to specialize.}We evaluate, for each  $ i $ , the alignment between the gating vector  $ \mathbf{w}_i $  and the dominant subspaces of (a) the input activations and (b) the parameters of the corresponding expert.
Similar to principal similarity defined in Equation~\ref{equa:principal_sim}
As shown in Figure~\ref{fig:gate_activation_alignment}, gating vectors exhibit strong alignment with the leading singular directions of the activation matrix.
Moreover, Figure~\ref{fig:gate_expert_aligment_deep} reveals consistent alignment between each gating vector  $ \mathbf{w}_i $  and the leading singular vector of its associated expert's parameter matrix.

\begin{figure*}[t]
    \centering
    \includegraphics[width=0.95\linewidth]{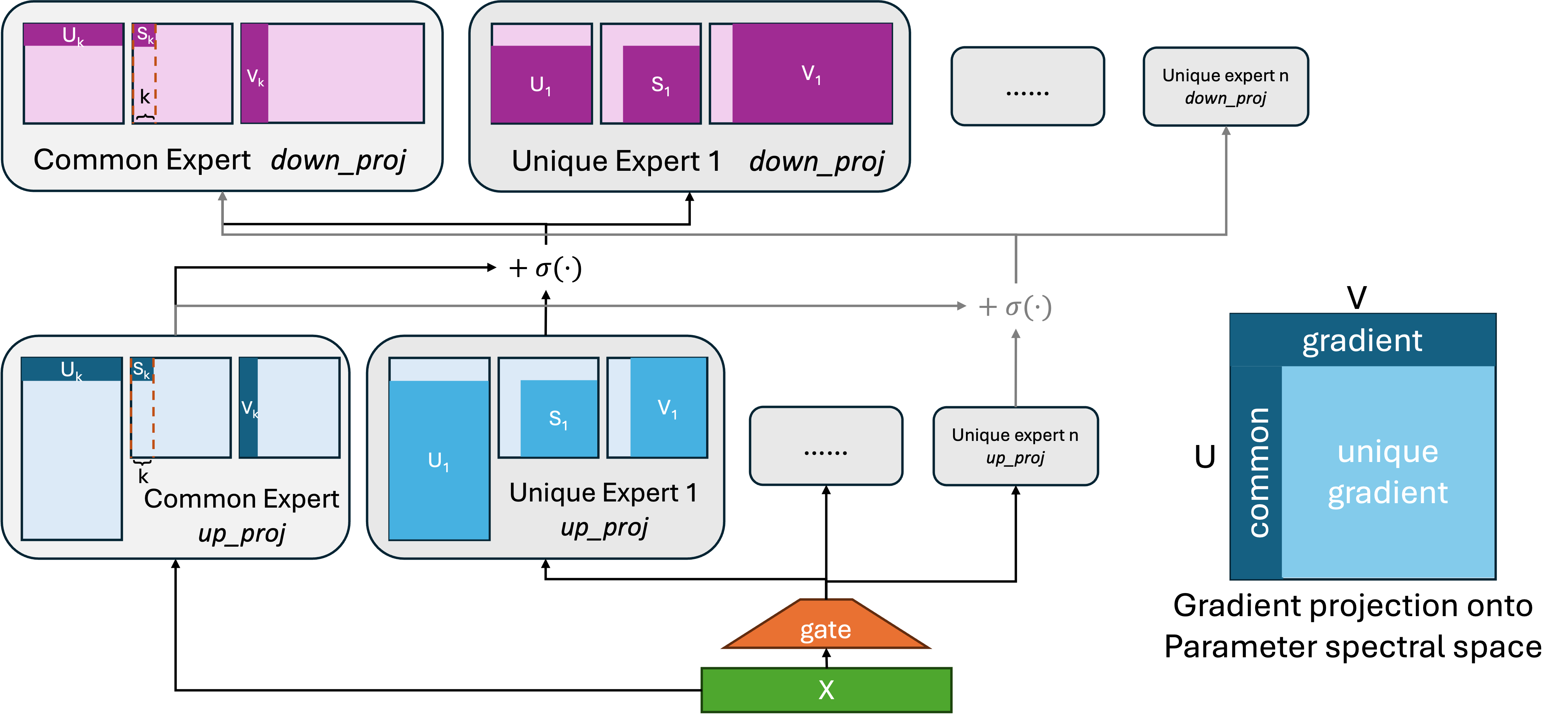}
    \caption{The architecture of SD-MoE. SD-MoE decompose each linear matrix into a shared low-rank subspace and multiple unique components in the orthogonal complement. During gradient updates, the gradient is also decomposed into these subspaces to update the experts accordingly.}
    \label{fig:method.model_structure}
\end{figure*}

\section{Spectral-Decoupled MoE}
\label{sec:method}
This section introduces Spectral-Decoupled Mixture of Experts (SD-MoE), a spectral decomposition-based approach that directly addresses the considerable overlap among experts in parameter and gradient spaces. As demonstrated in Section~\ref{sec:experiments}, this design reduces expert overlap by 70\%, improves training efficiency by 30\% on MoE architectures based on DeepSeek and Qwen, indicating its broad applicability across diverse existing MoE architectures.

\subsection{Expert Spectral Decomposition}

SD-MoE optimizes expert specialization though
(i) spectrally decouples each expert's parameter matrix into a shared low-rank component and an expert-specific component within its orthogonal complement; and
(ii) decouples the gradient updates for these two components to enable independent optimization. This design reduces inter-expert gradient interference and variance, relaxes learning rate constraints, and accelerates convergence. 

Figure~\ref{fig:method.model_structure} uses a two-layer MLP as an example to illustrate how to apply SD-MoE to the standard linear-gate MoE paradigm. Given an input token $\mathbf{x}$, routing scores are computed as $\mathbf{W}_{\text{gate}} \mathbf{x}$, where $\mathbf{W}_{\text{gate}}$ is a learnable gating matrix.
The top-n experts with the highest scores are selected for activation.  
Each expert $i$ consists of two weight matrix: an up-projection $\mathbf{W}_U^{(i)}$ and a down-projection $\mathbf{W}_D^{(i)}$.
The output is computed as
$\mathbf{W}_D^{(i)}\sigma(\mathbf{W}_U^{(i)} \mathbf{x}),$
where $\sigma(\cdot)$ denotes the activation function.

However, architectural decoupling alone is insufficient to ensure sustained specialization during training. Section~\ref{sec:method.initialization} addresses this by spectrally decoupling expert parameters at initialization, preventing dominant shared components from biasing early optimization, while Section~\ref{sec:method.gradient_update} further decomposes gradients during training to prevent cross-expert interference from re-emerging. Together, these steps ensure that spectral decoupling is preserved throughout training.

\subsection{Parameter spectral decoupling at initialization}
\label{sec:method.initialization}
For each weight matrix in the experts, SD-MoE decomposes it into (i) a shared component $\mathbf{W}_c$ for the common expert, and 
(ii) multiple expert-specific components $\mathbf{W}_u^{(i)}$ for unique experts.
Let $\mathbf{U} \boldsymbol{\Sigma} \mathbf{V}^\top$ be random sampled singular vectors and singular values, $k$ be the rank of dominant subspace. We define $\mathbf{U}_k=\mathbf{U}_{[:,:k]}$, $\mathbf{V}_k=\mathbf{V}_{[:,:k]}$, $\boldsymbol{\Sigma}_k = \boldsymbol{\Sigma}_{[:k, :k]}$.
The matrix in the common expert is initialized as
\begin{equation}
\mathbf{W}_c := \mathbf{U}_k\,\boldsymbol{\Sigma}_k\,\mathbf{V}_k^\top,
\end{equation}
The matrix in each unique expert $\mathbf{W}_u^{(i)}$ is then initialized such that it lies in the orthogonal complement of the common subspace: 
\begin{equation}
    \mathbf{U}_k^\top \mathbf{W}_u^{(i)}=\mathbf{0}, \quad \mathbf{W}_u^{(i)}\mathbf{V}_k=\mathbf{0}.
\end{equation}
Details of how unique experts are initialized are deferred to Appendix~\ref{appendix:method}.

\subsection{Gradient spectral decomposition during training}
\label{sec:method.gradient_update}
Before detailing our gradient decomposition strategy, we first outline the forward pass, which determines how gradients are subsequently computed.

\textbf{Forward Pass} The routing mechanism in SD-MoE follows the standard linear MoE router.
If a token $\mathbf{x}$ is routed to expert $i$, SD-MoE computes the forward pass of each linear matrix in the expert as
\begin{equation}
    \mathbf{W}^{(i)}\mathbf{x} = (\mathbf{W}_c+\mathbf{W}_u^{(i)})\mathbf{x}.
\end{equation}
where we refer to $\mathbf{W}^{(i)}$ as a \emph{proxy expert matrix} corresponding to the unique expert $\mathbf{W}_u^{(i)}$.
The same operation is applied to all linear matrices, and thus the output is $\mathbf{W}_D^{(i)}\sigma(\mathbf{W}_U^{(i)}\mathbf{x})\,$
where $\mathbf{W}_U^{(i)} = \mathbf{W}_{U,c} + \mathbf{W}_{U,u}^{(i)}, \quad \mathbf{W}_D^{(i)} = \mathbf{W}_{D,c} + \mathbf{W}_{D,u}^{(i)}$.

\textbf{Gradient Decomposition} During back-propagation, each proxy expert matrix $\mathbf{W}^{(i)}$ produces a gradient
$\mathbf{G}^{(i)}=\frac{\partial\mathcal{L}}{\partial\mathbf{W}^{(i)}}$.
We decompose this gradient into two parts:
(i) a \emph{low-rank} component $\mathbf{G}_c^{(i)}$ used to update the common expert matrix $\mathbf{W}_c$,
and
(ii) a \emph{long-tail} component $\mathbf{G}_u^{(i)}$ used to update the unique expert matrix $\mathbf{W}_u^{(i)}$.
Let $\mathbf{P}_U=\mathbf{U}_k\mathbf{U}_k^\top$ and $\mathbf{P}_V=\mathbf{V}_k\mathbf{V}_k^\top$.
The gradient is decomposed as
\begin{align}
    \mathbf{G}_c^{(i)} &:= \mathbf{P}_U \mathbf{G}^{(i)} + (\mathbf{I}-\mathbf{P}_U)\,\mathbf{G}^{(i)}\,\mathbf{P}_V, \\
    \mathbf{G}_u^{(i)} &:= \mathbf{G}^{(i)}-\mathbf{G}_c^{(i)}
\end{align}
An intuitive illustration of this decomposition is shown in Figure~\ref{fig:method.model_structure}.
In the spectral space spanned by the left and right singular subspaces $\mathbf{U}$ and $\mathbf{V}$, gradient components involving the dominant low-rank subspaces $\mathbf{U}_k\mathbf{V}_k^\top$ are absorbed into $\mathbf{G}_c^{(i)}$, while the remaining components in the double orthogonal complement are assigned to $\mathbf{G}_u^{(i)}$.
Since the basis for the common expert may change after update, we perform SVD on $\mathbf{W}_c$ periodically to obtain correct $\mathbf{U}_k, \mathbf{V}_k$.

\section{Experiments}
\label{sec:experiments}

\begin{table*}[t]
\centering
\begin{tabular}{lcccccc}
    \toprule
    \multirow{2}{*}{Metric} & \multicolumn{4}{c}{Small (2B--0.8B)} & \multicolumn{2}{c}{Large (7B--1.5B)} \\
    \cmidrule(lr){2-5} \cmidrule(lr){6-7}
    & DeepSeek & DeepSeek w. SD & Qwen & Qwen w. SD & Qwen & Qwen w. SD \\
    \midrule
    ARC-challenge & 30.80 & \textbf{31.66} & 31.06 & \textbf{36.09} & 35.75 & \textbf{40.70} \\
    ARC-easy & 62.16 & \textbf{63.43} & 62.04 & \textbf{68.39} & 69.91 & \textbf{71.84} \\
    HellaSwag & 54.11 & \textbf{55.57} & 55.34 & \textbf{59.48} & 61.12 & \textbf{66.13} \\
    LAMBADA\_OPENAI & 39.63 & \textbf{40.21} & 35.09 & \textbf{42.97} & \textbf{49.31} & 47.29 \\
    PIQA & 71.98 & \textbf{72.80} & 72.31 & \textbf{74.16} & 74.81 & \textbf{76.39} \\
    RACE & \textbf{50.51} & 50.06 & 48.54 & \textbf{50.61} & 50.65 & \textbf{51.48} \\
    SIQA & 40.53 & \textbf{41.25} & 40.89 & \textbf{41.71} & 41.76 & \textbf{41.91} \\
    Winogrande & 54.33 & \textbf{56.91} & \textbf{58.33} & \textbf{58.33} & 59.43 & \textbf{62.43} \\
    \midrule
    Avg & 50.51 & \textbf{51.49} & 50.45 & \textbf{53.97} & 55.34 & \textbf{57.23} \\
    \bottomrule
\end{tabular}
\caption{
    Accuracy (\%) on 8 downstream tasks. Our SD-MoE framework is applied to various MoE architectures, including Qwen and DeepSeek, demonstrating its broad applicability across different model scales and structures. This table shows that integrating SD into existing MoE models improves performance across nearly all tasks, making it a reliable enhancement for any MoE-based architecture.
}
\label{tab:downstream}
\end{table*}

\subsection{Experiment Setup}
\label{sec:experiments.setup}
Our experiments are based on widely available open-source MoEs DeepSeek~\cite{dai2024deepseekmoe} and Qwen~\cite{yang2025qwen3}, which are explicitly designed to encourage expert specialization.
We conduct comparisons at two scales: (2B/0.8B) and (7B/1.5B).
All models are trained on a 100B subset of the DCLM corpus~\cite{li2024dclm}.
The shared expert subspace rank is set to 8, and we empirically fix the periodic SVD update interval to every 16 steps.
Full details on hyperparameters are provided in Appendix~\ref{appendix:exp-settings}.

\subsection{Main Results}
\begin{figure}[h]
    \centering
    \includegraphics[width=0.85\linewidth]{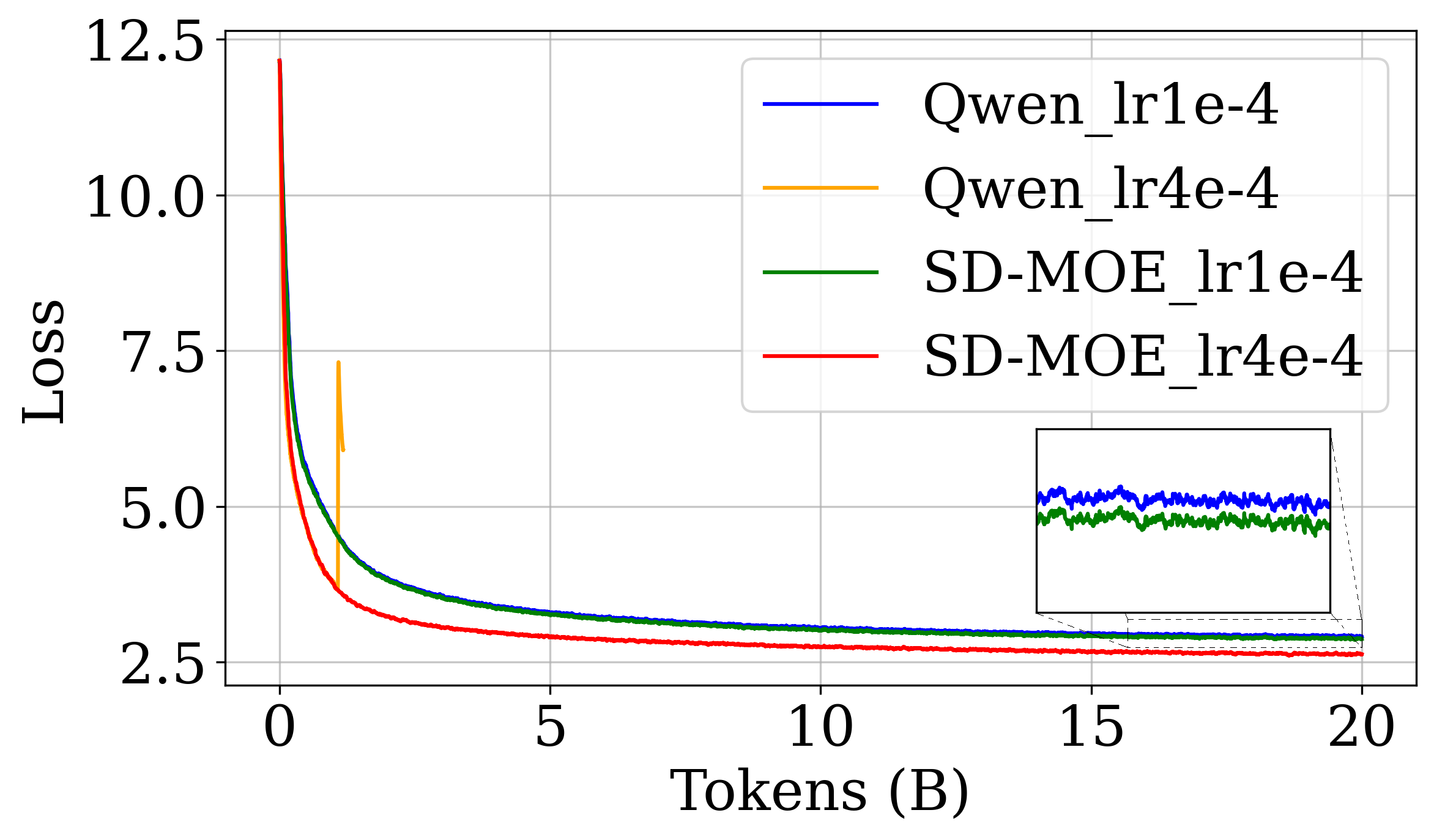}
    \caption{
        Compared to Qwen baseline, SD-MoE achieves lower loss and tolerates higher learning rates at 4$\times$.
    }
    \label{fig:main_results_lr}
\end{figure}

On downstream tasks (Table~\ref{tab:downstream}), SD-MoE delivers better performance across nearly all benchmarks compared to baselines. 
Notably, spectral sharing and decomposition further enable training with larger learning rates with out instability, delivering a 30\% training efficiency.
As is shown in Figure~\ref{fig:main_results_lr}, while baseline Qwen model diverges at learning rates beyond $1\times 10^{-4}$, SD-MoE remains stable up to a 4× increase $4\times 10^{-4}$.
More details for the larger learning rate experiments can be referred to in Appendix~\ref{app:same_lr}.

\paragraph{Spectral conditioning and training stability.}
Prior work has shown that training stability and convergence speed are closely tied to the spectral geometry of network mappings: well-conditioned singular spectra, often described through \emph{dynamical isometry}, permit larger stable learning rates and faster optimization \cite{pennington2017resurrecting,xiao2018dynamical}. Related approaches further demonstrate that explicitly regulating dominant singular components, such as through spectral normalization, stabilizes training by controlling the effective Lipschitz constants of learned transformations \cite{miyato2018spectral, chen2023over, shi2023train}.

In MoE models, orthogonality- and decoupling-based methods reduce cross-expert redundancy and gradient interference, improving specialization and optimization~\cite{liu2023diversifying}. Building on this insight, our spectral decomposition separates shared low-rank components from expert-specific subspaces, removes common spectral “spikes,” yields more benign conditioning across experts, and enables larger stable learning rates in practice.

\subsection{Analysis}
\label{sec:exp.spectral}
\begin{figure}[h]
    \centering
     \subfigure[]{
        \includegraphics[width=.45\linewidth]{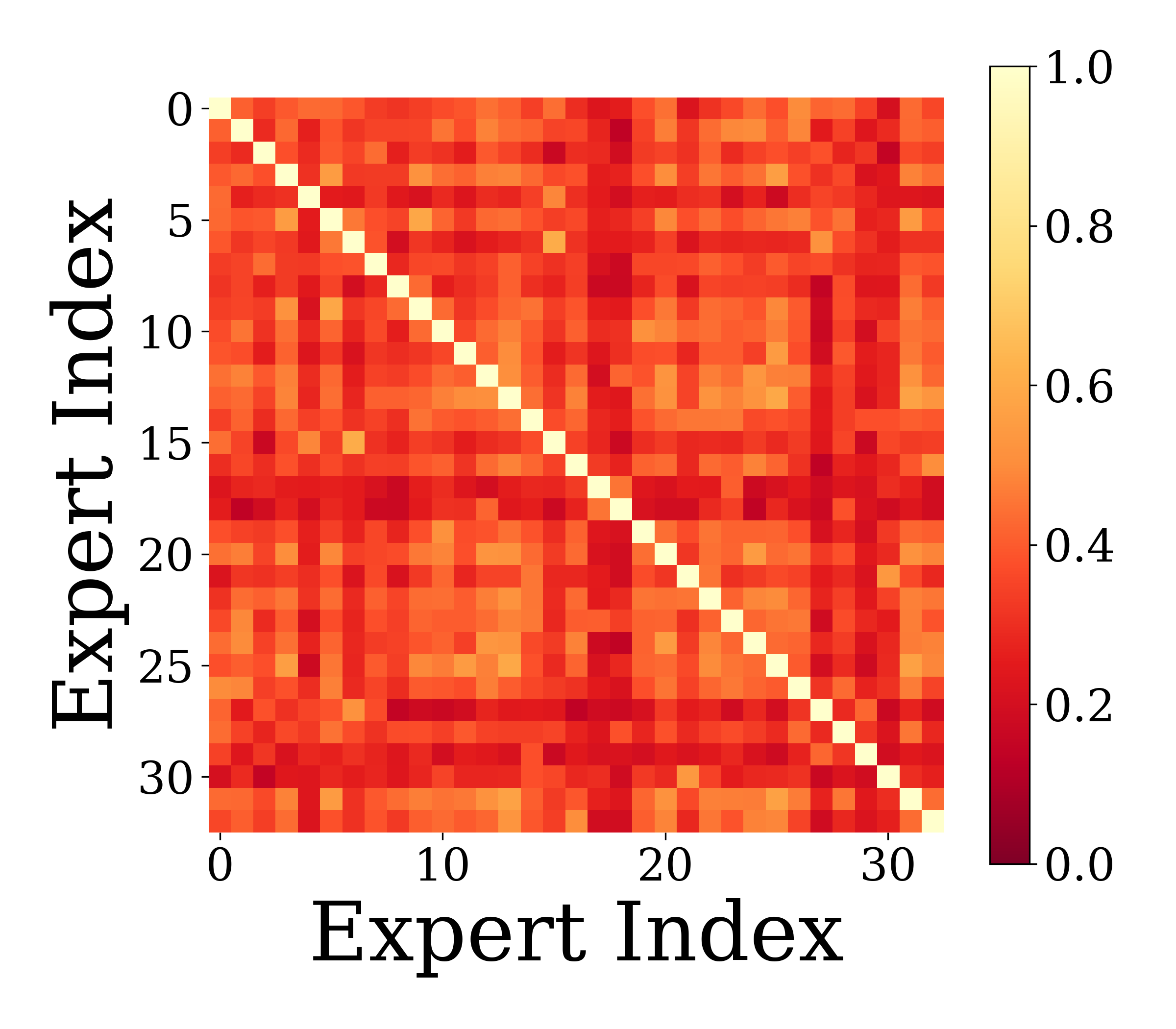}
        \label{fig:exp.qwen_subspace_similarity}
    }
    \subfigure[]{
        \includegraphics[width=.45\linewidth]{figures/our_expert_similarity_exp.png}
        \label{fig:exp.ours_subspace_similarity}
    }
    \caption{
        Pairwise principal subspace similarity between the dominant spectral subspaces of all experts in \textbf{(a)} Qwen baseline model, \textbf{(b)} our SD-MoE. The near-zero similarities across all expert pairs indicate effective decoupling of parameter directions, demonstrating successful expert specialization and reduced parameter redundancy. Supplementary results in more models and layers are in Appendix~\ref{append:more_results} Figure~\ref{fig:append.all_expert_prin_sim_qwen_native} and \ref{fig:append.all_expert_prin_sim_sdmoe}.
    }
    \label{fig:spectral_disentanglement}
\end{figure}

\begin{figure}[h]
    \centering
    \subfigure[]{
        \includegraphics[width=.45\linewidth]{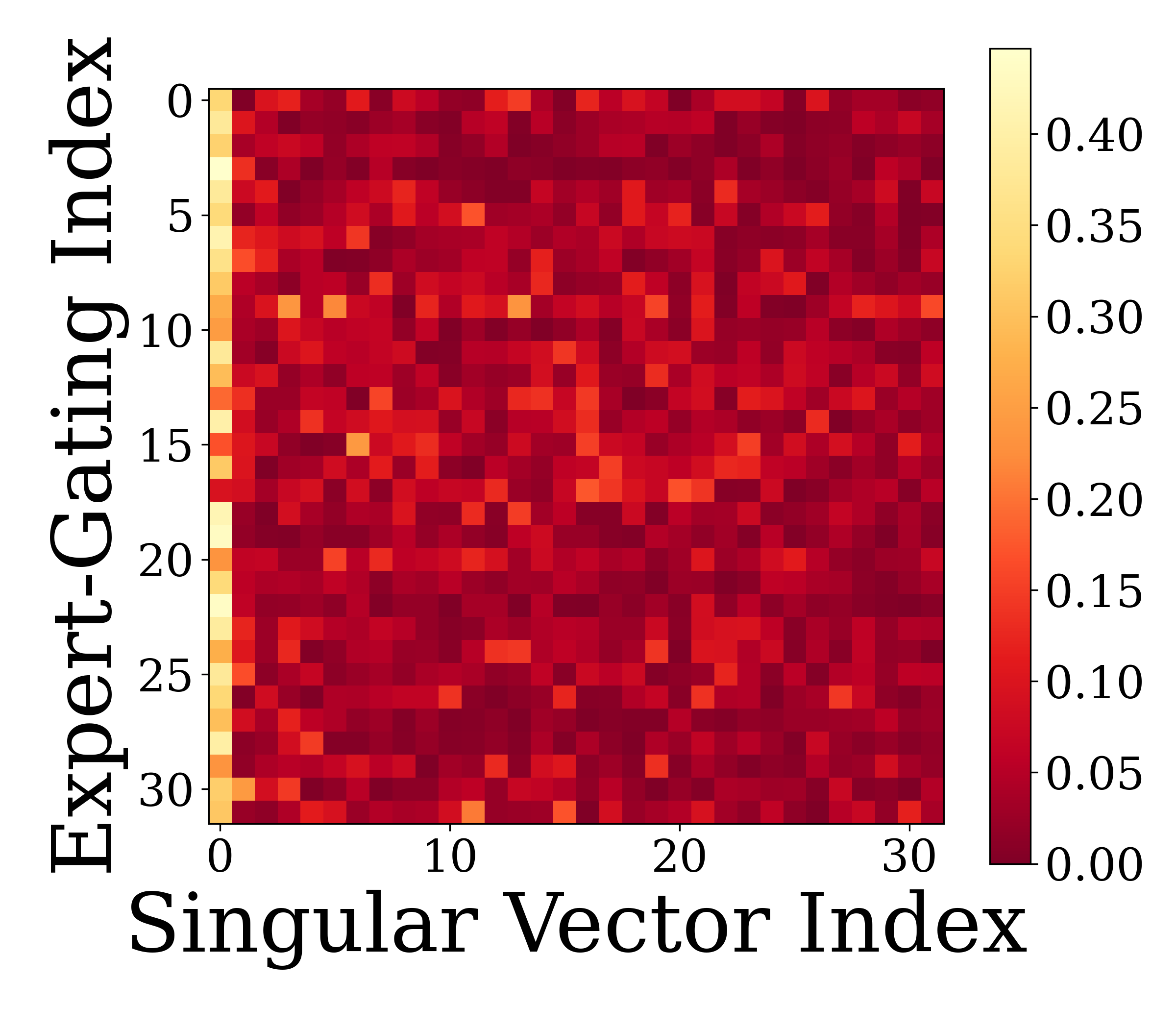}
        \label{fig:qwen_gate_expert_alignment}
    }
    \subfigure[]{
        \includegraphics[width=.45\linewidth]{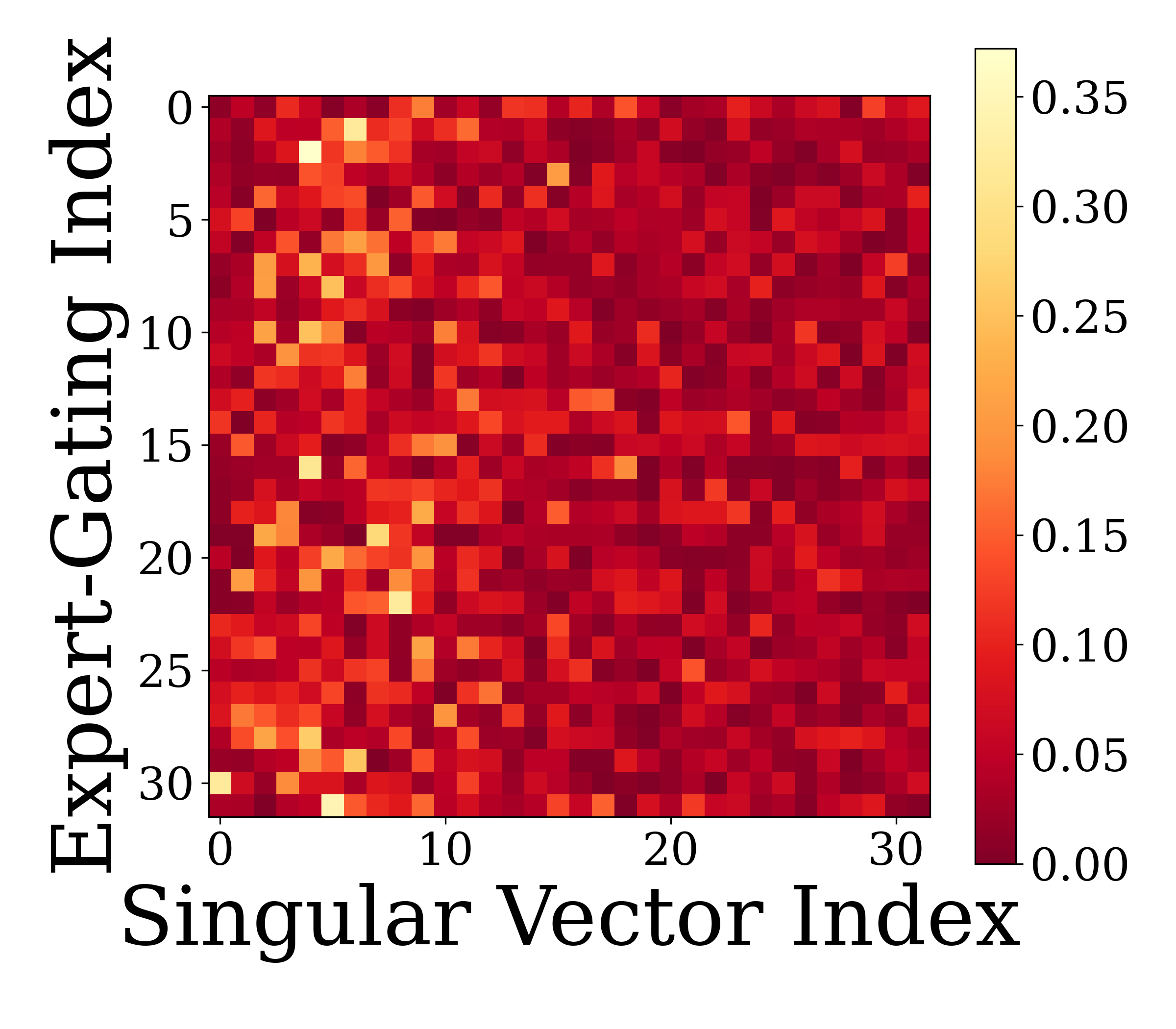}
        \label{fig:ours_gate_expert_alignment}
    }
    \caption{
        The alignment of row vectors of the gating matrix with the leading singular directions of the expert parameters in \textbf{(a)} Qwen baseline model and \textbf{(b)} our SD-MoE. In our SD-MoE, gating is not dominated by the largest singular vector. Supplementary results in more models and layers are in Appendix~\ref{append:more_results} Figure~\ref{fig:append.all_gate_param_spectral_alignment_qwen_native} and \ref{fig:append.all_gate_param_spectral_alignment_sdmoe}.
    }
    \label{fig:main_results_gating}
\end{figure}

\paragraph{Expert \& Gating Specialization.}
To verify whether our model achieves effective specialization across experts as expected, we compute the principal subspace similarity among all experts as defined in Section~\ref{sec:analysis}.
As shown in Figure~\ref{fig:spectral_disentanglement}, the pairwise principal subspace similarities are lower than 0.1 at convergence, indicating minimal overlap in their parameter directions.
The gating mechanism also exhibits improved specialization.
Compared to Qwen model in Figure~\ref{fig:qwen_gate_expert_alignment}, the row vectors of the gating matrix in SD-MoE in Figure~\ref{fig:ours_gate_expert_alignment} no longer exhibit strong alignment with the top singular directions of the expert parameters. Instead, they distribute their influence across a broader set of singular vectors, reflecting the use of richer, more diverse semantic cues during routing.
This confirms that SD-MoE successfully promotes expert specialization and  reduces parameter redundancy.

\paragraph{Sensitivity to Common Rank.}
We systematically evaluate $k \in \{2, 4, 8, 16, 32\}$.
The average task performance is listed in Table~\ref{tab:k_sensitivity}.
Despite the peak at $k=8$, performance remains stable across the entire range, varying by less than 0.6\%.
This suggests that the shared structure present in human-generated text is inherently low-rank. Moreover, our method is robust to the choice of $k$: as long as $k$ is modestly sized, it effectively captures the common subspace and thereby enhances expert specialization.

\begin{table}[h]
\centering
\caption{
    Model performance with different common subspace rank $k$.
    Detailed results are in Appendix~\ref{appendix:rank_abl} Table~\ref{tab:rank_ablation}.
}
\label{tab:k_sensitivity}
\begin{tabular}{lcccccc}
    \toprule
    $k$ & 2 & 4 & 8 & 16 & 32 \\
    \midrule
    Task Acc.  & 53.64 & 53.39 & \textbf{53.97} & 53.57 & 53.68 \\
    \bottomrule
\end{tabular}
\end{table}

\paragraph{Computation Overhead.}
Table~\ref{tab:throughput} reports training throughput (tokens/sec) under identical hardware conditions.
SD-MoE incurs around 5\% overhead compared to standard MoE at training time.
At inference time, SD-MoE introduces no additional computational cost, as the number of activated parameters remains unchanged. Overall, SD-MoE still achieves higher practical learning efficiency in wall-clock time for sustaining larger learning rates.

\begin{table}[h]
\centering
\caption{
    Training throughput (tokens/second) under the same computational resources for each model scale.
}
\label{tab:throughput}
\begin{tabular}{lccc}
    \toprule
    Model Scale & Qwen & SD-MoE (Ours) & $\Delta$ \\
    \midrule
    2B-A0.8B  & 229K & 218K & 4.80\% \\
    7B-A1.5B  & 248K & 235K & 5.24\% \\
    \bottomrule
\end{tabular}
\end{table}

\section{Related Works}
\label{sec:related_works}

\textbf{Mixture of Experts}
\cite{smoe} extended the MoE structure to deep neural networks and proposed a deep MoE model composed of multiple layers of routers and experts.
Since then, the MoE layer with different base neural network structures~\cite{gatedConv, transformer} has been proposed.
Following works explored various routing strategies like (i) letting tokens select the top-k experts~\cite{G-shard, switchtransformer, stochasticexperts, representationcollapse, stablemoe, smoeasdropout}, (ii) letting experts select the top-k tokens~\cite{expertchoicerouting}, to (iii) globally decide expert assignment~\cite{baselayers, scalinglawformoe}.
Recently, MoEs have been widely adopted as a core component in large-scale language models deployed by leading organizations, including models such as Qwen~\cite{yang2025qwen3}, DeepSeek~\cite{dai2024deepseekmoe}, and Mixtral~\cite{jiang2024mixtral}.

\textbf{Specialization failure and routing-based remedies.}
Despite the success of sparse routing, MoE models do not always yield meaningful expert specialization in practice: 
experts may become functionally similar, some experts can dominate the routing decisions~\cite{krishnamurthy2023improving, cai2025longtailed}, 
and the overall capacity gain can be reduced by redundancy and representation collapse~\cite{representationcollapse, liu2023diversifying, zhang2025mixtureexpertslargelanguage}.
Specialization from a semantic perspective have also been proposed~\cite{dong2024once, zhou2025oracle}. 
A common line of work therefore focuses on \emph{routing-level} improvements,
such as load-balancing objectives, capacity constraints, and stochastic routing/regularization variants~\cite{smoeasdropout, stochasticexperts}.
More recently, alternative routing paradigms (e.g., expert-choice routing) have also been explored to mitigate imbalance by changing the assignment mechanism~\cite{expertchoicerouting}.

\section{Conclusion}
\label{sec:conclusion}
This work addresses failure of expert specialization in MoE models. Our spectral analysis reveals that shared spectral components in parameters and gradients limit gating and expert differentiation. We resolve this with SD-MoE, which improves gating and expert specialization, enhances downstream performance, and is broadly applicable to existing MoE architectures.

\section*{Impact Statement}
This work aims to deliver insights to the Machine Learning community and will not lead to any direct societal consequences.
While it is associated with LLMs that may output misleading or harmful content, such issues are outside the scope of this work and will not be particularly specified here.

\newpage
\bibliographystyle{icml2026}

% \bibliography{references}

\newpage
\appendix
\section{More Results in Analysis}
\label{append:more_results}

\begin{figure*}[h]
    \centering
    \subfigure[]{
        \includegraphics[width=.22\linewidth]{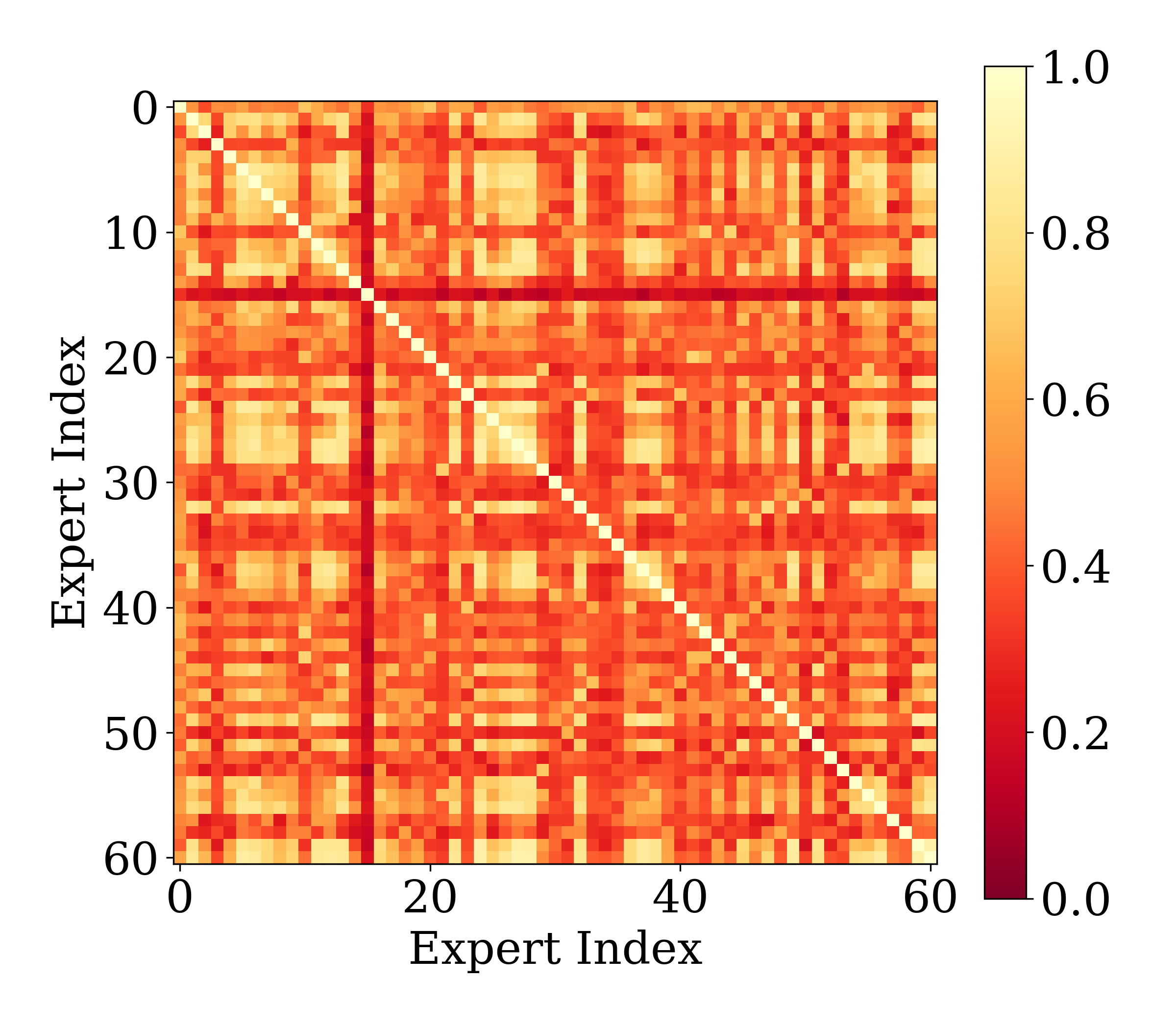}
    }
    ~ % space
    \subfigure[]{
        \includegraphics[width=.22\linewidth]{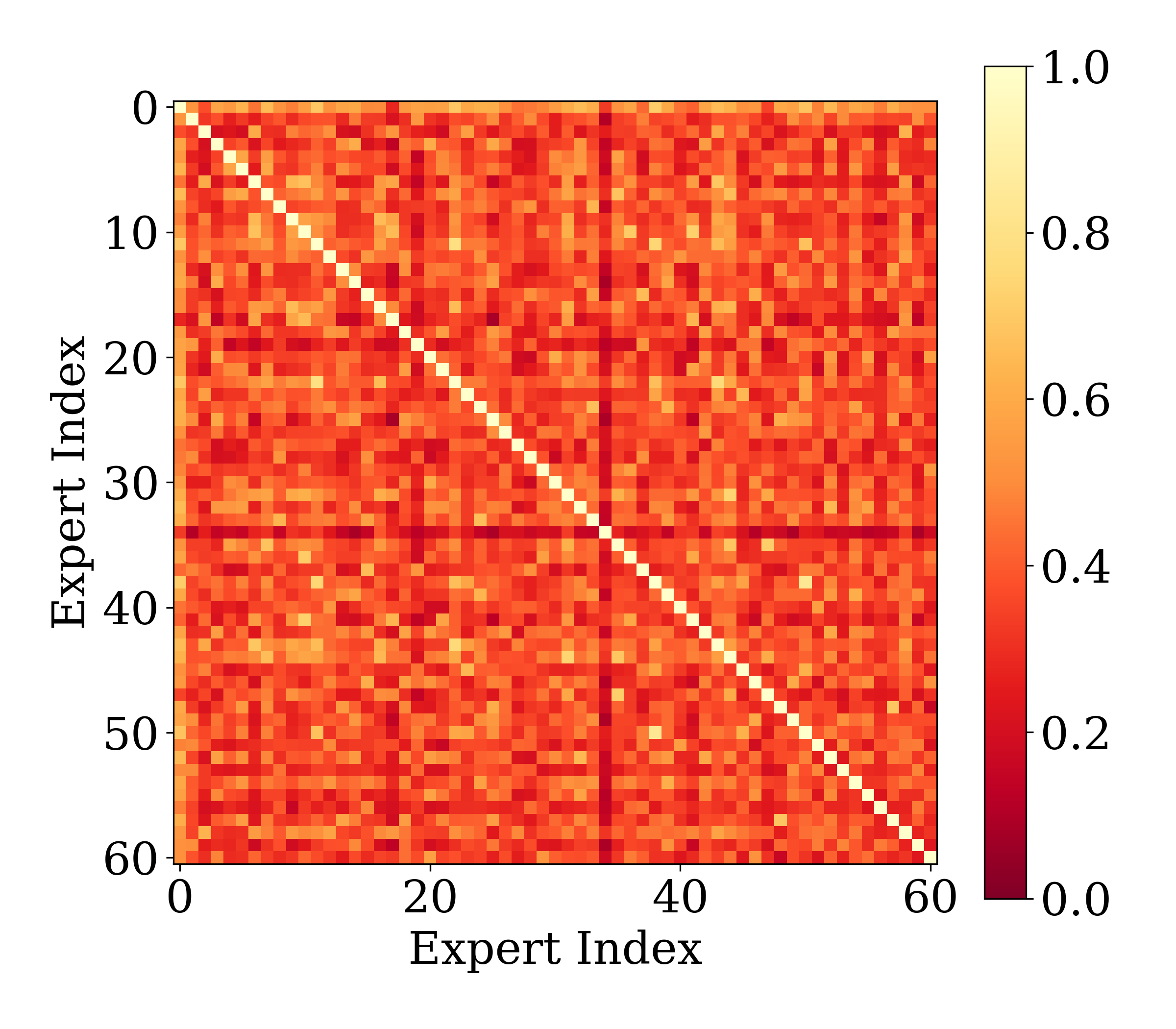}
    }
    ~ % space
    \subfigure[]{
        \includegraphics[width=.22\linewidth]{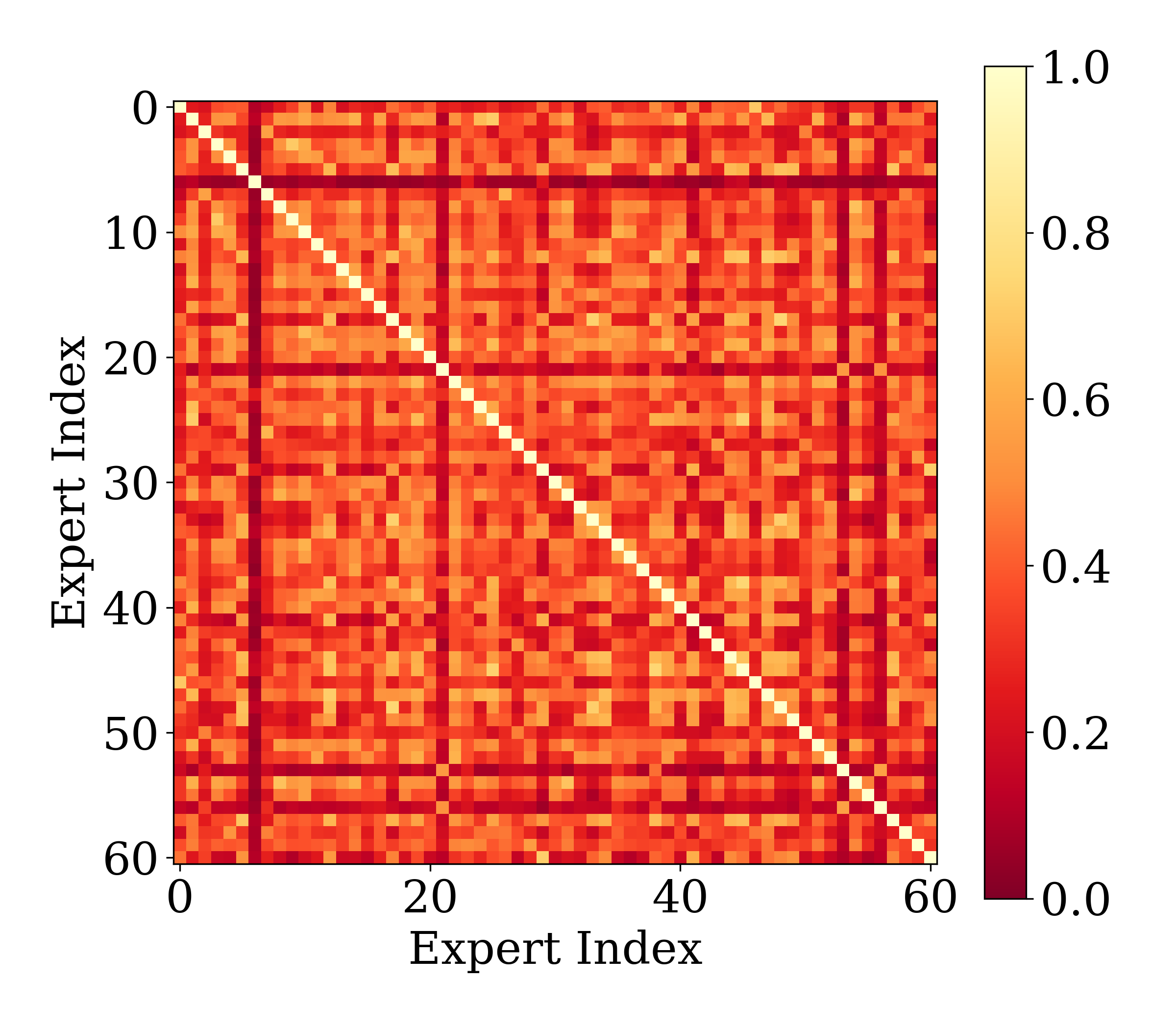}
    }
    ~
    \subfigure[]{
        \includegraphics[width=.22\linewidth]{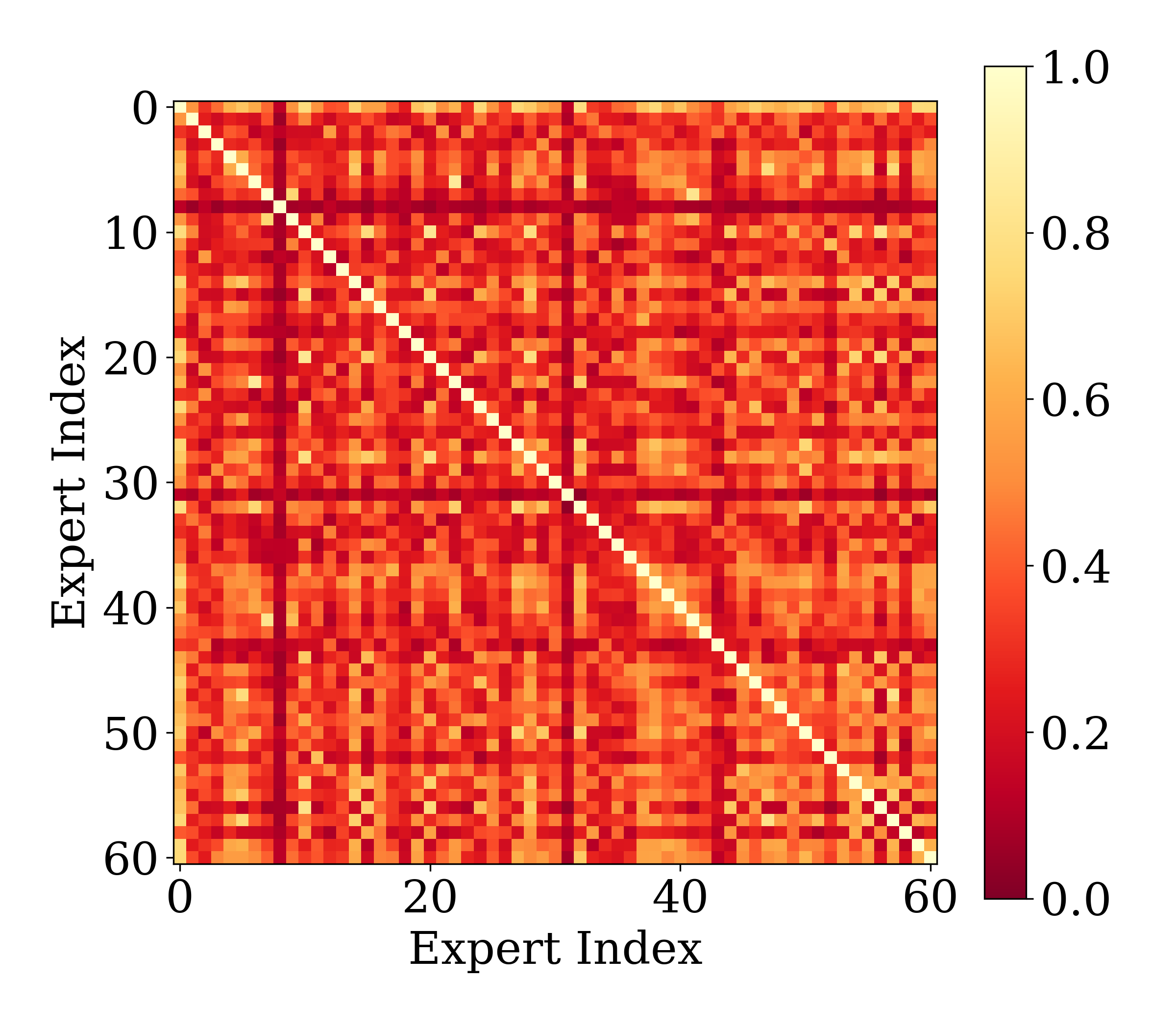}
    }
    \caption{Pairwise expert principal similarity in Qwen~\cite{qwen_moe} models from \textbf{(a)} layer 0, \textbf{(b)} layer 8, \textbf{(c)} layer 16, \textbf{(d)} layer 23. The overlap in expert dominant 1\% spectral subspace is consistent in all model layers.}
    \label{fig:append.all_expert_prin_sim_qwen}
\end{figure*}

\begin{figure*}[h]
    \centering
    \subfigure[]{
        \includegraphics[width=.22\linewidth]{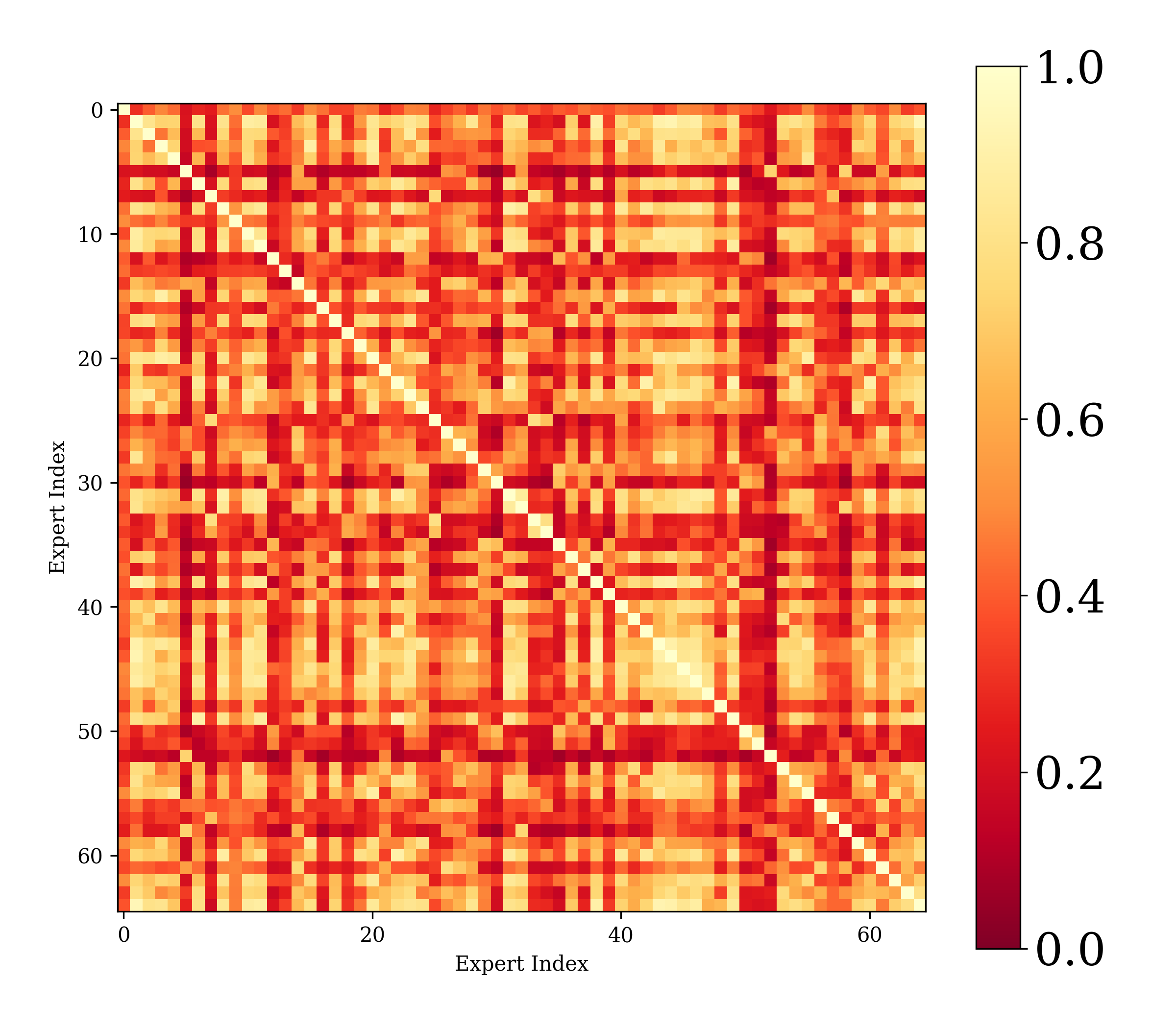}
    }
    ~ % space
    \subfigure[]{
        \includegraphics[width=.22\linewidth]{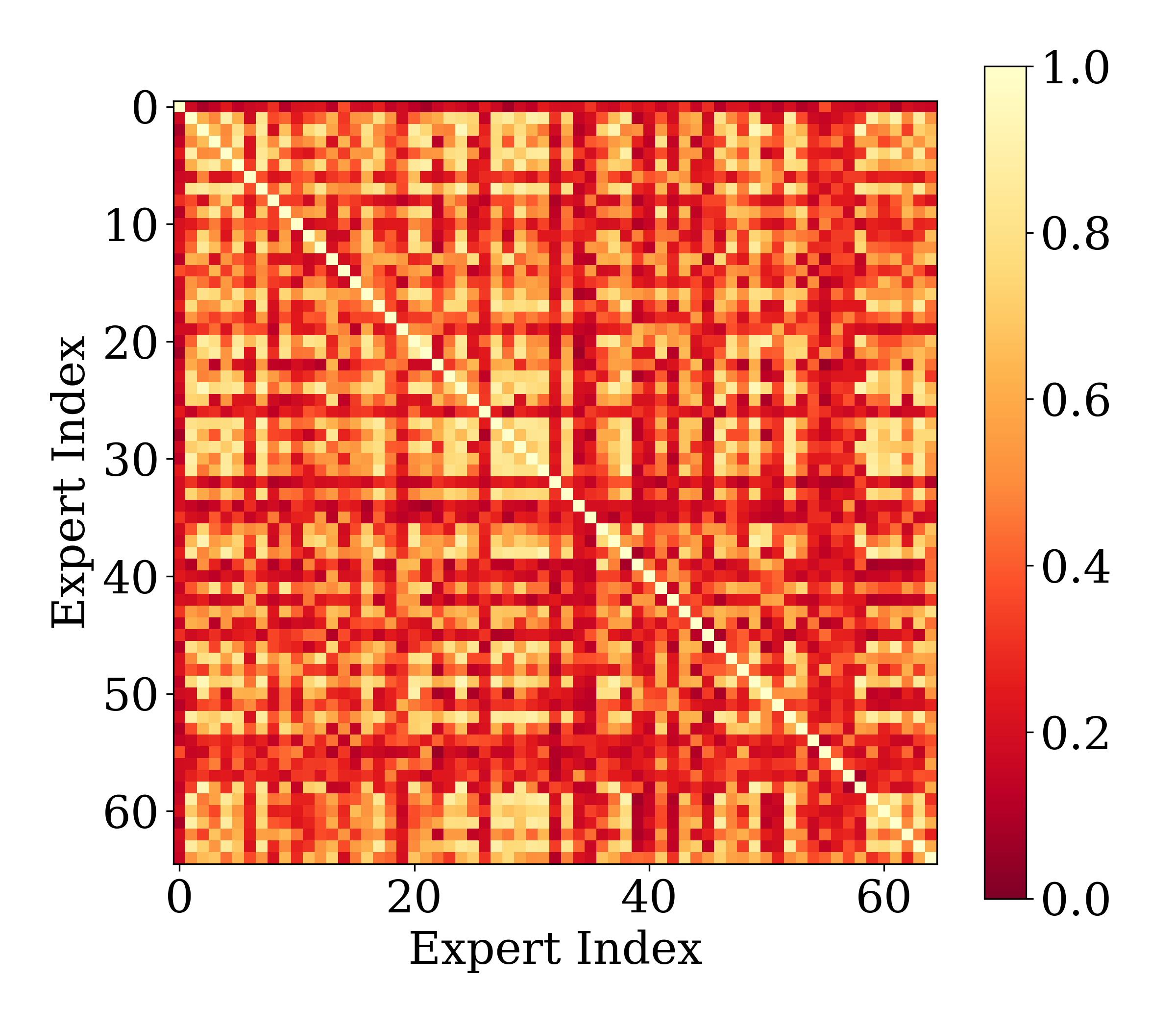}
    }
    ~ % space
    \subfigure[]{
        \includegraphics[width=.22\linewidth]{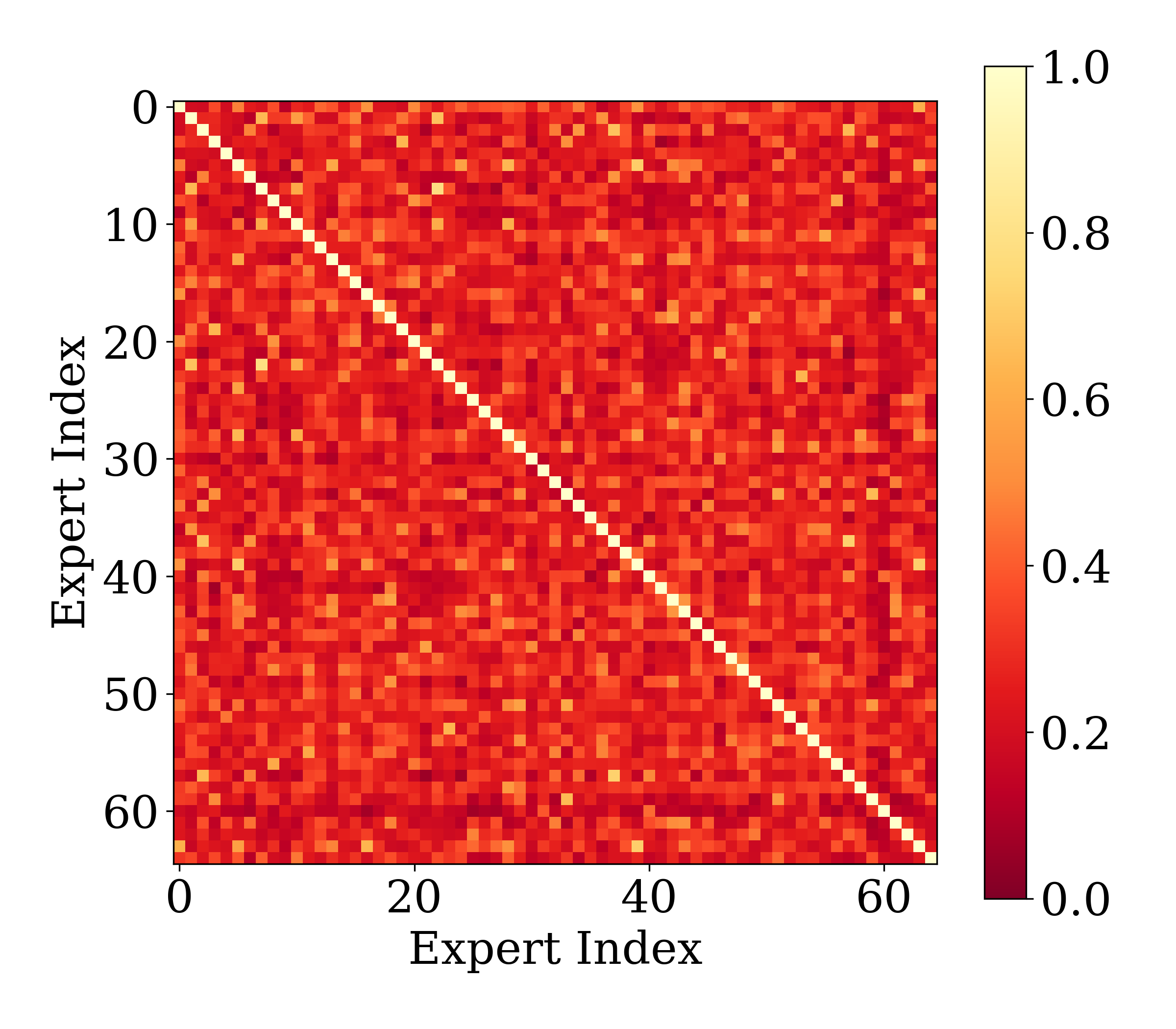}
    }
    ~
    \subfigure[]{
        \includegraphics[width=.22\linewidth]{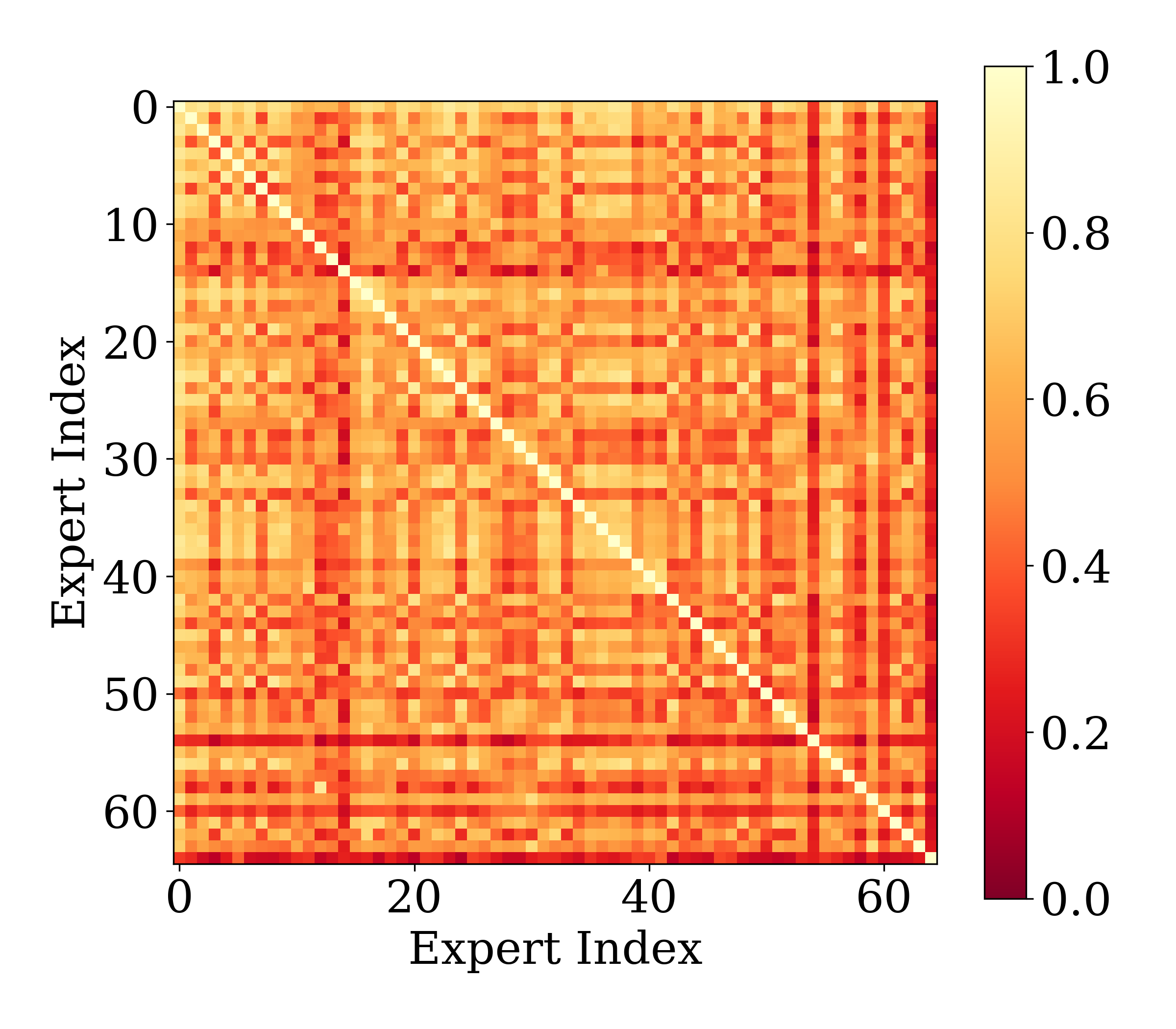}
    }
    \caption{Pairwise expert principal similarity in DeepSeek~\cite{dai2024deepseekmoe} models from \textbf{(a)} layer 1, \textbf{(b)} layer 8, \textbf{(c)} layer 16, \textbf{(d)} layer 26. The overlap in expert dominant 1\% spectral subspace is consistent in all model layers.}
    \label{fig:append.all_expert_prin_sim_ds}
\end{figure*}

\begin{figure*}[h]
    \centering
    \subfigure[]{
        \includegraphics[width=.22\linewidth]{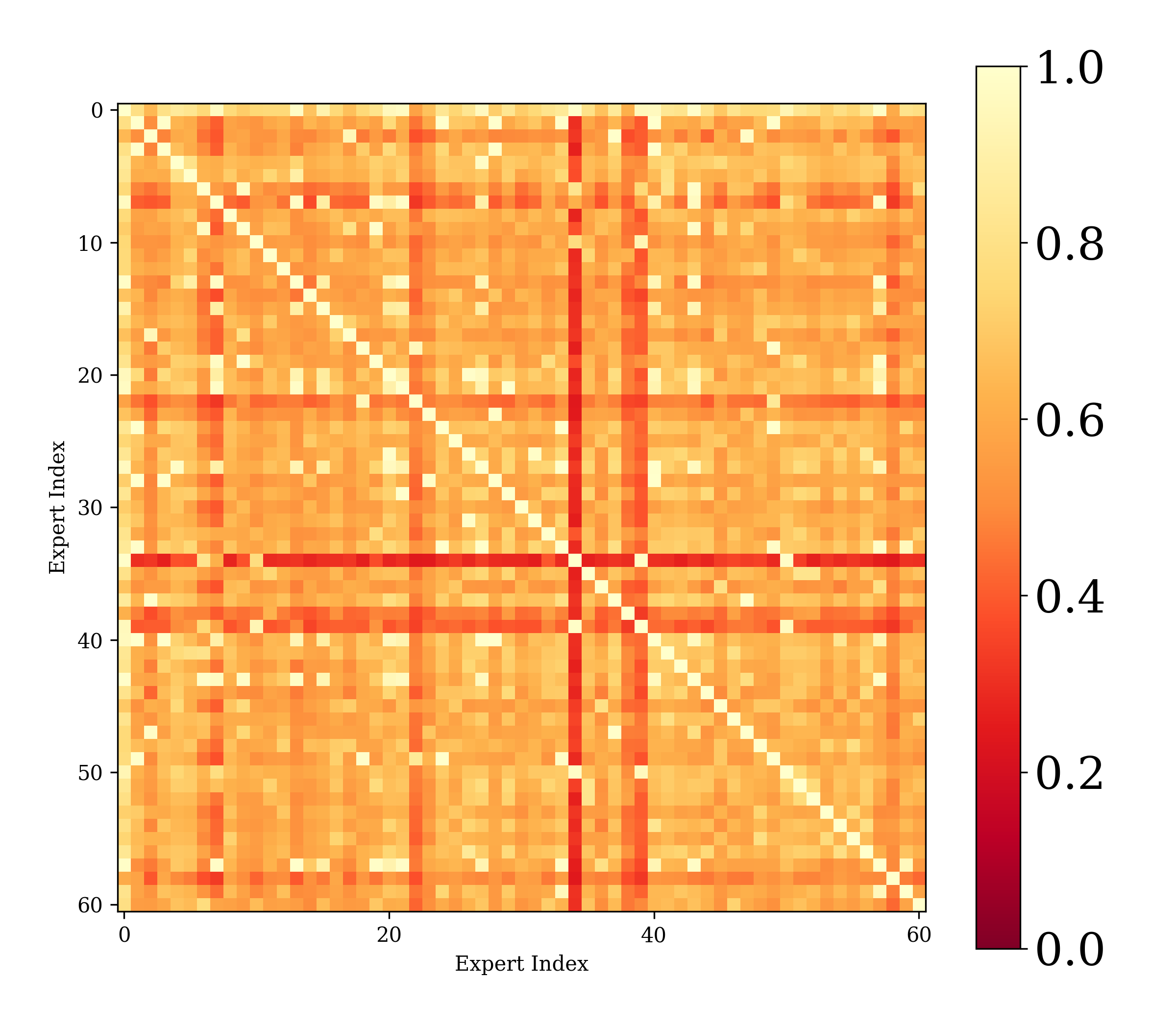}
    }
    ~ % space
    \subfigure[]{
        \includegraphics[width=.22\linewidth]{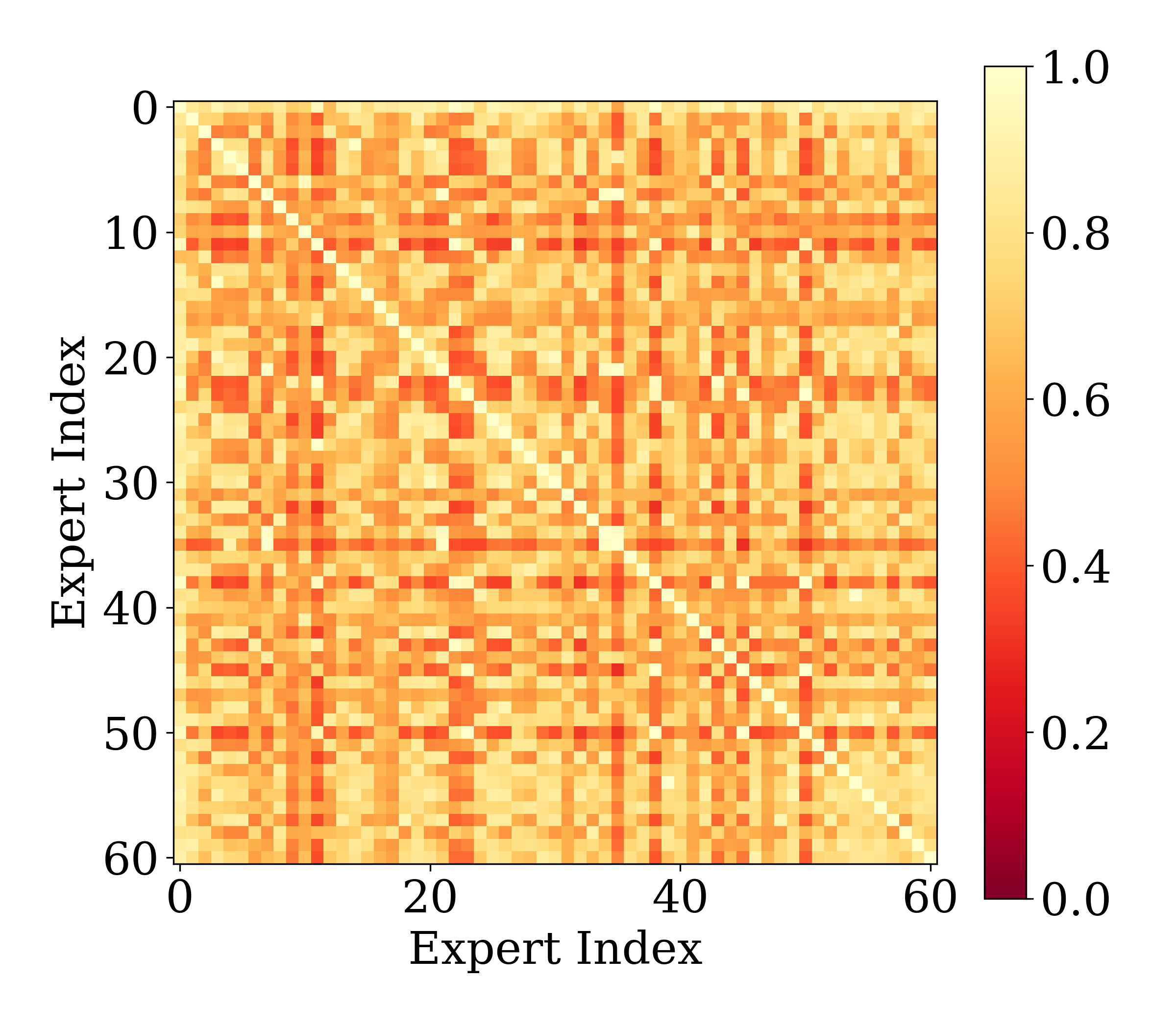}
    }
    ~ % space
    \subfigure[]{
        \includegraphics[width=.22\linewidth]{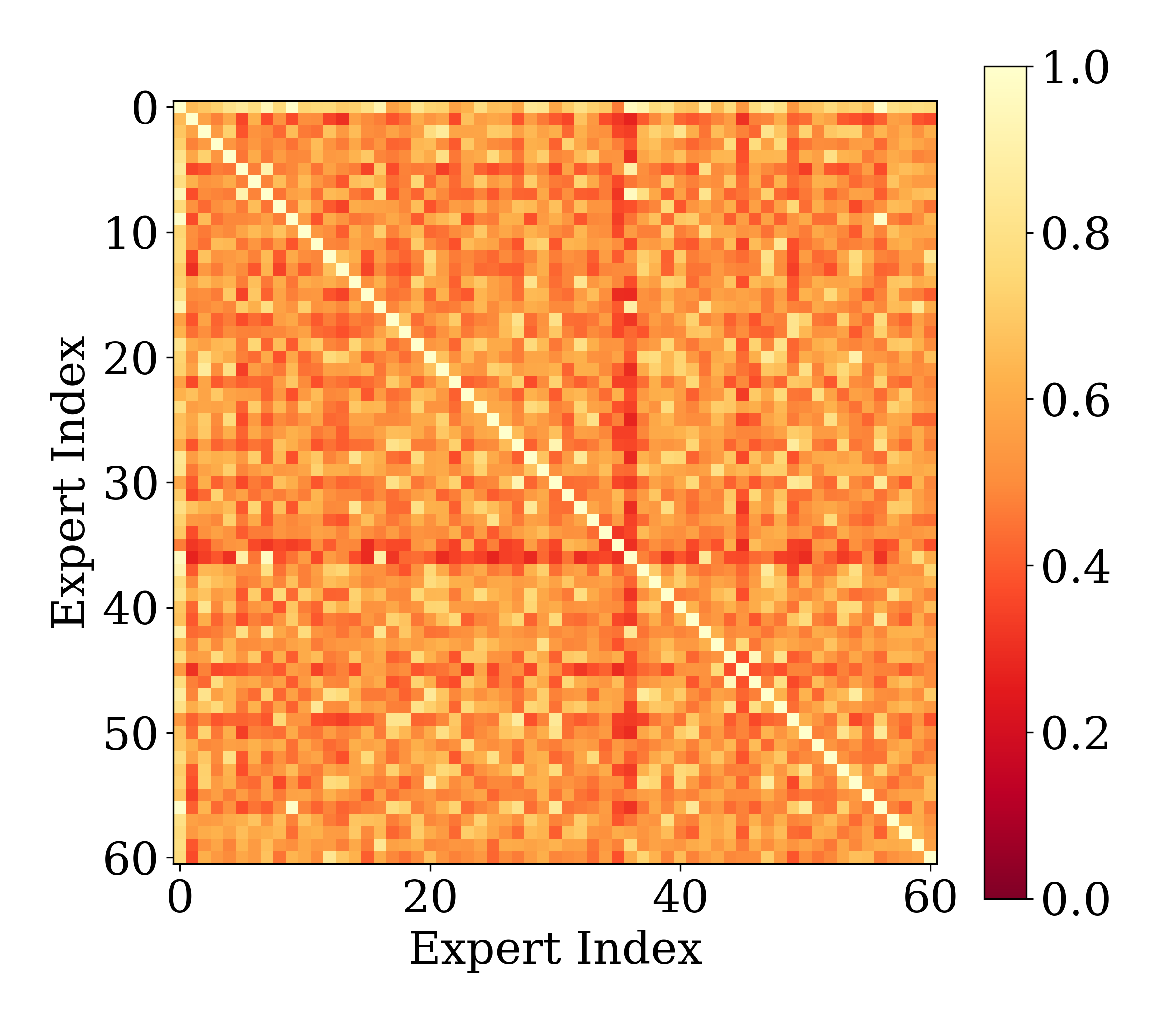}
    }
    ~
    \subfigure[]{
        \includegraphics[width=.22\linewidth]{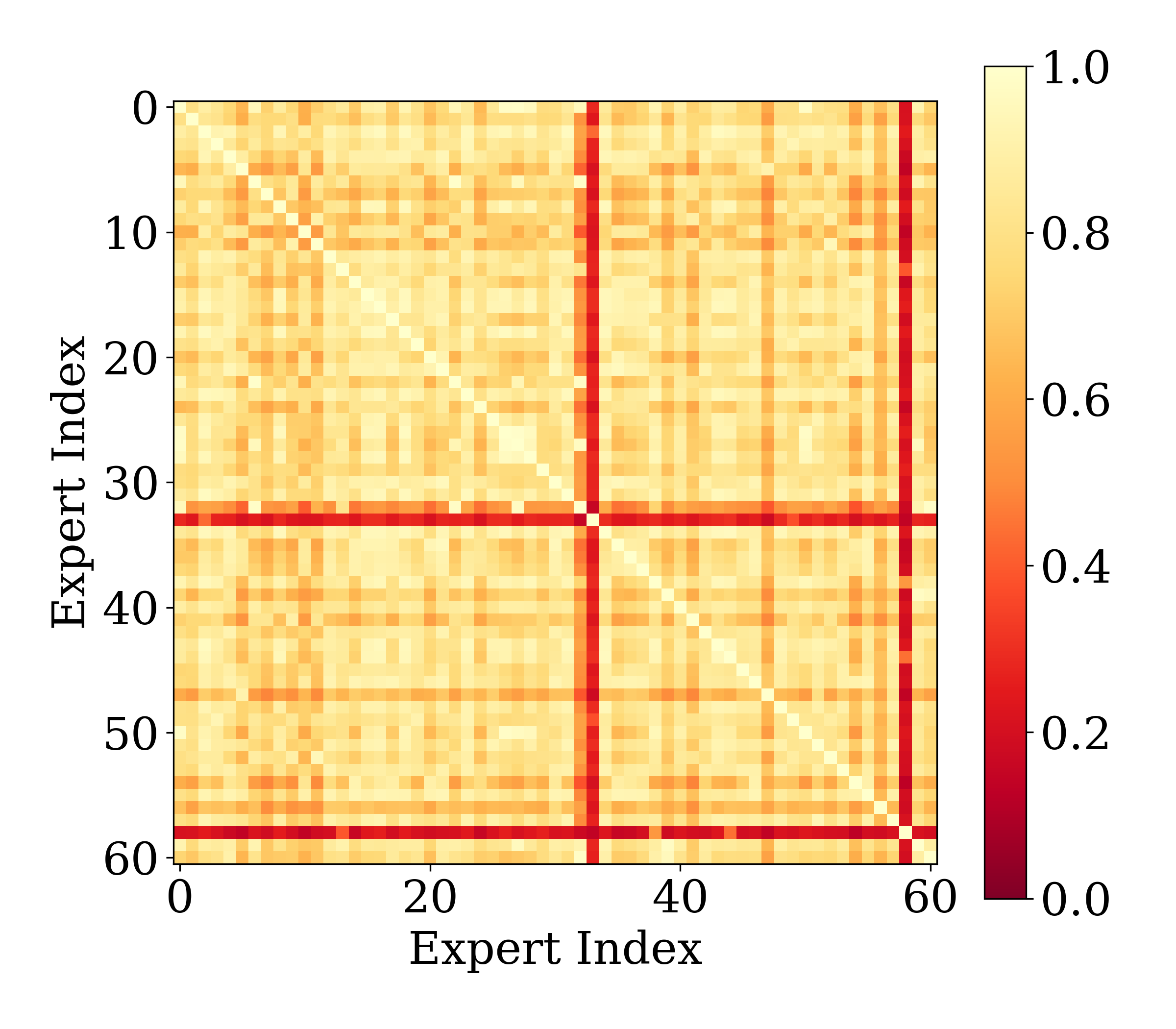}
    }
    \caption{Pairwise expert gradient principal similarity in Qwen~\cite{qwen_moe} models from \textbf{(a)} layer 0, \textbf{(b)} layer 8, \textbf{(c)} layer 16, \textbf{(d)} layer 23. The overlap in expert gradient dominant 1\% spectral subspace is consistent in all model layers.}
    \label{fig:append.all_expert_grad_prin_sim}
\end{figure*}

\begin{figure*}[h]
    \centering
    \subfigure[]{
        \includegraphics[width=.22\linewidth]{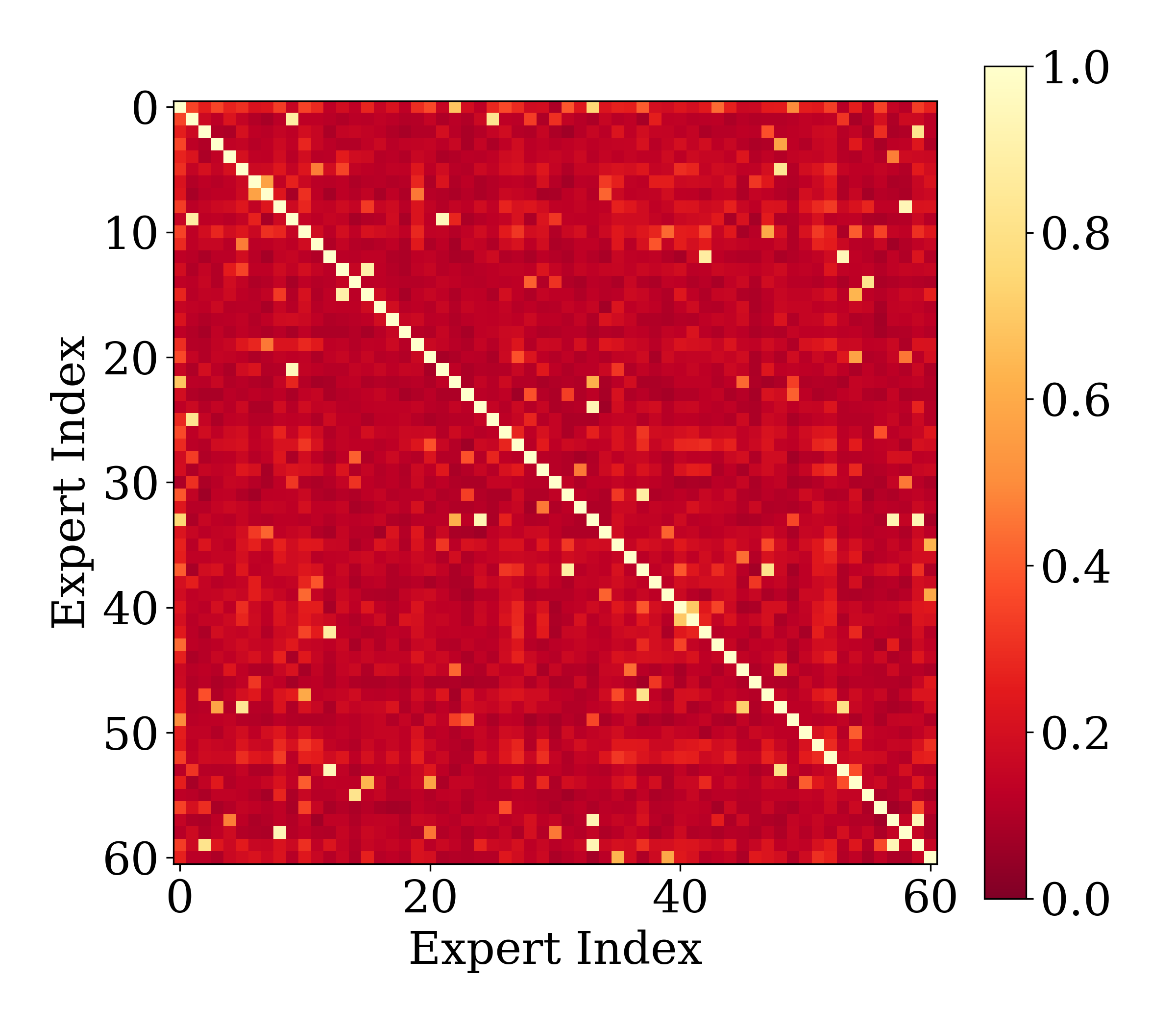}
    }
    ~ % space
    \subfigure[]{
        \includegraphics[width=.22\linewidth]{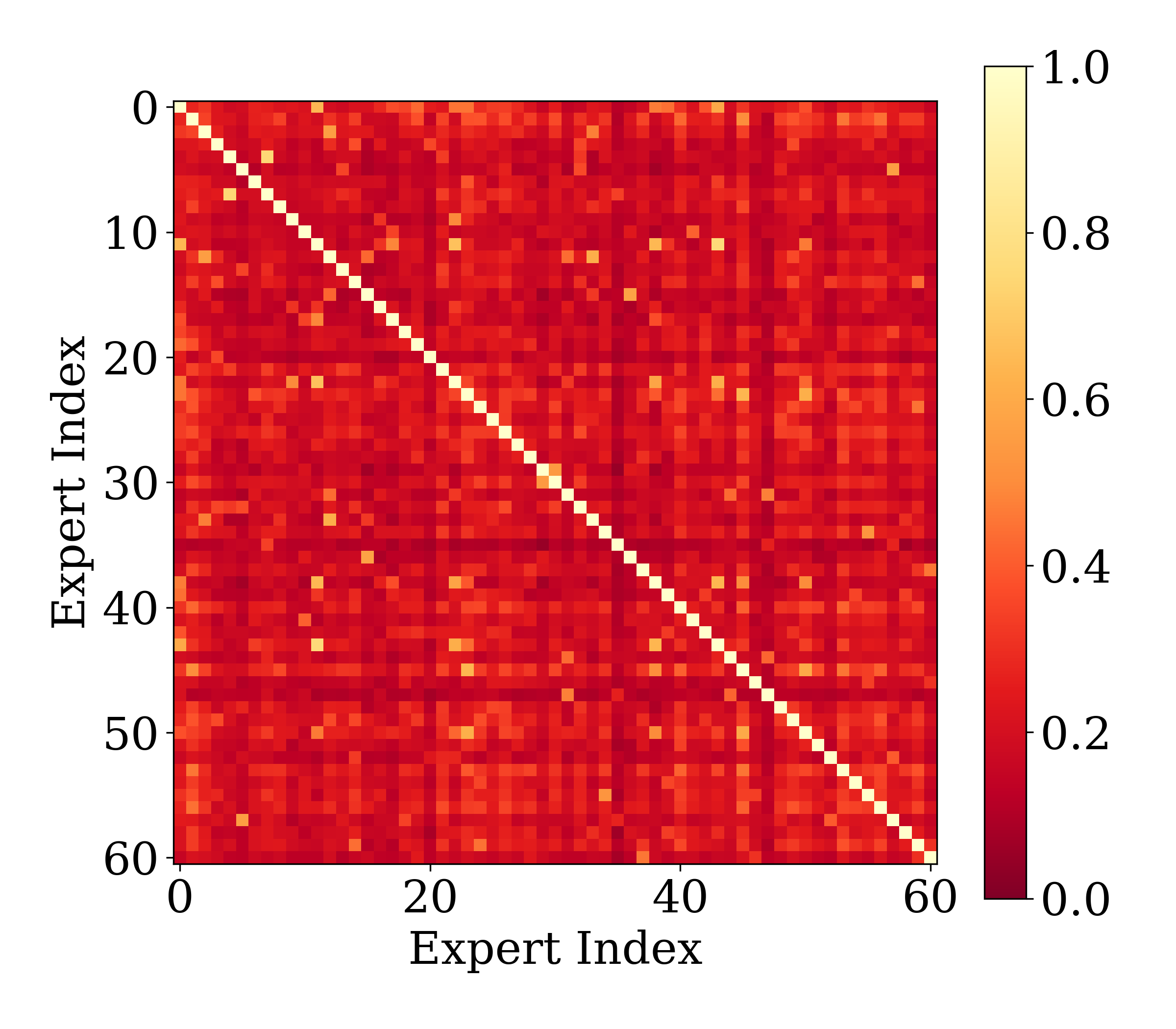}
    }
    ~ % space
    \subfigure[]{
        \includegraphics[width=.22\linewidth]{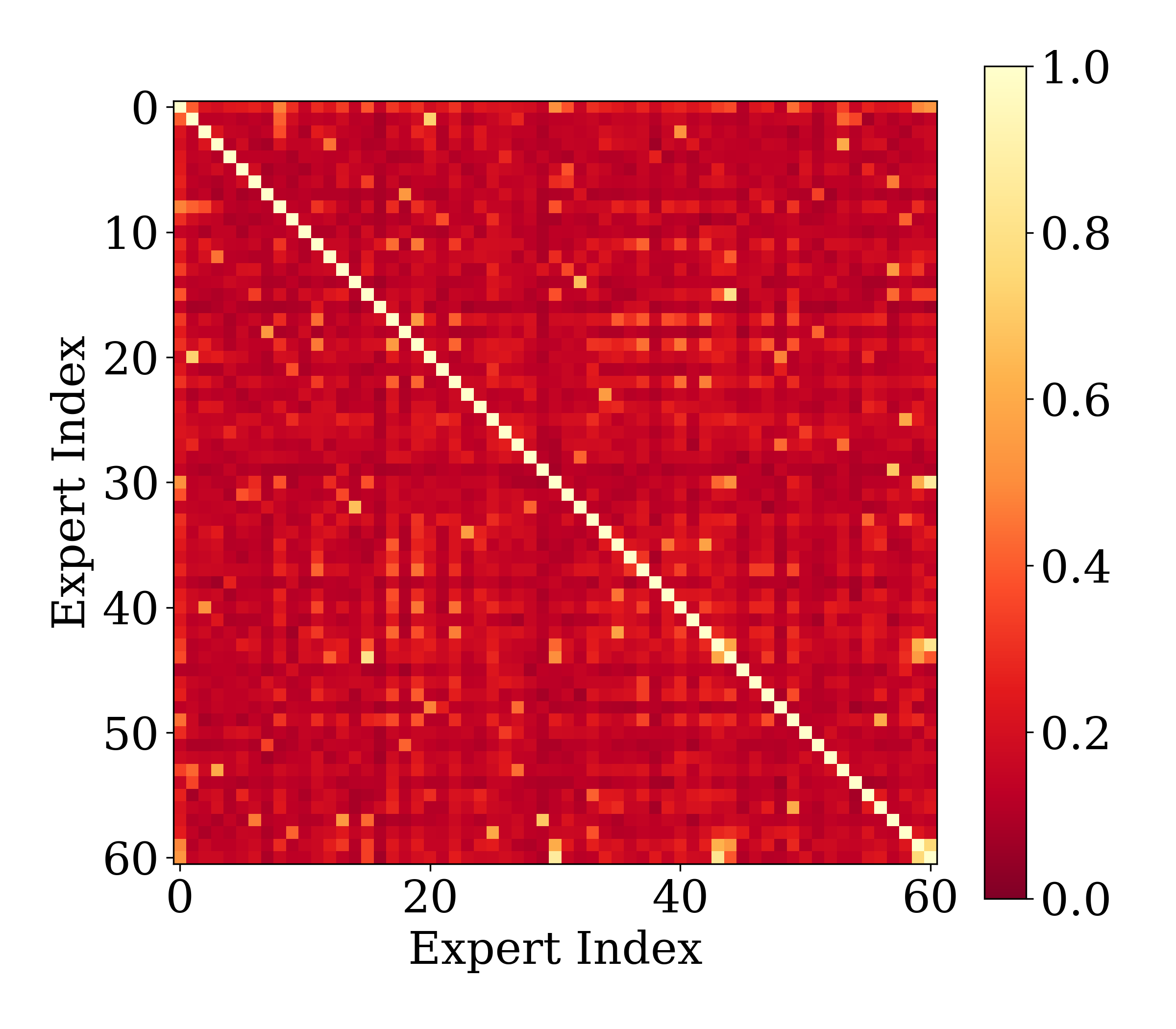}
    }
    ~
    \subfigure[]{
        \includegraphics[width=.22\linewidth]{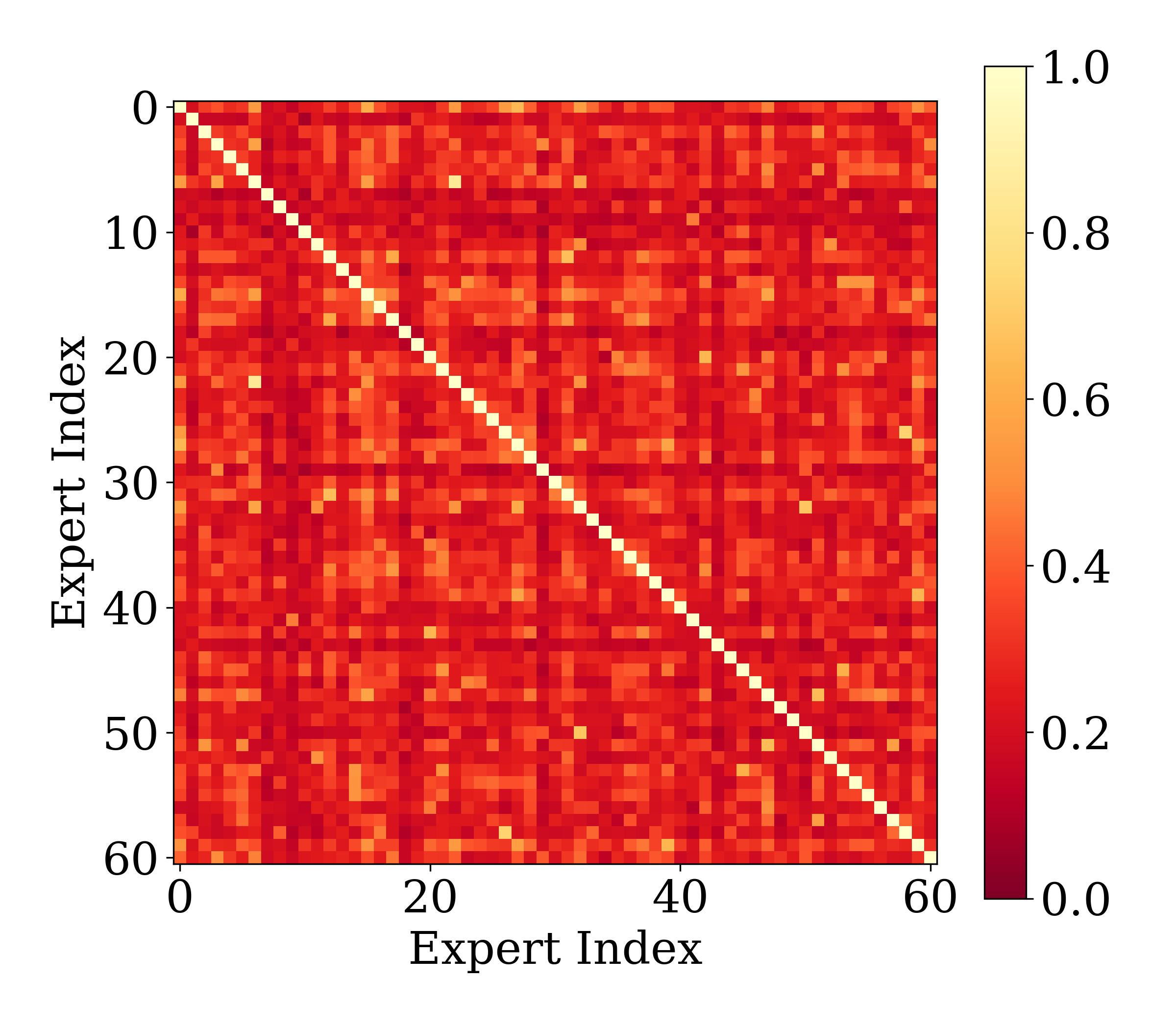}
    }
    \caption{Pairwise expert gradient long-tail subspace principal similarity in Qwen~\cite{qwen_moe} models from \textbf{(a)} layer 0, \textbf{(b)} layer 8, \textbf{(c)} layer 16, \textbf{(d)} layer 23. The overlap in expert gradient long-tail spectral subspace is consistent in all model layers.}
    \label{fig:append.all_expert_grad_tail_sim}
\end{figure*}

\begin{figure*}[h]
    \centering
    \subfigure[]{
        \includegraphics[width=.22\linewidth]{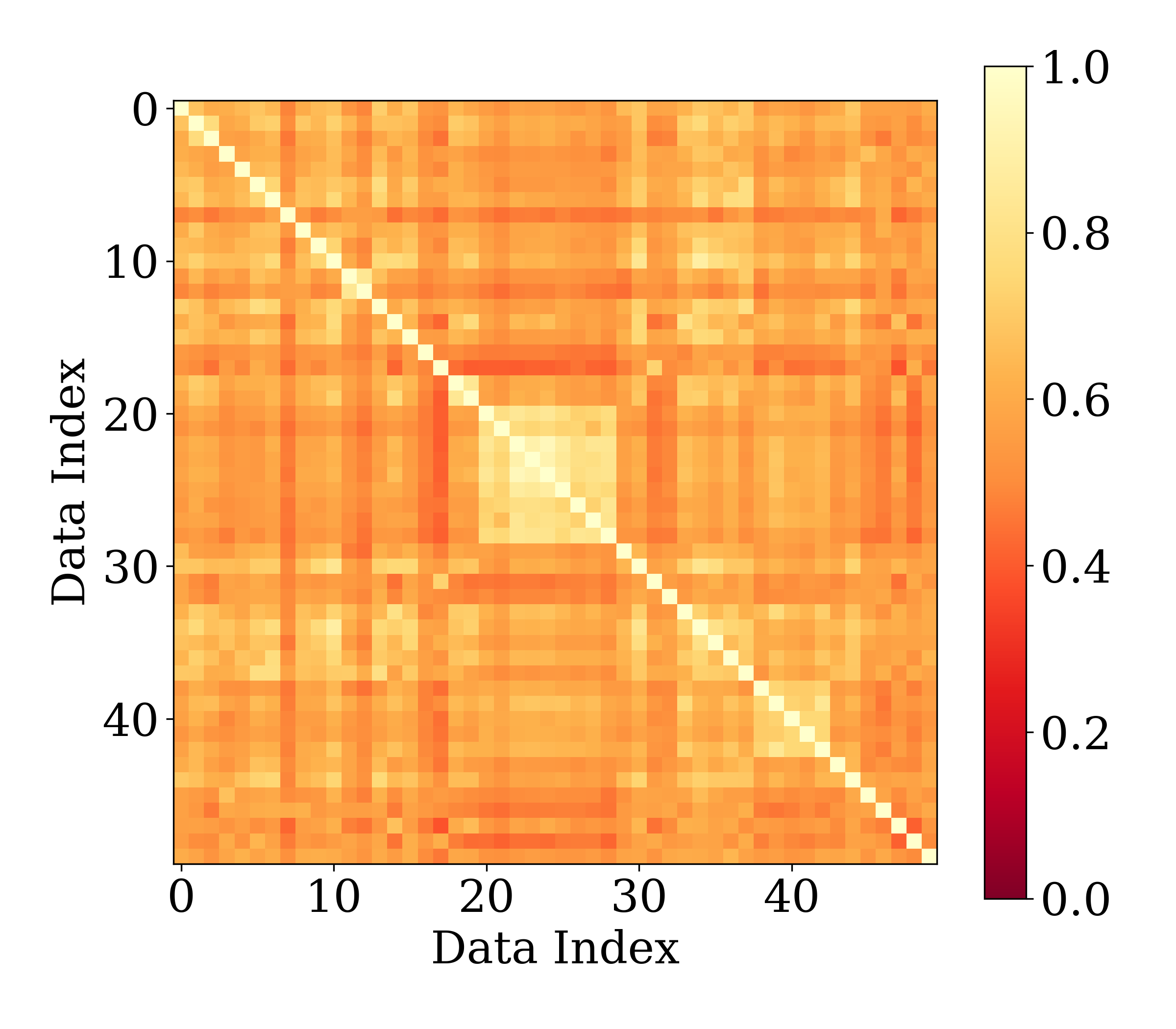}
    }
    ~ % space
    \subfigure[]{
        \includegraphics[width=.22\linewidth]{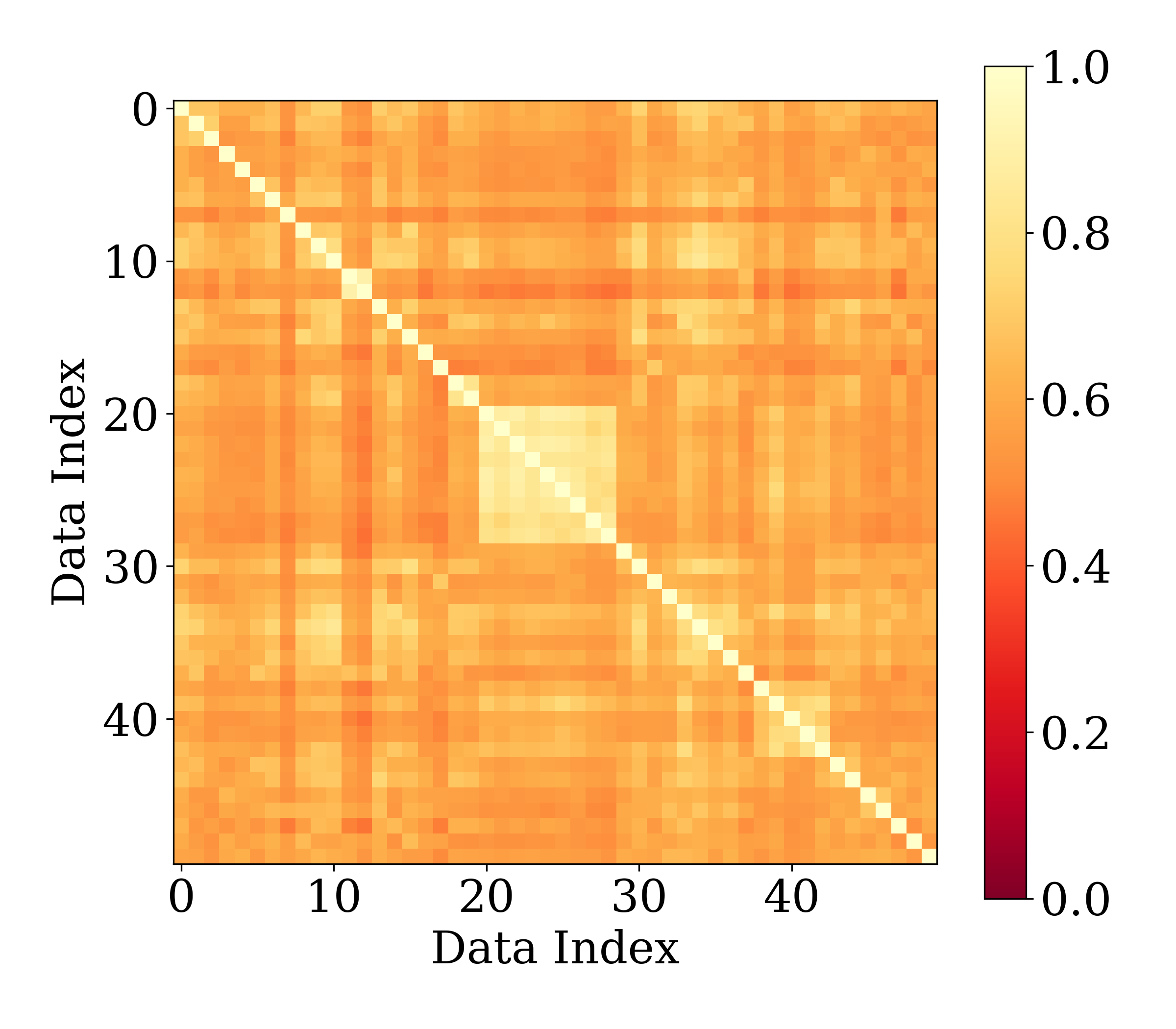}
    }
    ~ % space
    \subfigure[]{
        \includegraphics[width=.22\linewidth]{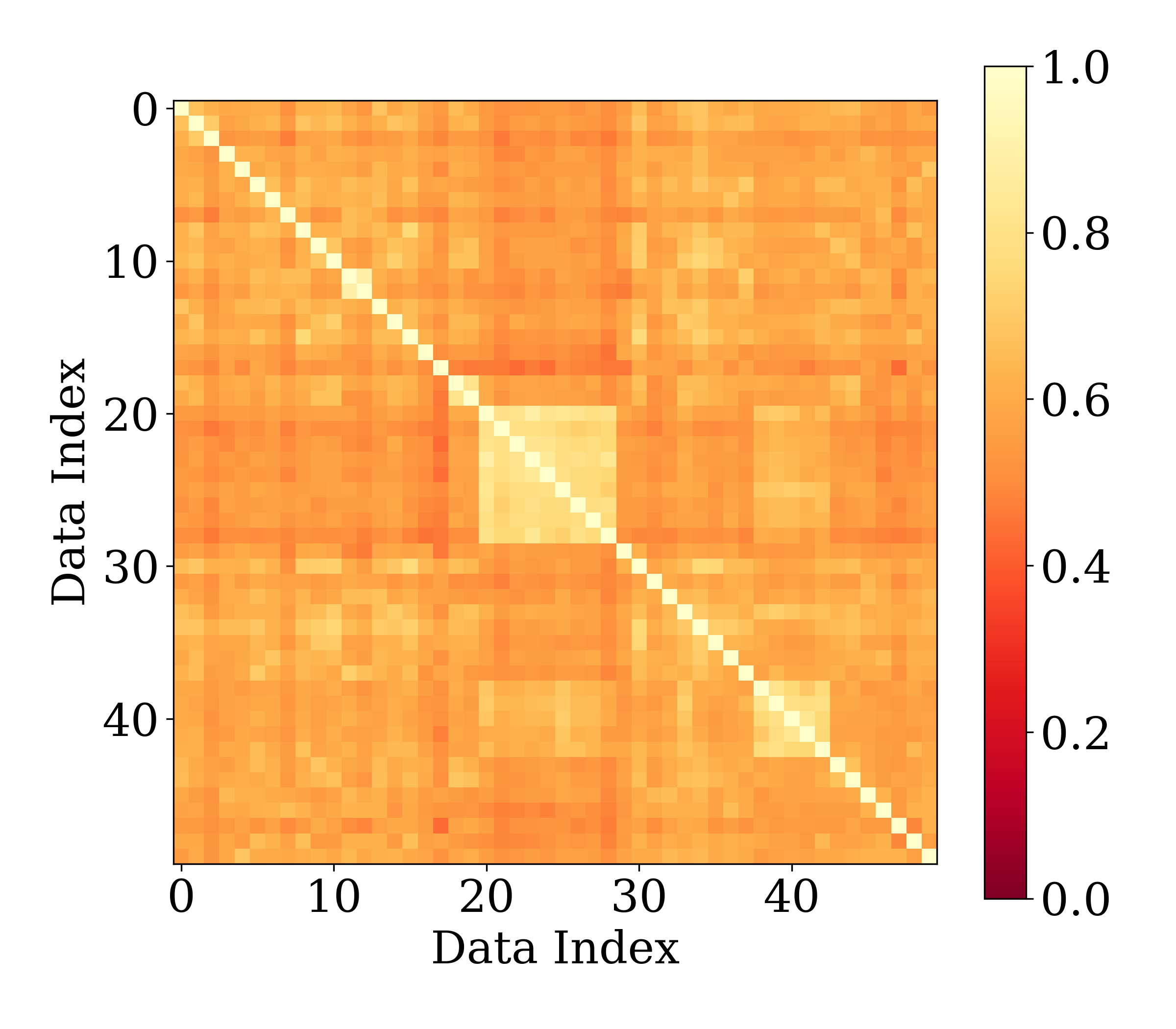}
    }
    ~
    \subfigure[]{
        \includegraphics[width=.22\linewidth]{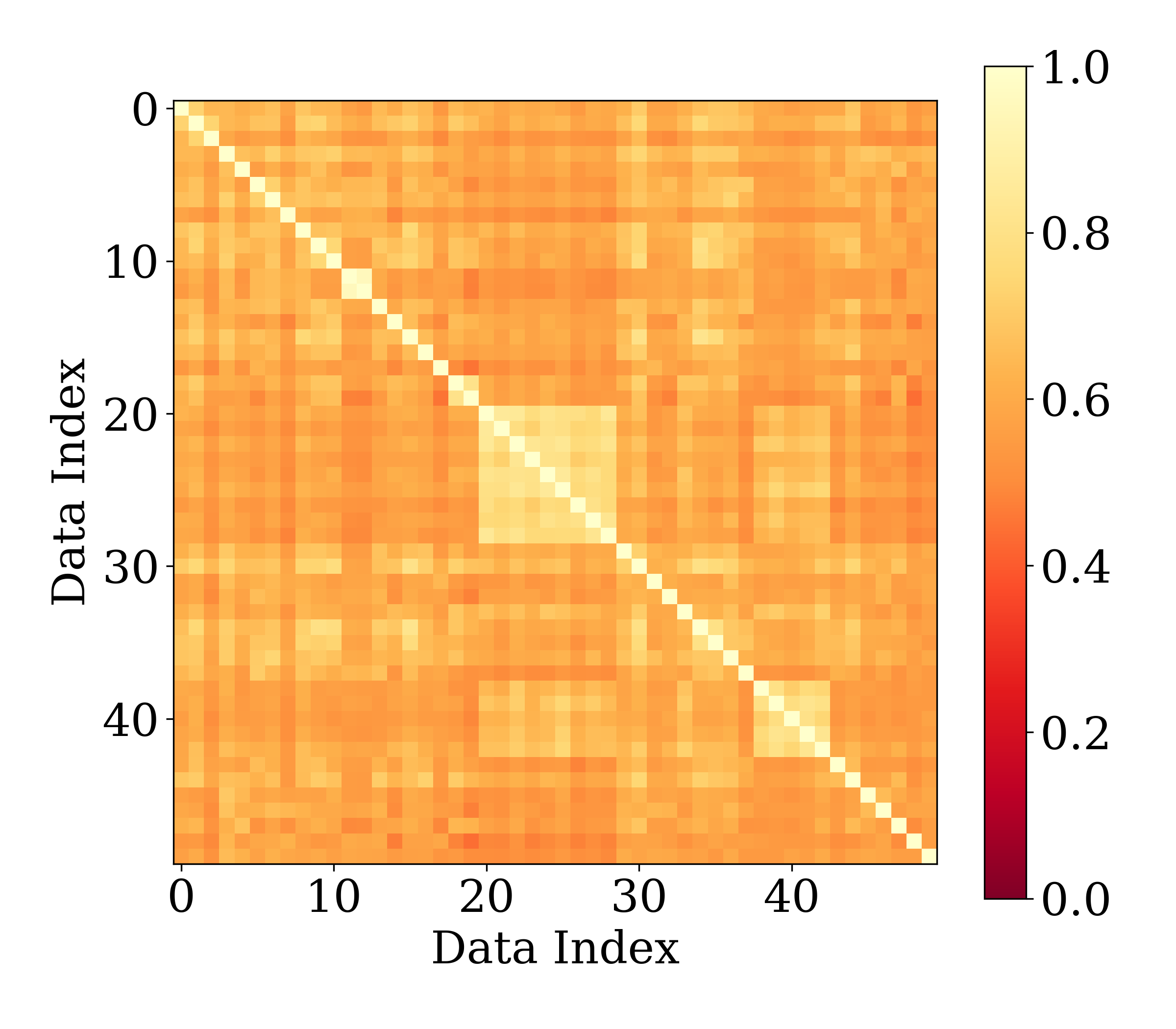}
    }
    \caption{Data activation principal subspace principal similarity in Qwen~\cite{qwen_moe} models from \textbf{(a)} layer 0, \textbf{(b)} layer 8, \textbf{(c)} layer 16, \textbf{(d)} layer 23. The overlap in data activation subspace is consistent in all model layers.}
    \label{fig:append.all_data_act_alignment}
\end{figure*}

\begin{figure*}[h]
    \centering
    \subfigure[]{
        \includegraphics[width=.22\linewidth]{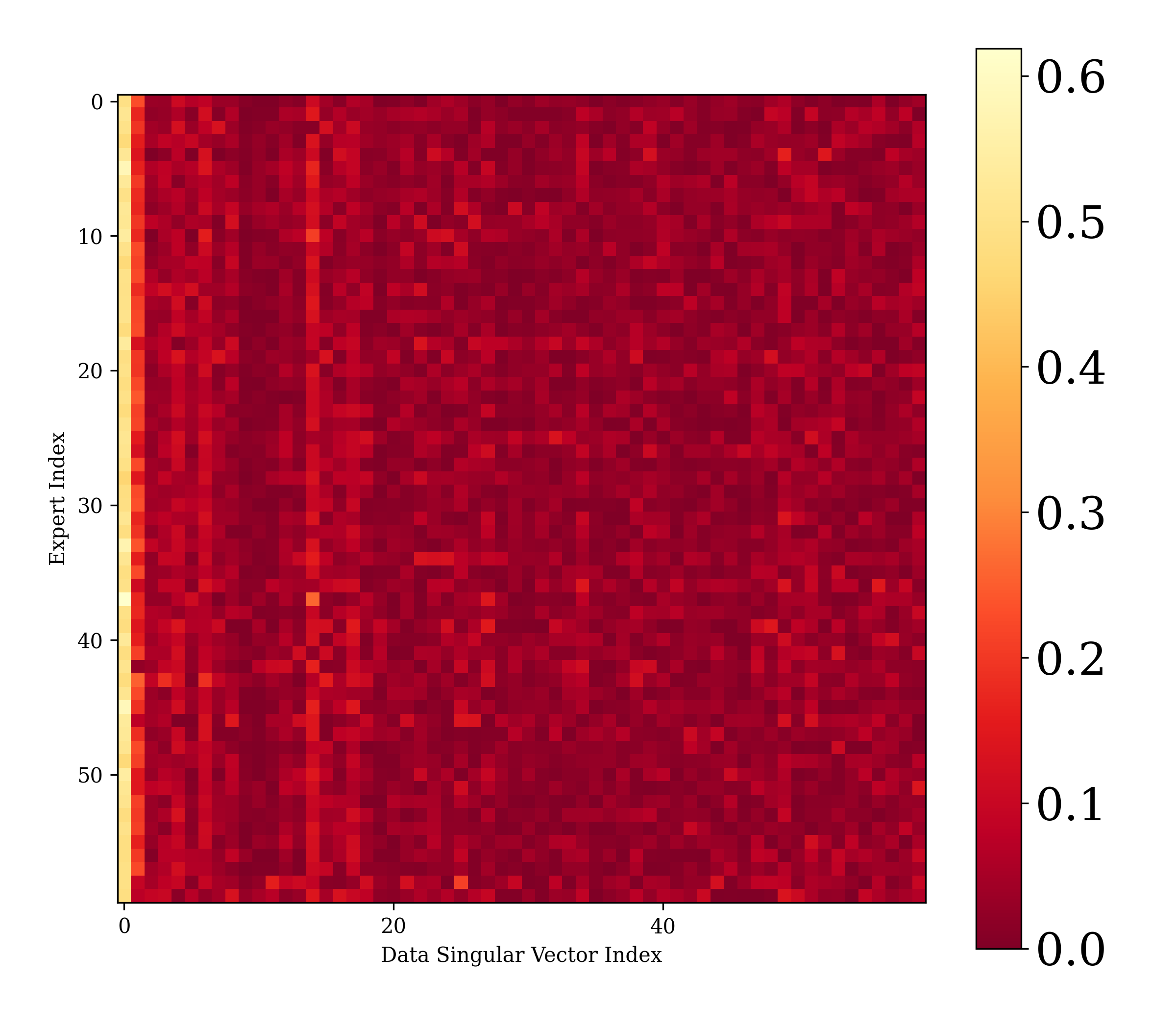}
    }
    ~ % space
    \subfigure[]{
        \includegraphics[width=.22\linewidth]{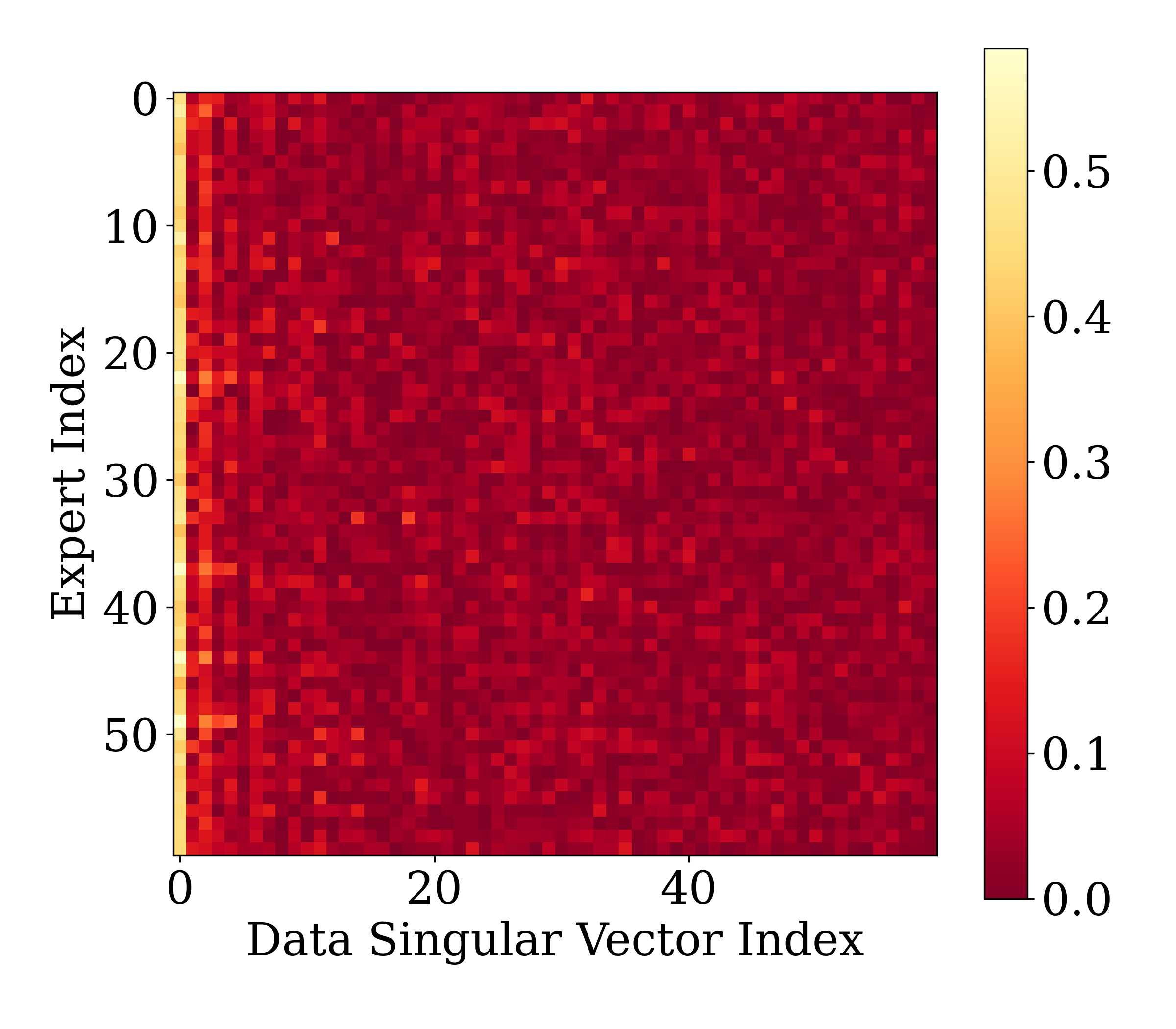}
    }
    ~ % space
    \subfigure[]{
        \includegraphics[width=.22\linewidth]{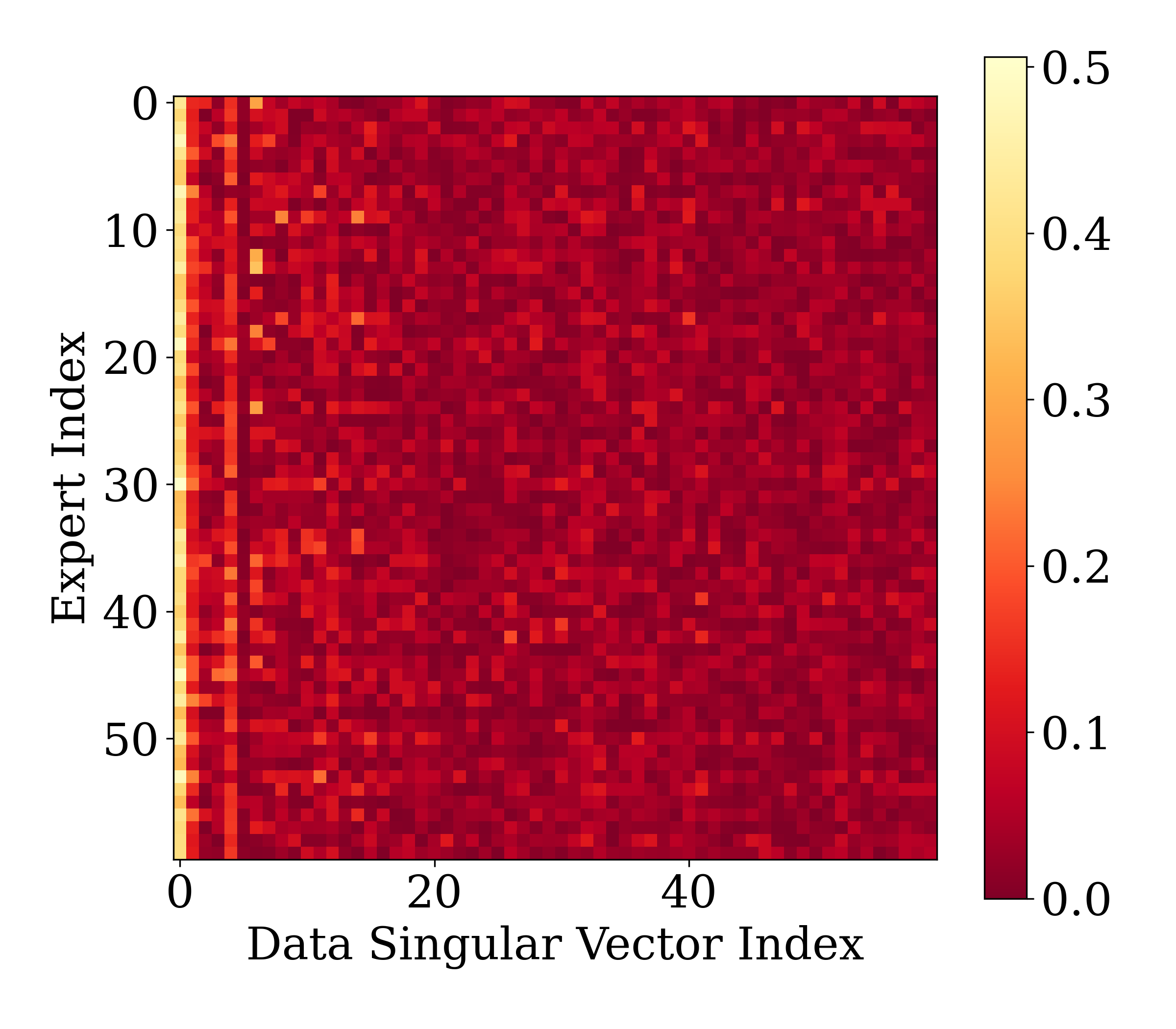}
    }
    ~
    \subfigure[]{
        \includegraphics[width=.22\linewidth]{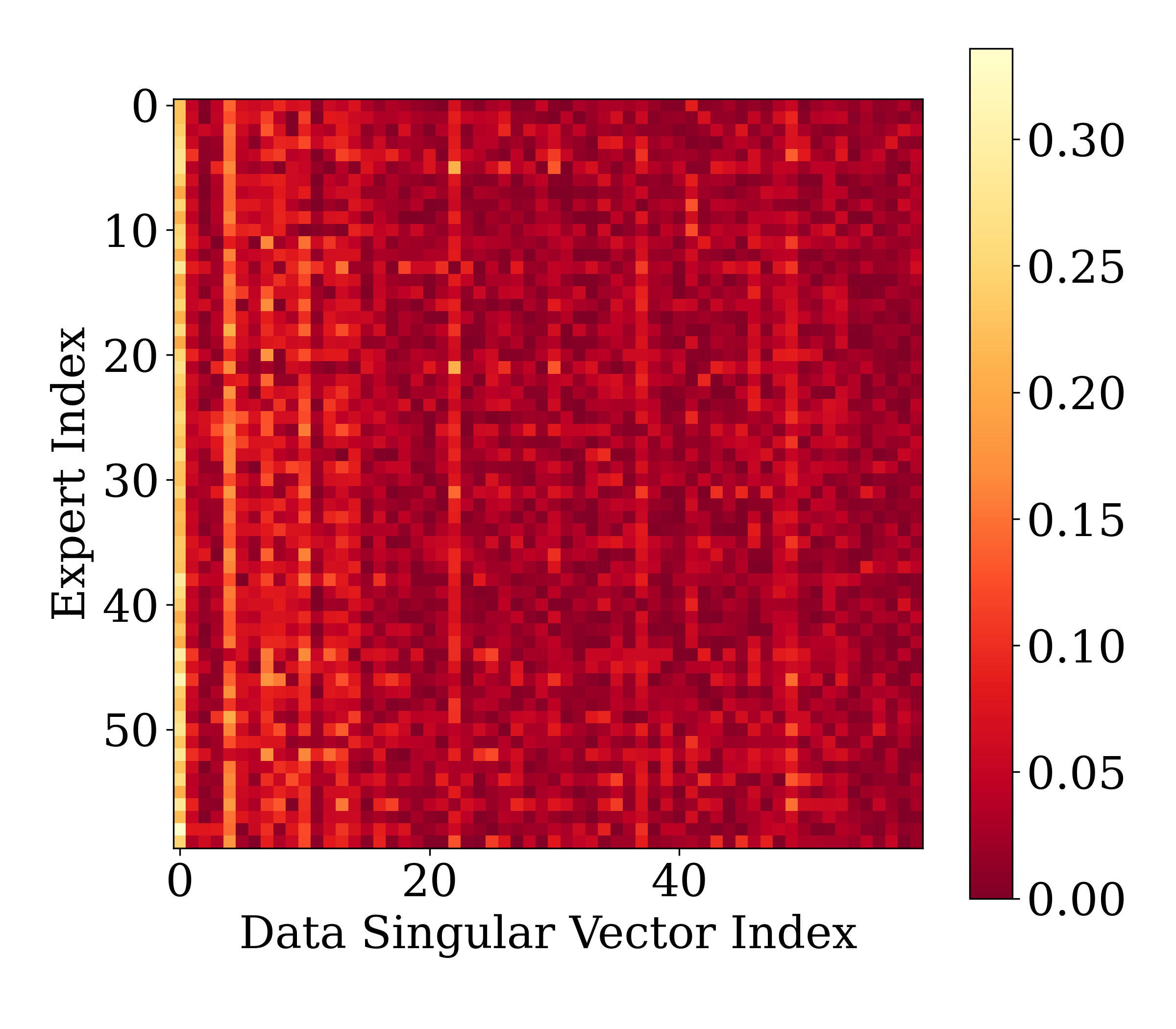}
    }
    \caption{Gate vector and data activation alignment in Qwen~\cite{qwen_moe} models from \textbf{(a)} layer 0, \textbf{(b)} layer 8, \textbf{(c)} layer 16, \textbf{(d)} layer 23. The gate vector is primarily aligned with the leading singular vectors of activation matrix in all model layers.}
    \label{fig:append.all_gate_act_alignment}
\end{figure*}

\begin{figure*}[h]
    \centering
    \subfigure[]{
        \includegraphics[width=.22\linewidth]{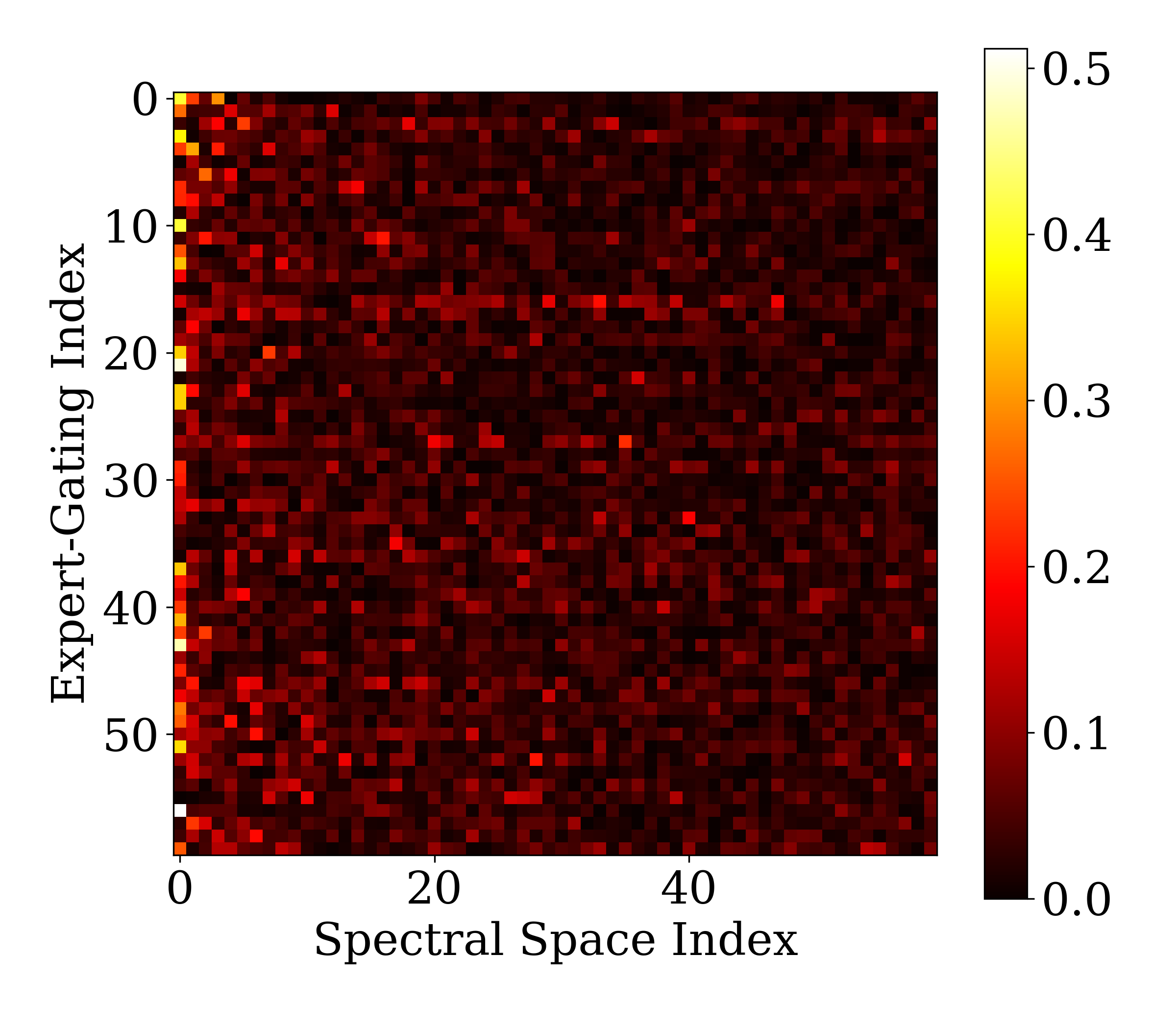}
    }
    ~ % space
    \subfigure[]{
        \includegraphics[width=.22\linewidth]{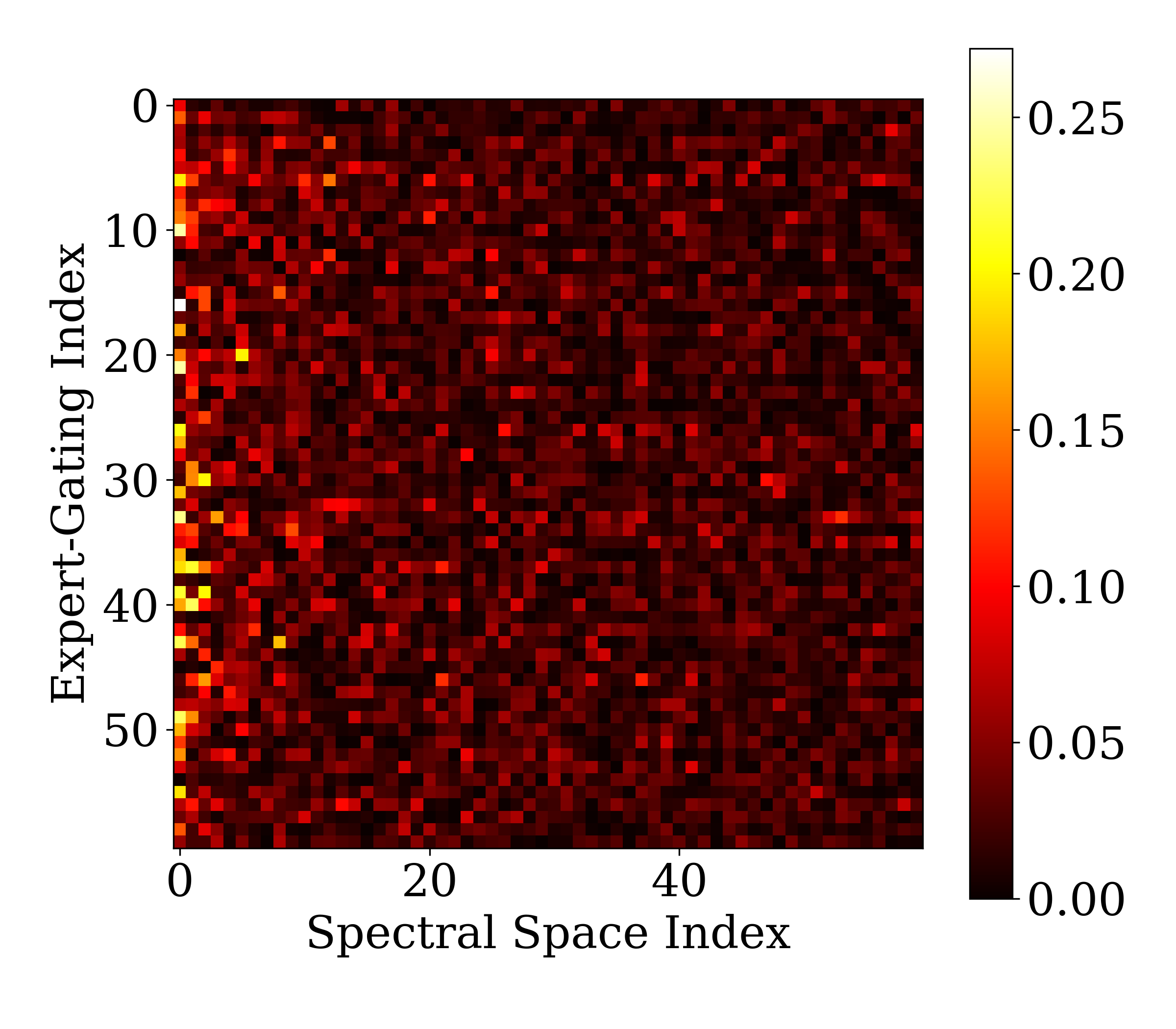}
    }
    ~ % space
    \subfigure[]{
        \includegraphics[width=.22\linewidth]{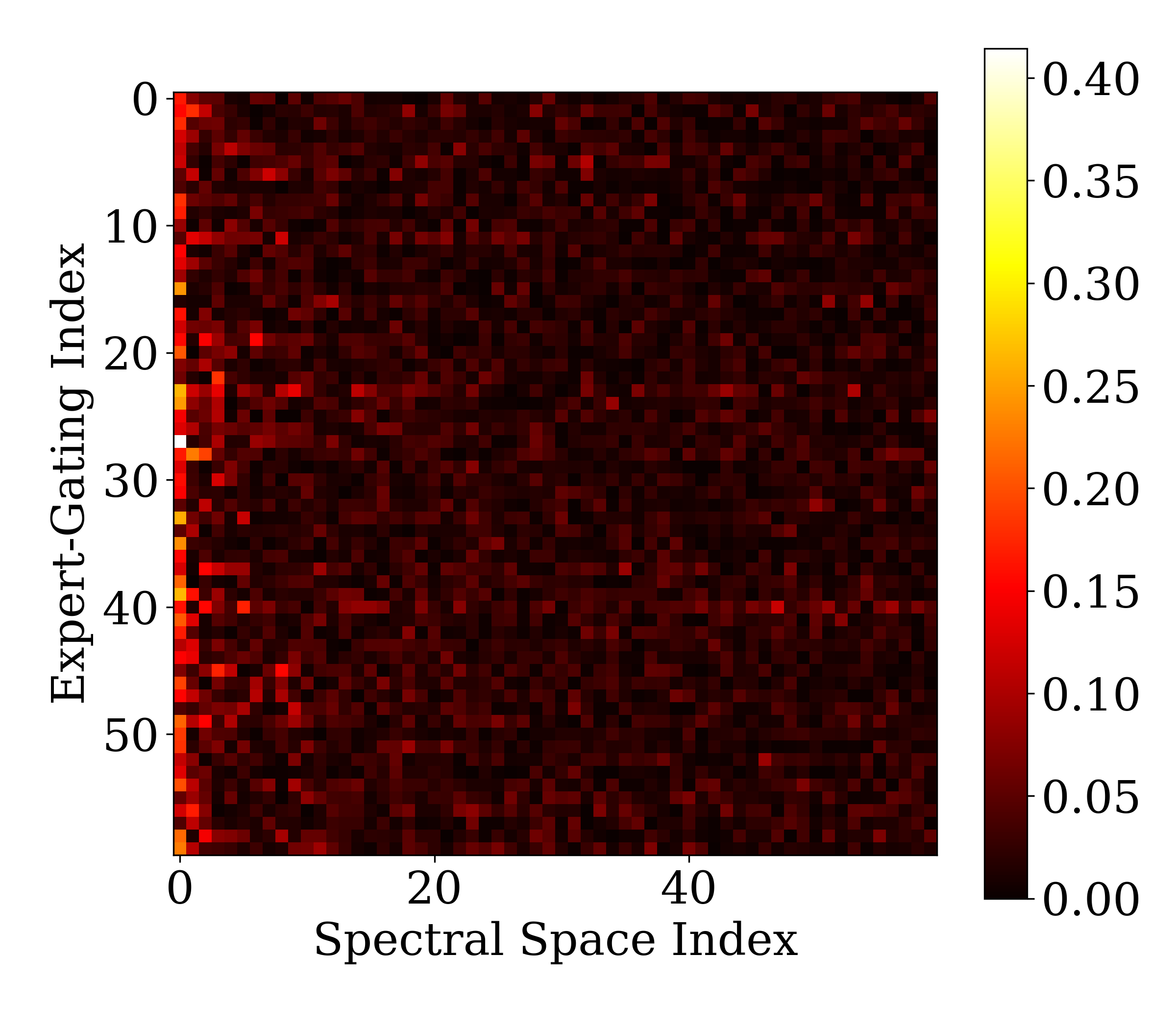}
    }
    ~
    \subfigure[]{
        \includegraphics[width=.22\linewidth]{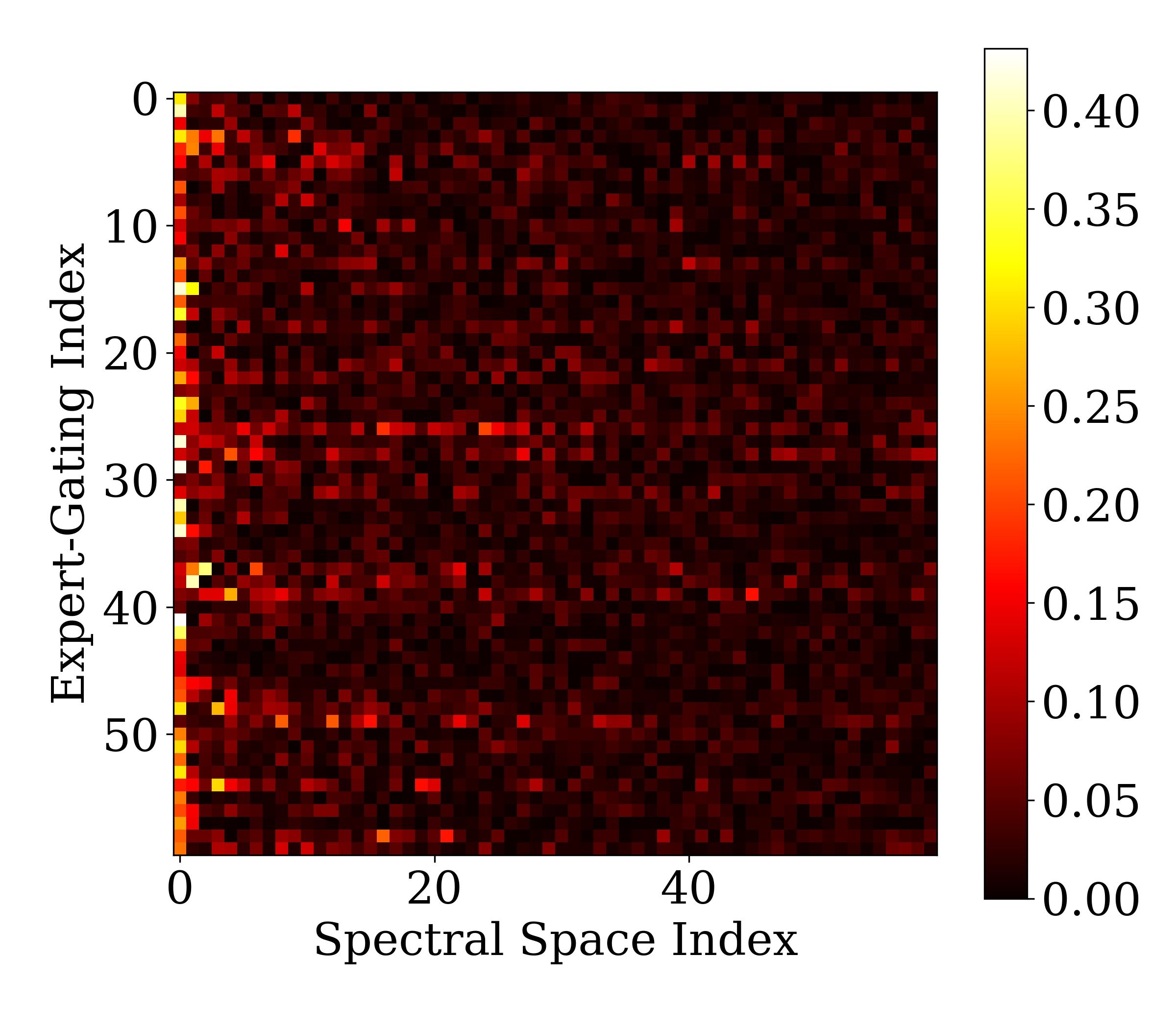}
    }
    \caption{Gate vector and corresponding expert parameter singular vectors alignment in Qwen~\cite{qwen_moe} models from \textbf{(a)} layer 0, \textbf{(b)} layer 8, \textbf{(c)} layer 16, \textbf{(d)} layer 23. The gate vector is primarily aligned with the leading singular vectors.}
    \label{fig:append.all_gate_param_spectral_alignment_qwen}
\end{figure*}

Supplementary result for Section~\ref{sec:analysis} are listed below in Figure~\ref{fig:append.all_expert_sigvals} to Figure~\ref{fig:append.all_gate_param_spectral_alignment_qwen}. All claims or observations we mentioned in Section~\ref{sec:analysis} are consistent in all model layers.

\begin{figure*}[h]
    \centering
    \subfigure[]{
        \includegraphics[width=.22\linewidth]{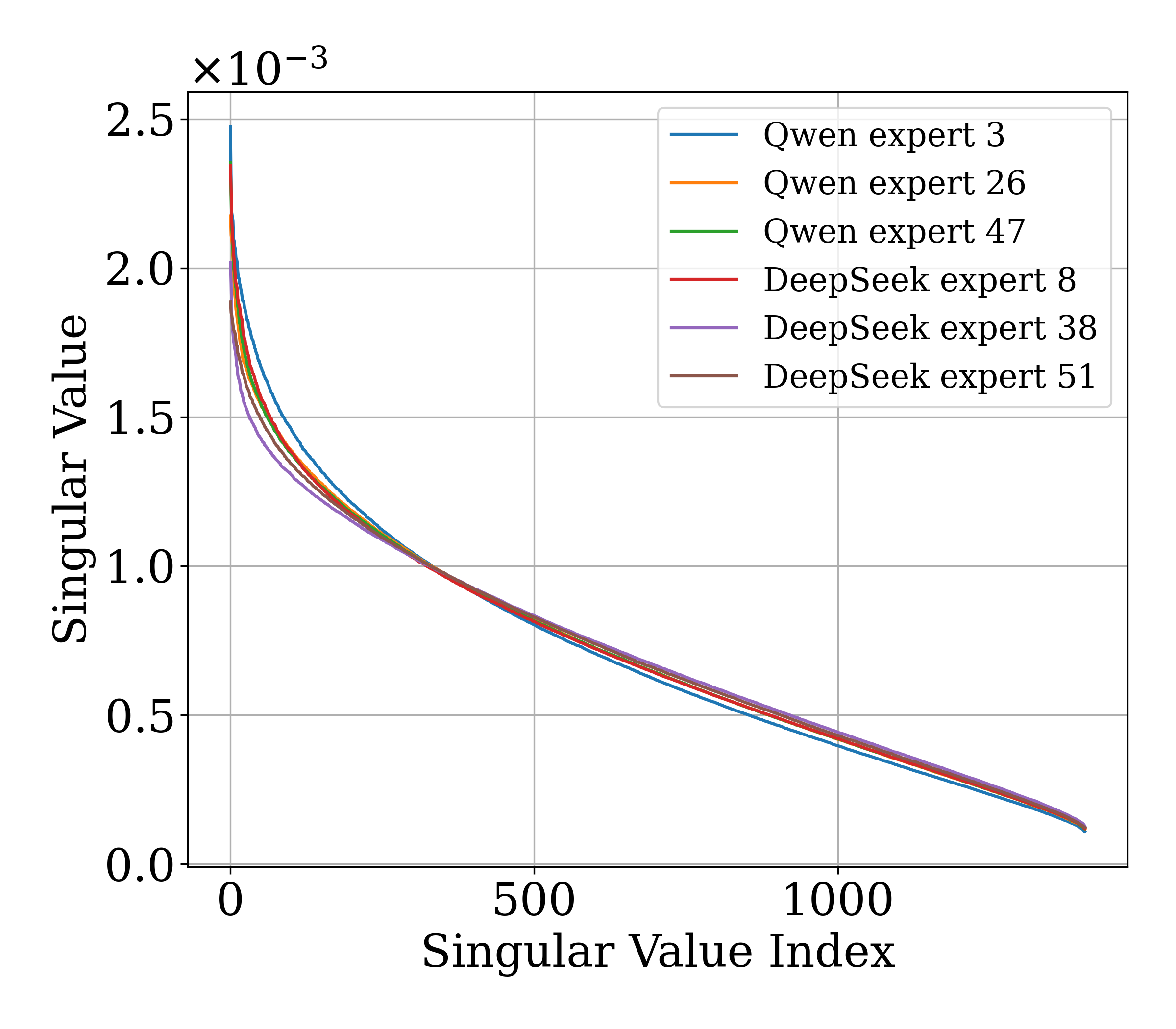}
    }
    ~ % space
    \subfigure[]{
        \includegraphics[width=.22\linewidth]{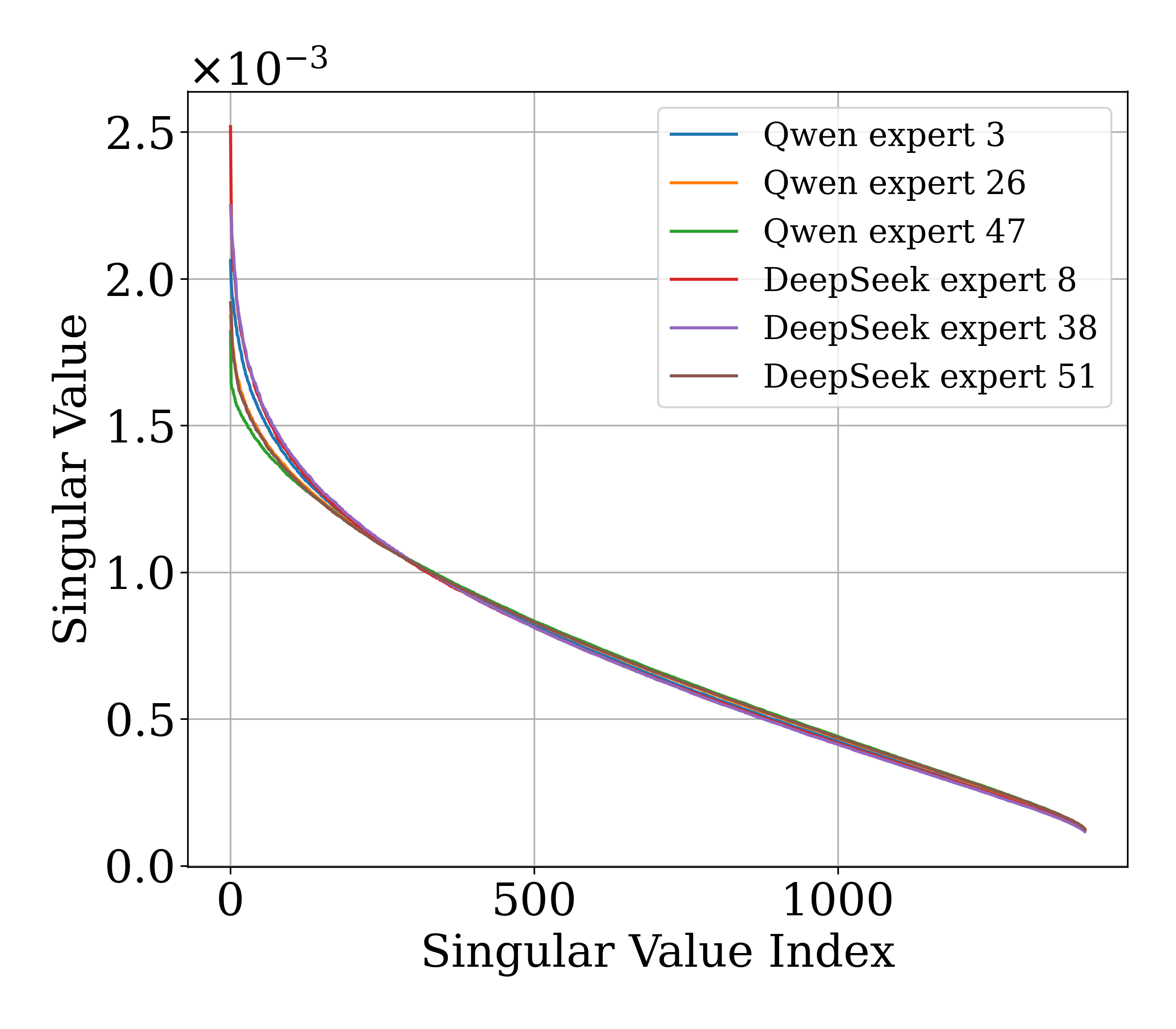}
    }
    ~ % space
    \subfigure[]{
        \includegraphics[width=.22\linewidth]{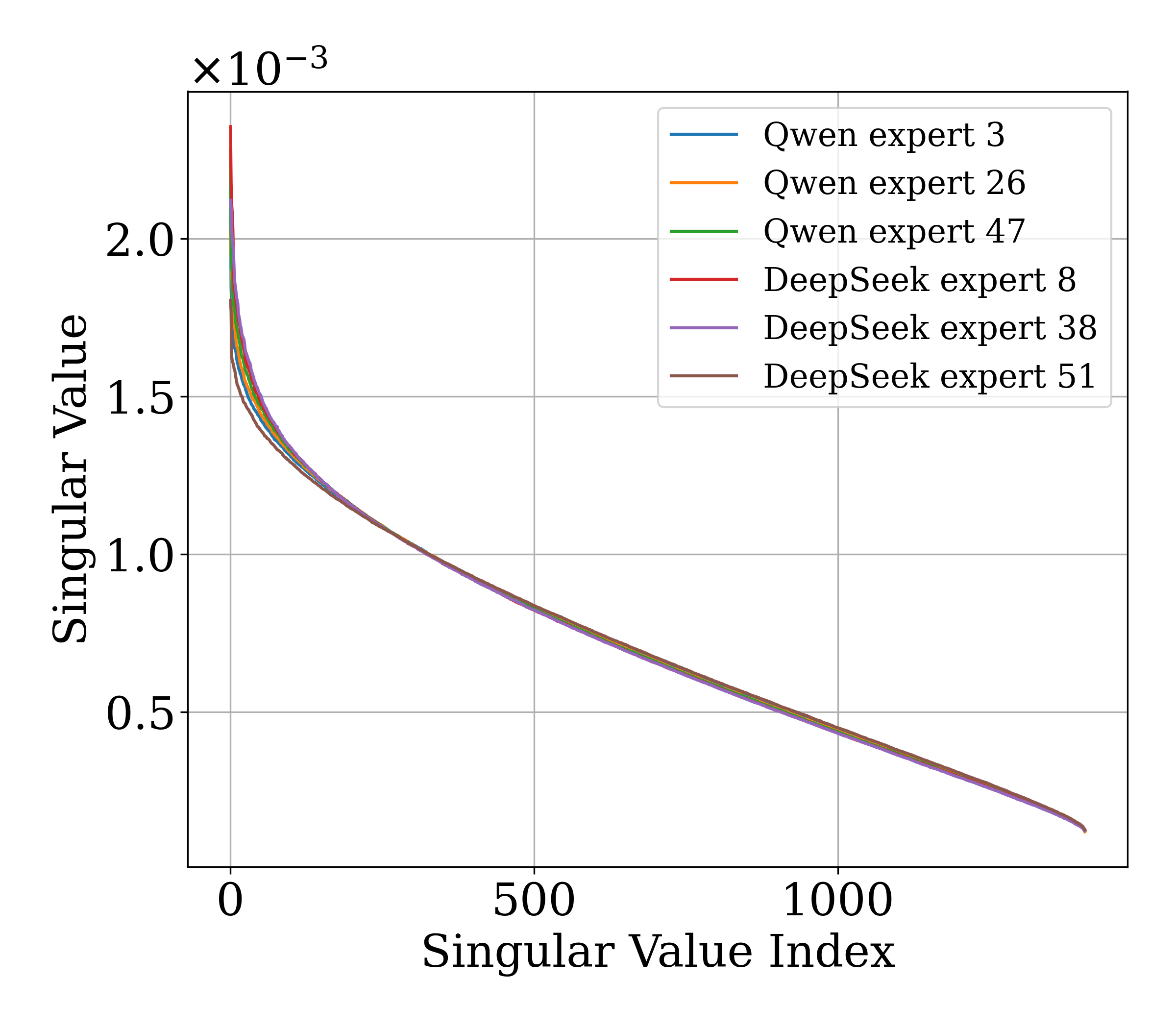}
    }
    ~
    \subfigure[]{
        \includegraphics[width=.22\linewidth]{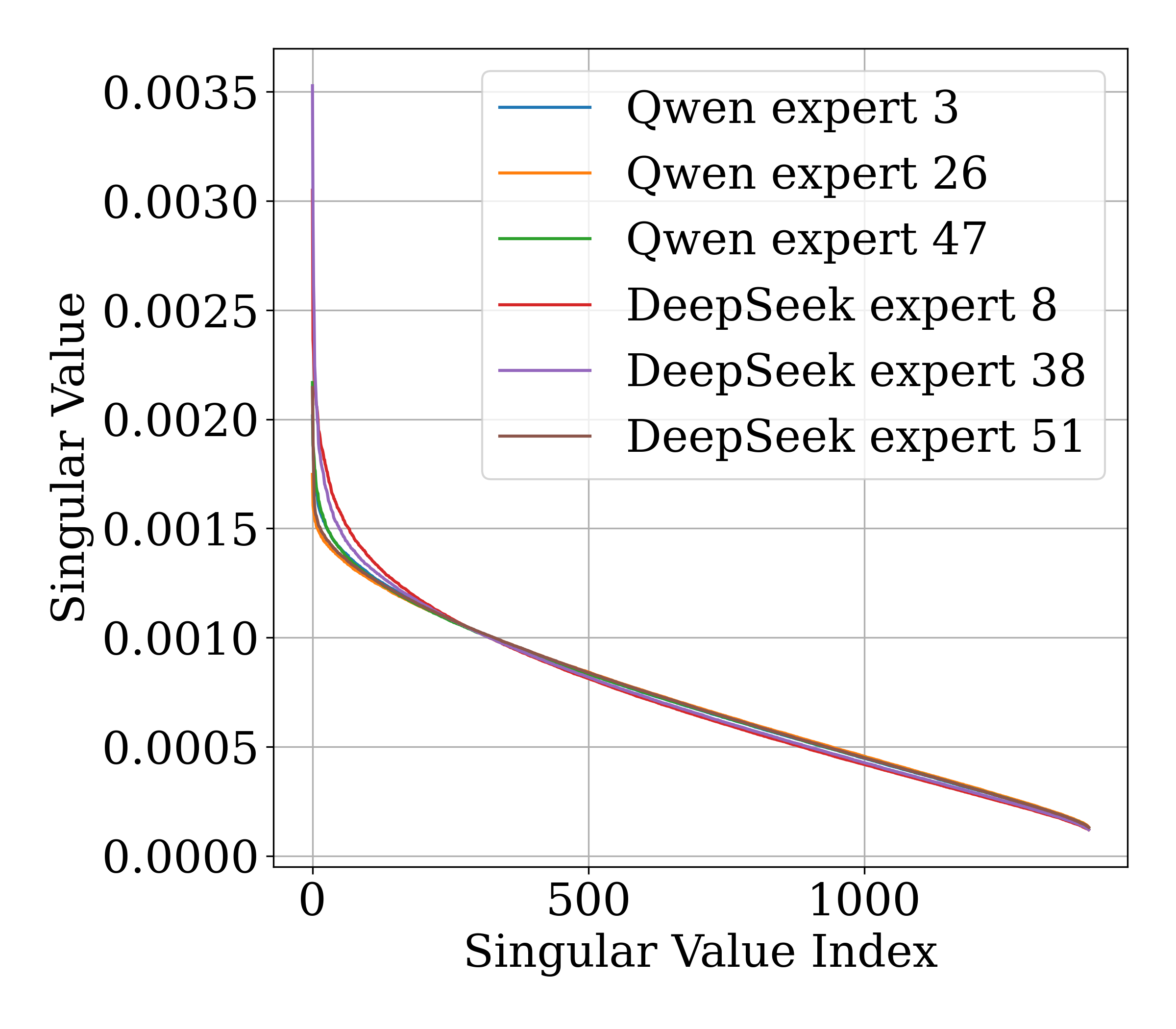}
    }
    \caption{Expert singular value in Qwen~\cite{qwen_moe} and DeepSeek~\cite{dai2024deepseekmoe} models from \textbf{(a)} layer 0, \textbf{(b)} layer 8, \textbf{(c)} layer 16, \textbf{(d)} layer 23. All experts in all model layers exhibit anisotropy, where 1\% leading singular values constitute to more than 30\% energy of the parameter spectrum.}
    \label{fig:append.all_expert_sigvals}
\end{figure*}

% TODO: Add Gate-Expert Parameter Spectral Alignment figures here.

\begin{figure*}[h]
    \centering
    \subfigure[]{
        \includegraphics[width=.22\linewidth]{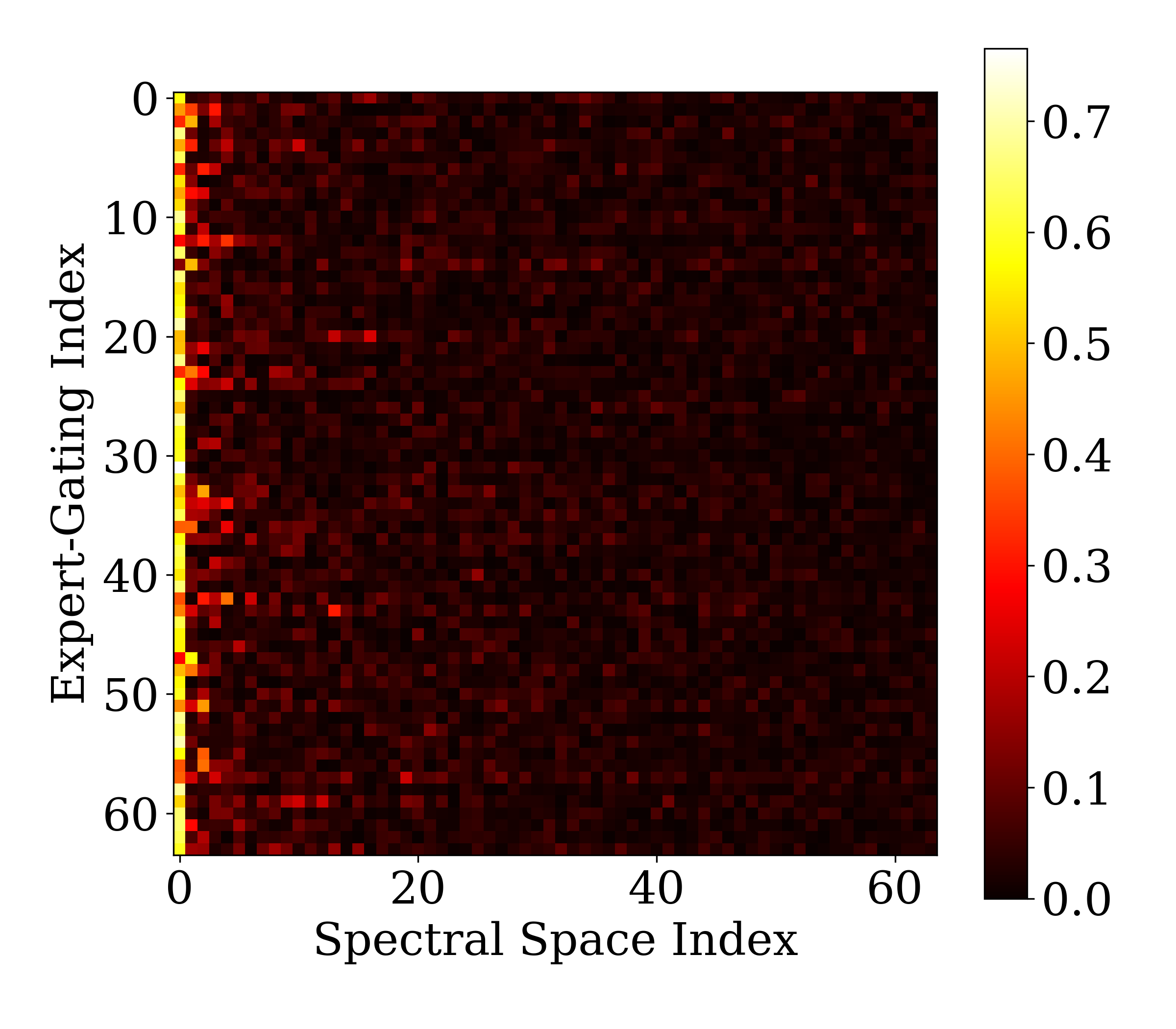}
    }
    ~ % space
    \subfigure[]{
        \includegraphics[width=.22\linewidth]{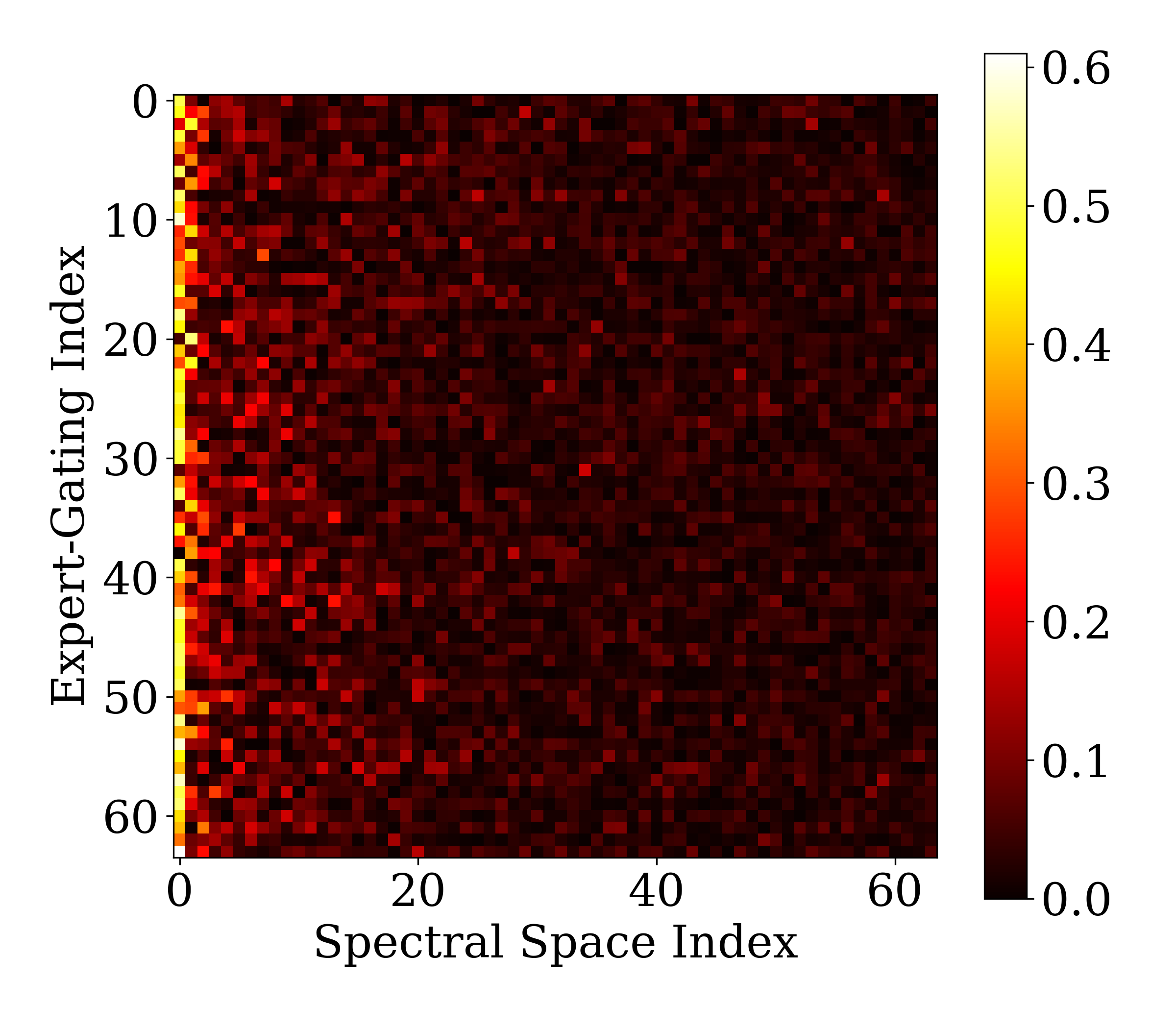}
    }
    ~ % space
    \subfigure[]{
        \includegraphics[width=.22\linewidth]{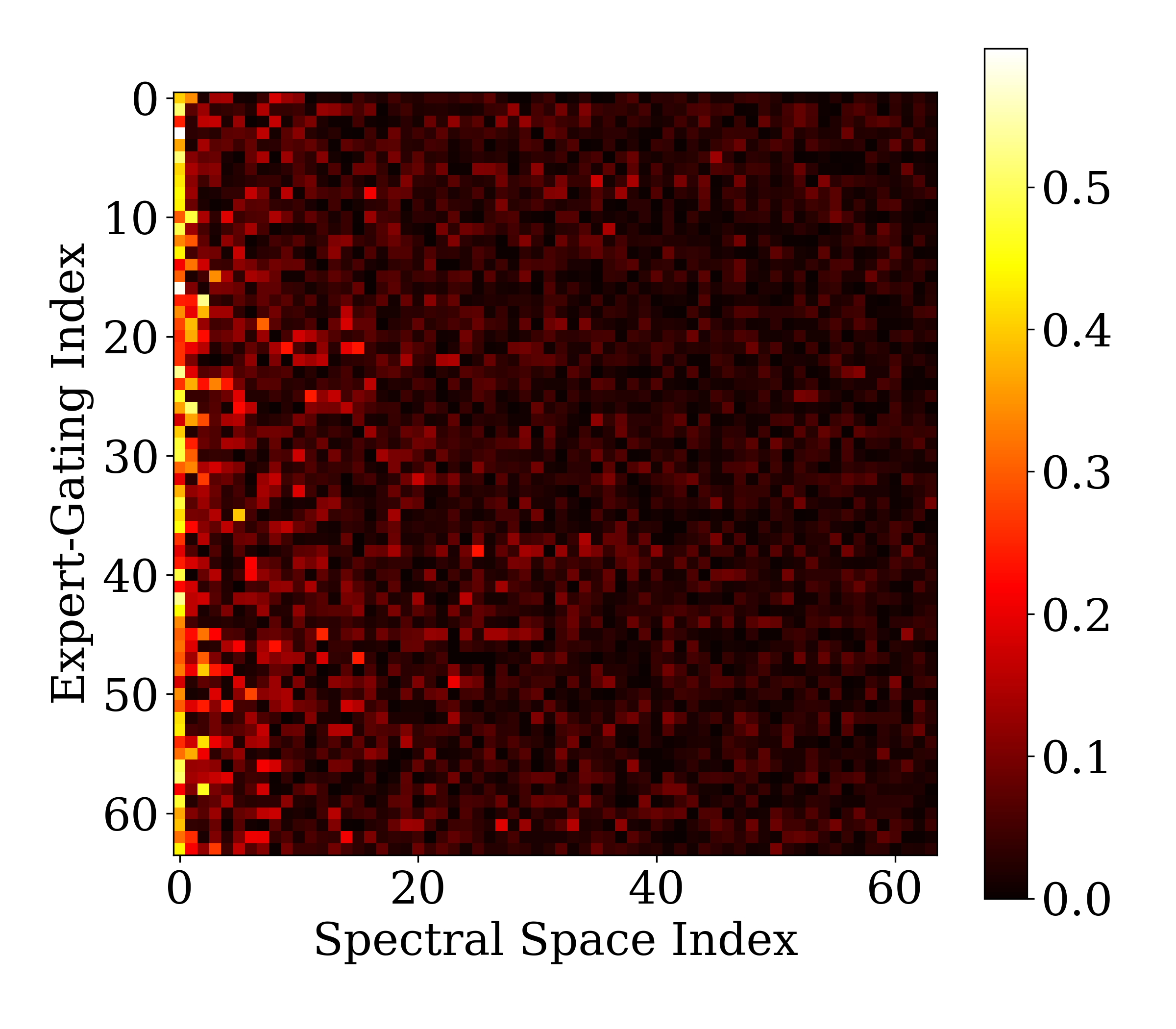}
    }
    ~
    \subfigure[]{
        \includegraphics[width=.22\linewidth]{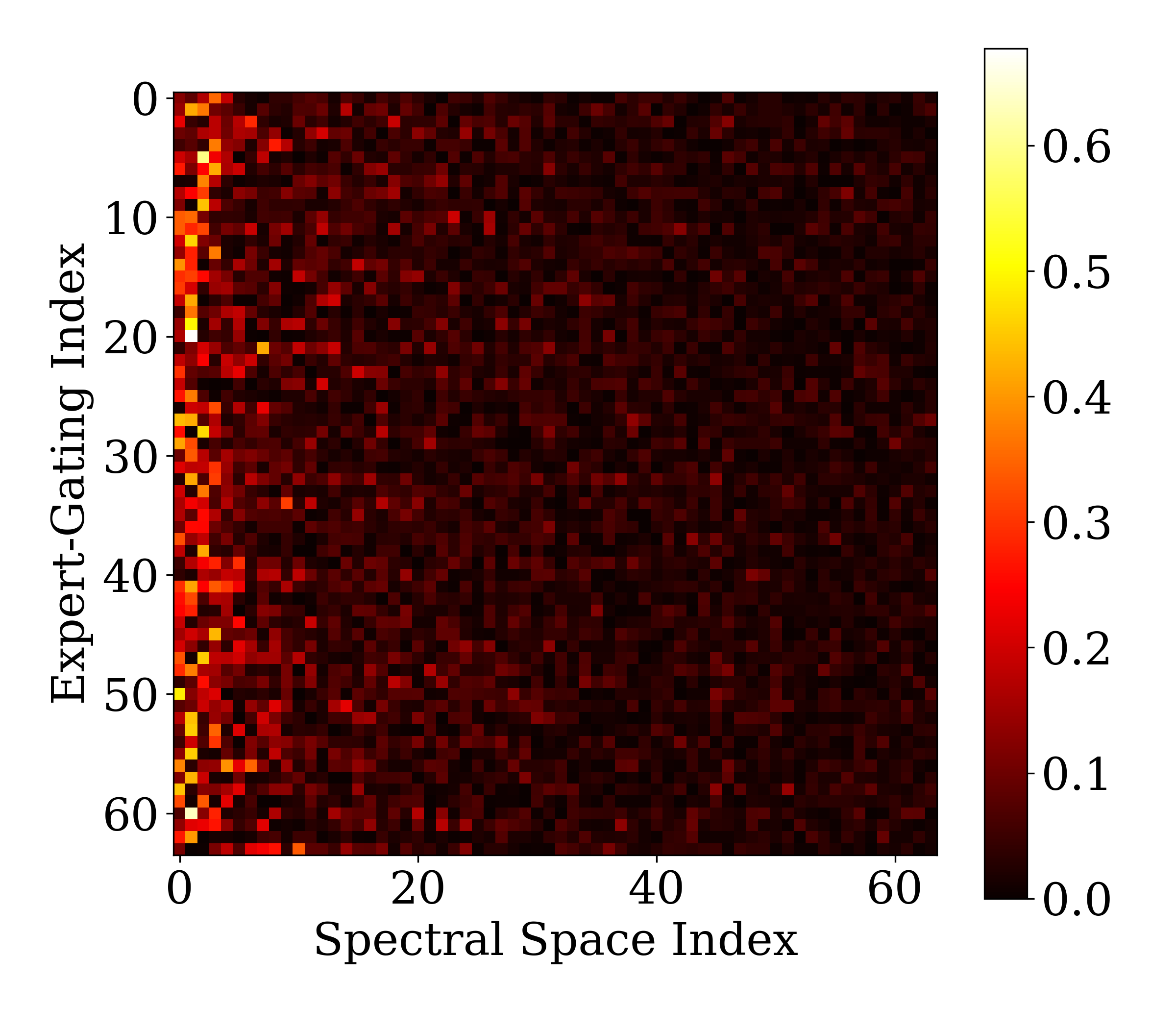}
    }
    \caption{Gate vector and corresponding expert parameter singular vectors alignment in DeepSeek~\cite{dai2024deepseekmoe} models from \textbf{(a)} layer 1, \textbf{(b)} layer 9, \textbf{(c)} layer 17, \textbf{(d)} layer 25. The gate vector is primarily aligned with the leading singular vectors.}
    \label{fig:append.all_gate_param_spectral_alignment_deepseek}
\end{figure*}

\begin{figure*}[h]
    \centering
    \subfigure[]{
        \includegraphics[width=.22\linewidth]{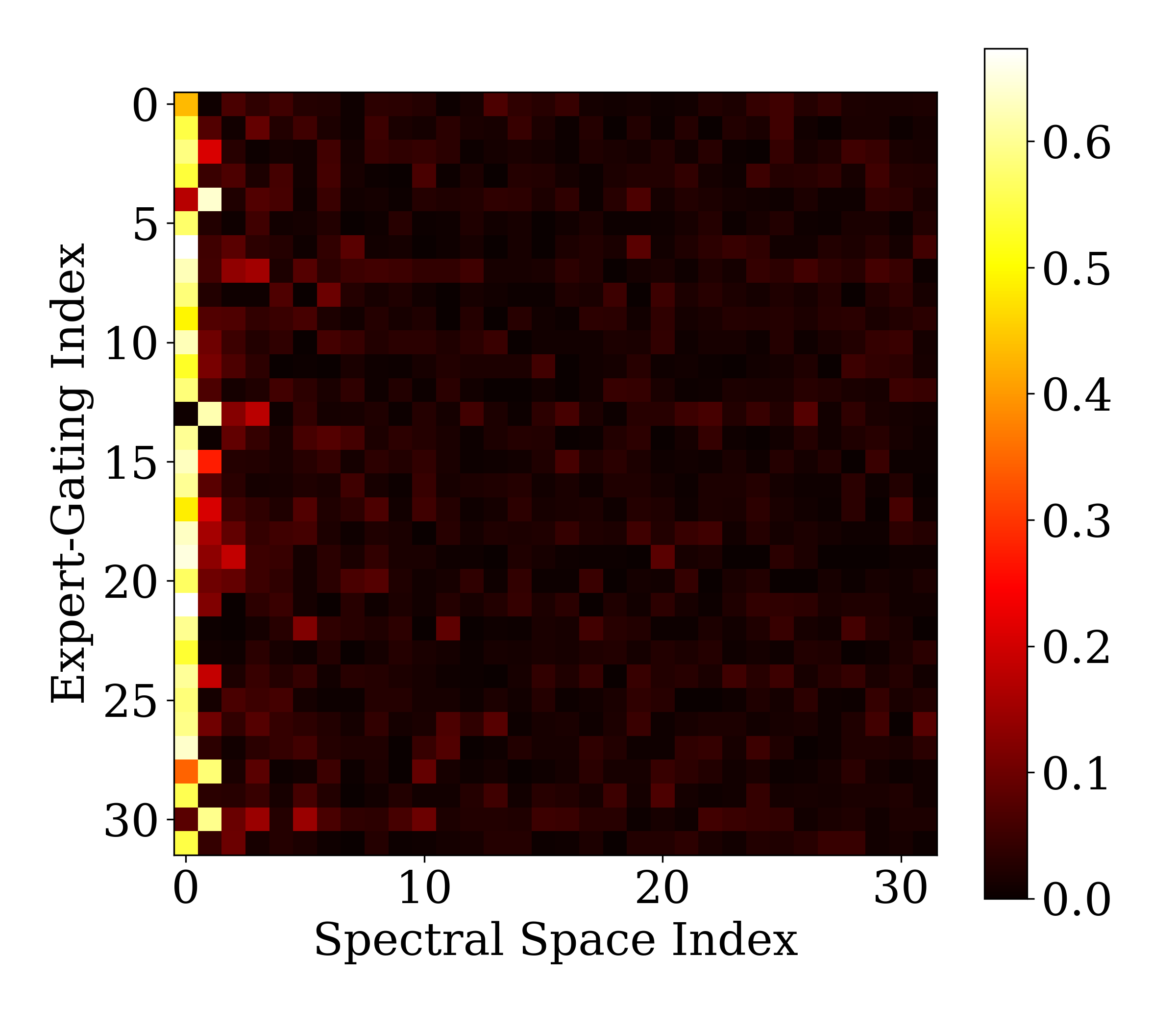}
    }
    ~ % space
    \subfigure[]{
        \includegraphics[width=.22\linewidth]{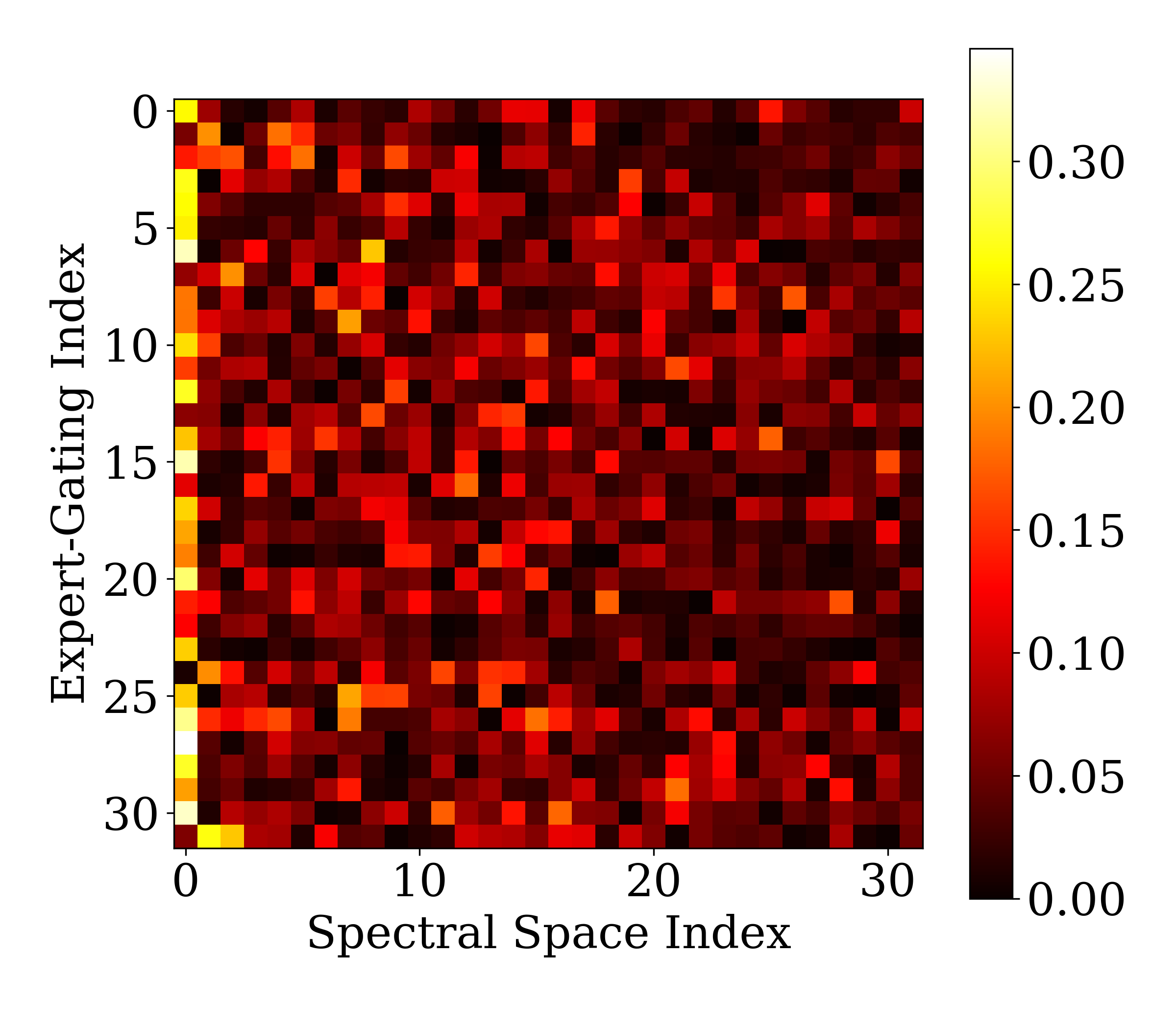}
    }
    ~ % space
    \subfigure[]{
        \includegraphics[width=.22\linewidth]{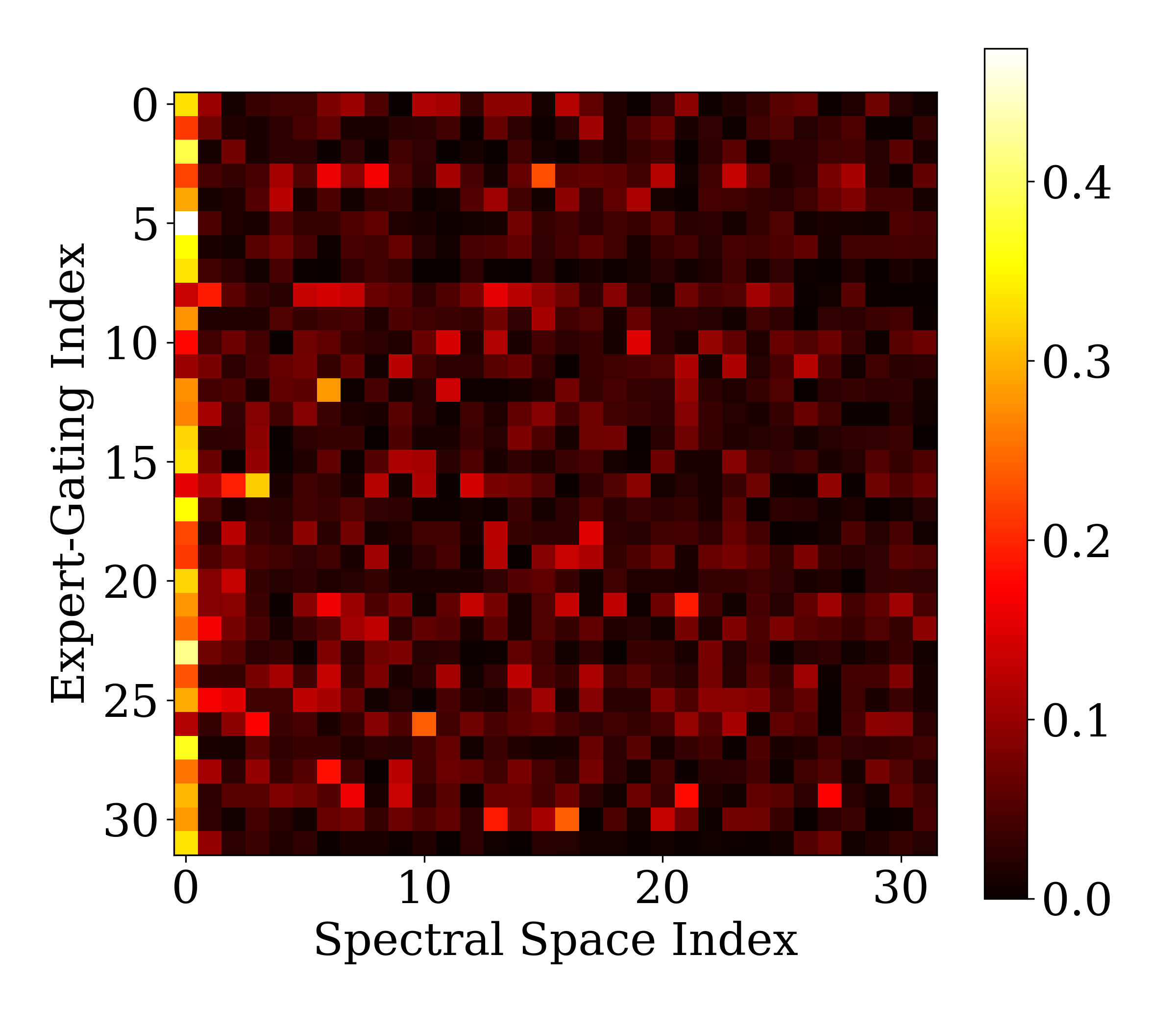}
    }
    ~
    \subfigure[]{
        \includegraphics[width=.22\linewidth]{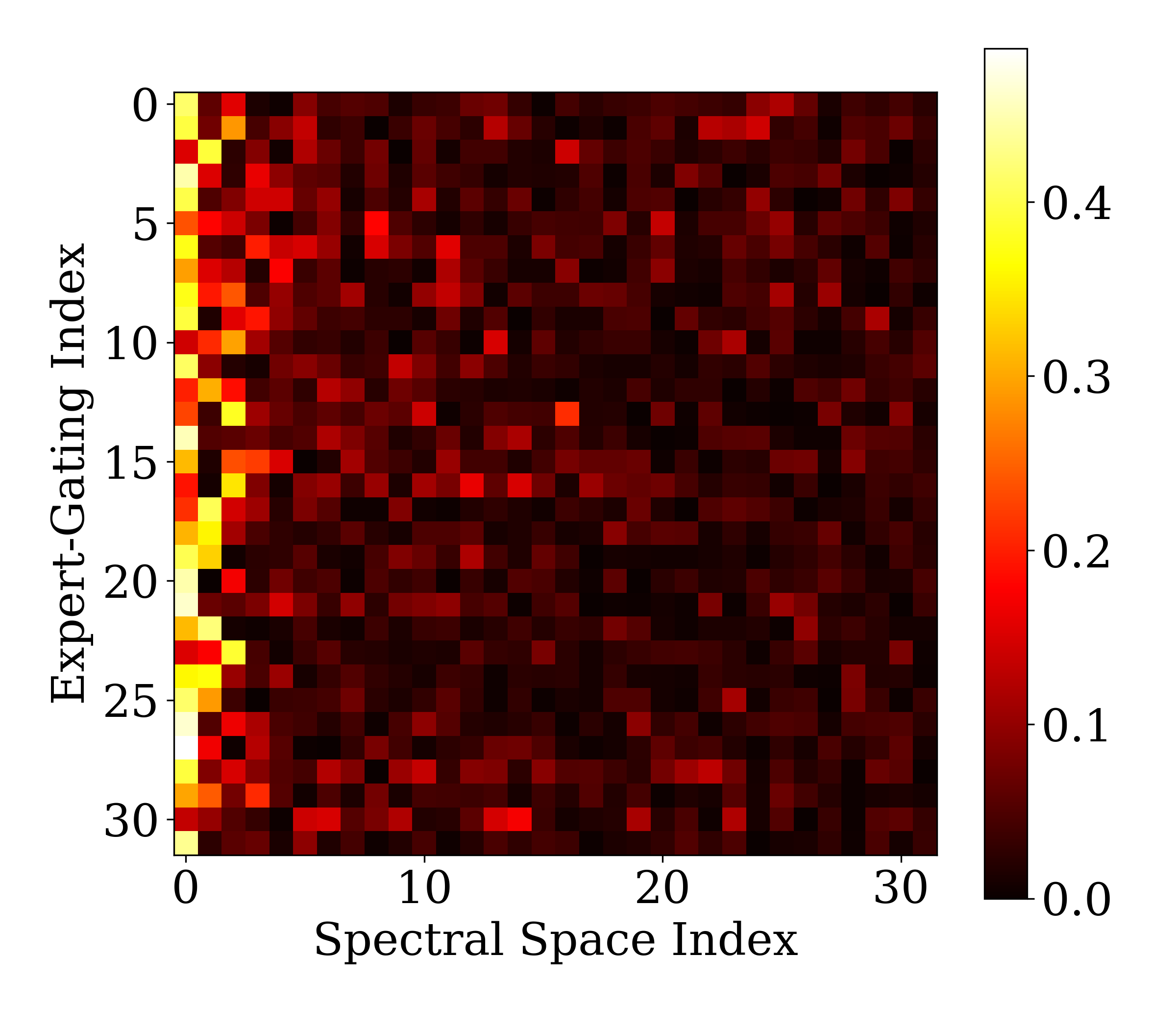}
    }
    \caption{Gate vector and corresponding expert parameter singular vectors alignment in Qwen baseline models (pretrained from scratch in our experiments) from \textbf{(a)} layer 0, \textbf{(b)} layer 15, \textbf{(c)} layer 30, \textbf{(d)} layer 39. The gate vector is primarily aligned with the leading singular vectors.}
    \label{fig:append.all_gate_param_spectral_alignment_qwen_native}
\end{figure*}

\begin{figure*}[h]
    \centering
    \subfigure[]{
        \includegraphics[width=.22\linewidth]{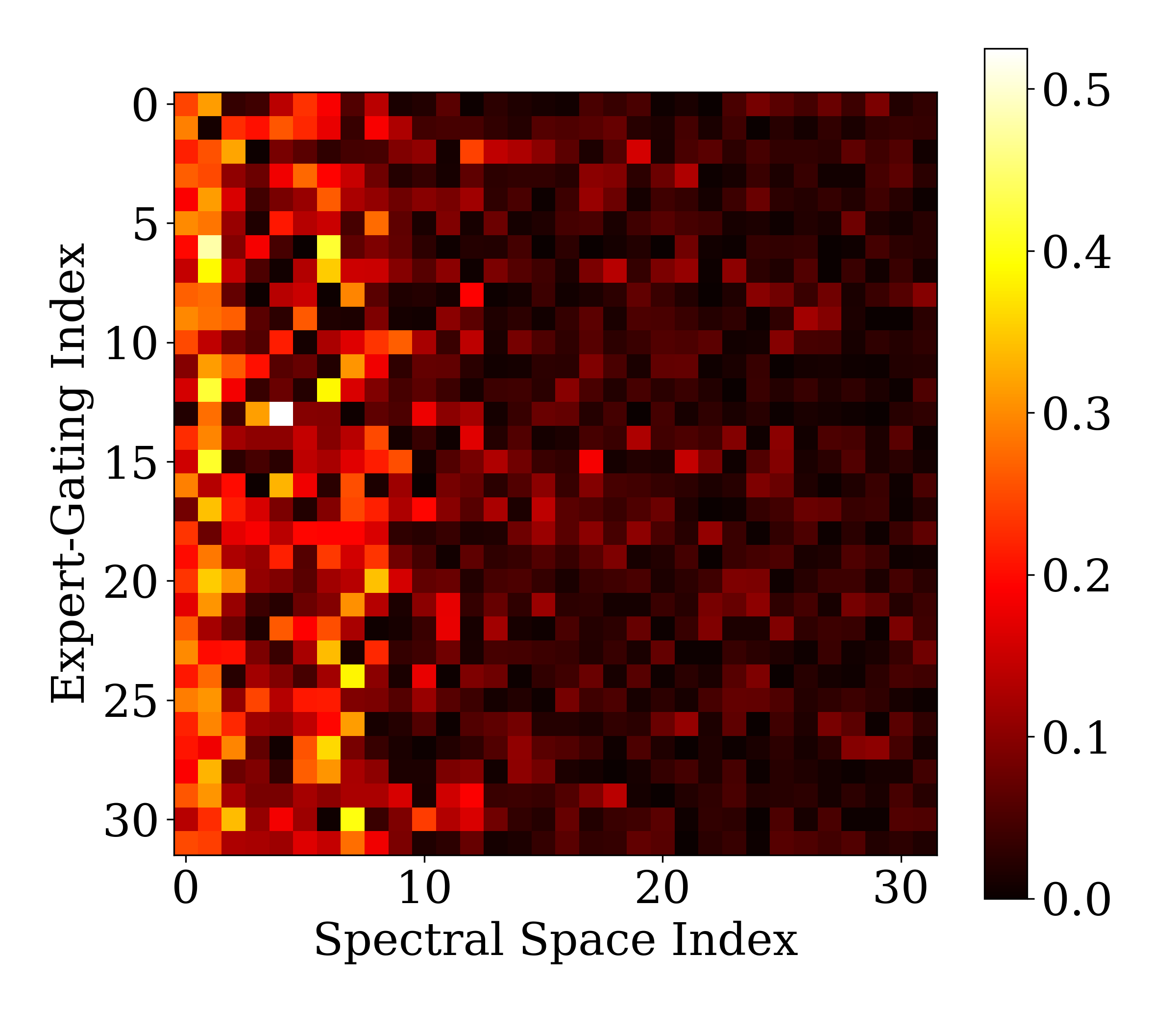}
    }
    ~ % space
    \subfigure[]{
        \includegraphics[width=.22\linewidth]{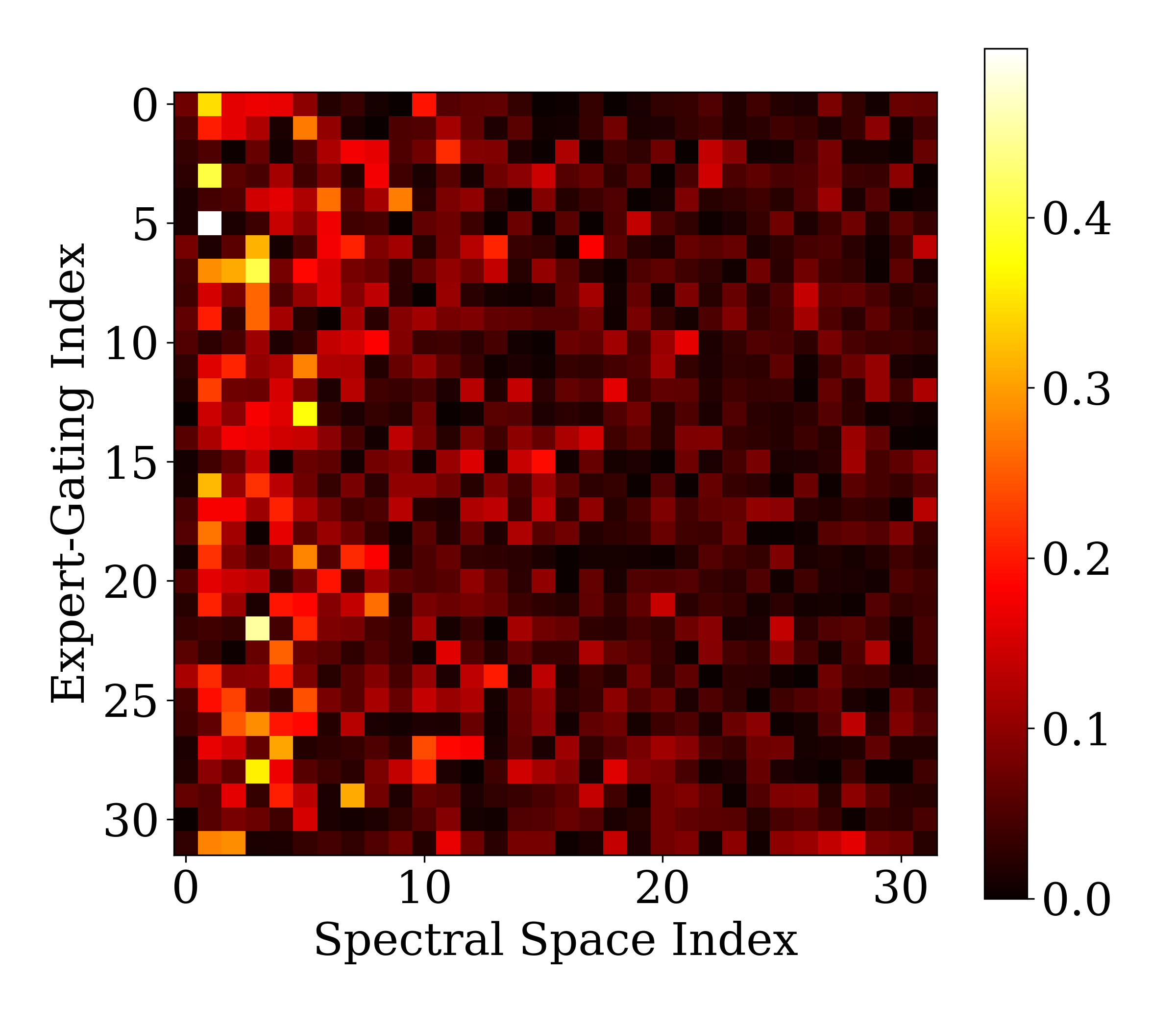}
    }
    ~ % space
    \subfigure[]{
        \includegraphics[width=.22\linewidth]{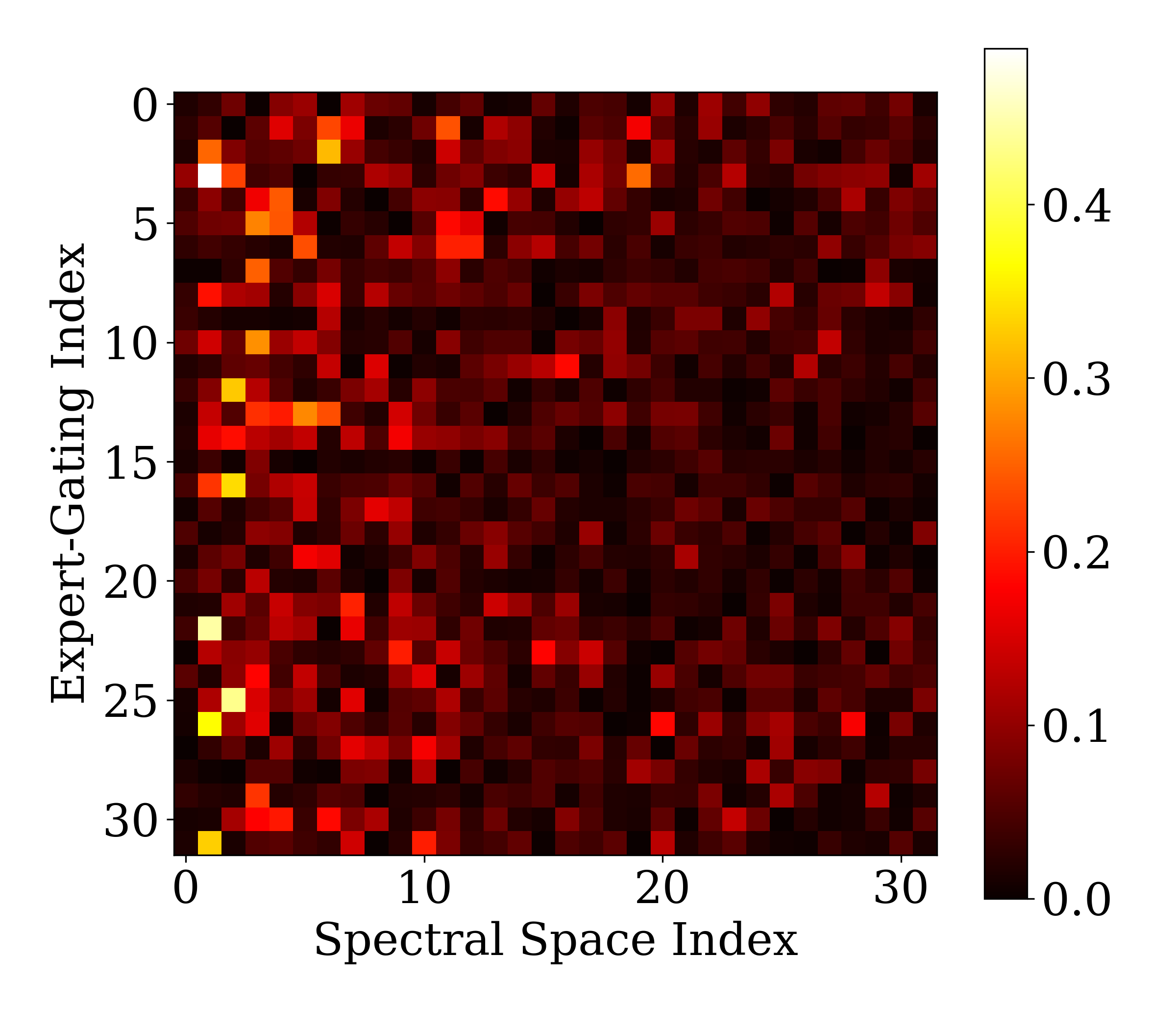}
    }
    ~
    \subfigure[]{
        \includegraphics[width=.22\linewidth]{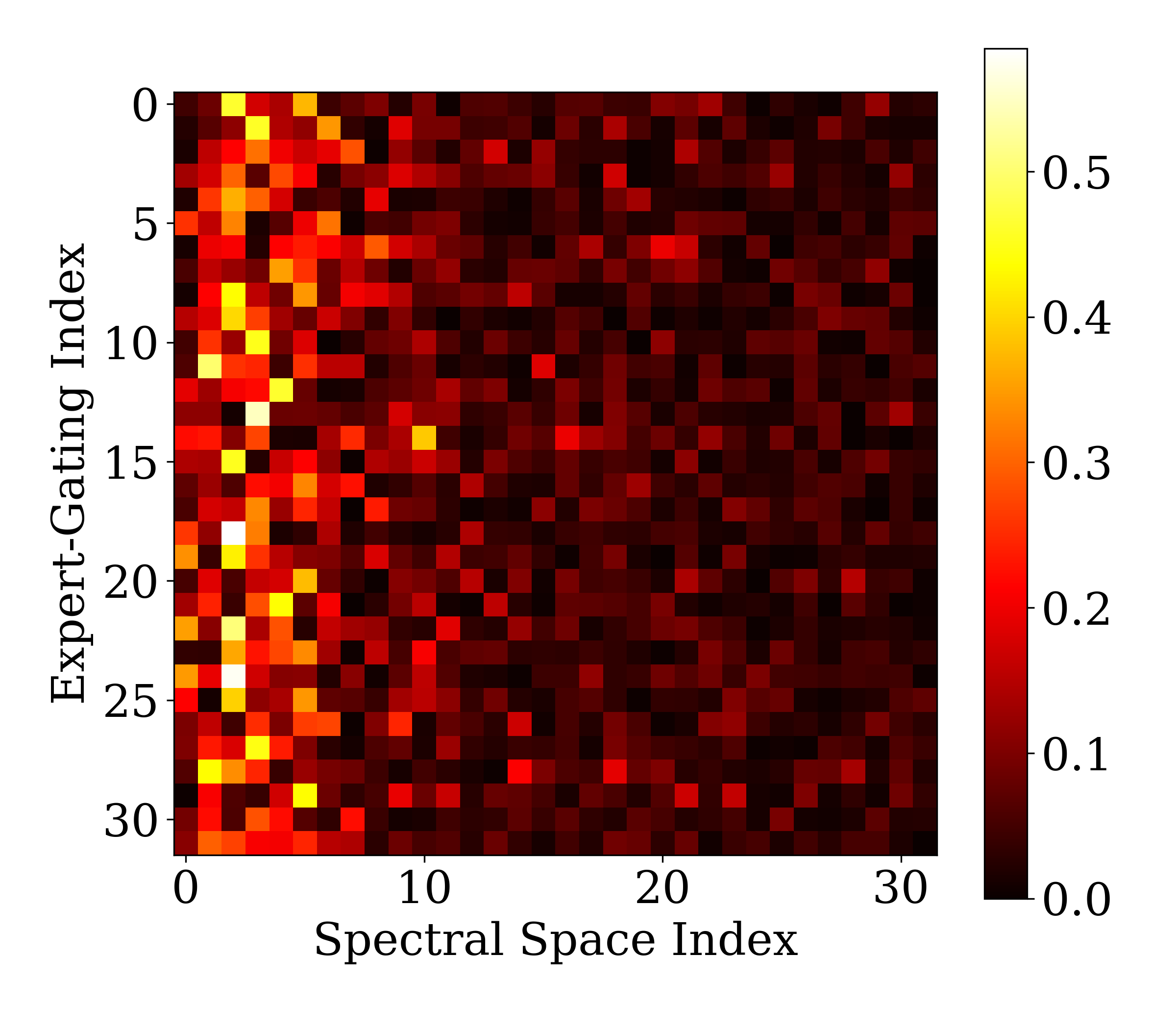}
    }
    \caption{Gate vector and corresponding expert parameter singular vectors alignment in SD-MoE from \textbf{(a)} layer 0, \textbf{(b)} layer 15, \textbf{(c)} layer 30, \textbf{(d)} layer 39. Gating is not dominated by the largest singular vector.}
    \label{fig:append.all_gate_param_spectral_alignment_sdmoe}
\end{figure*}

\begin{figure*}[h]
    \centering
    \subfigure[]{
        \includegraphics[width=.22\linewidth]{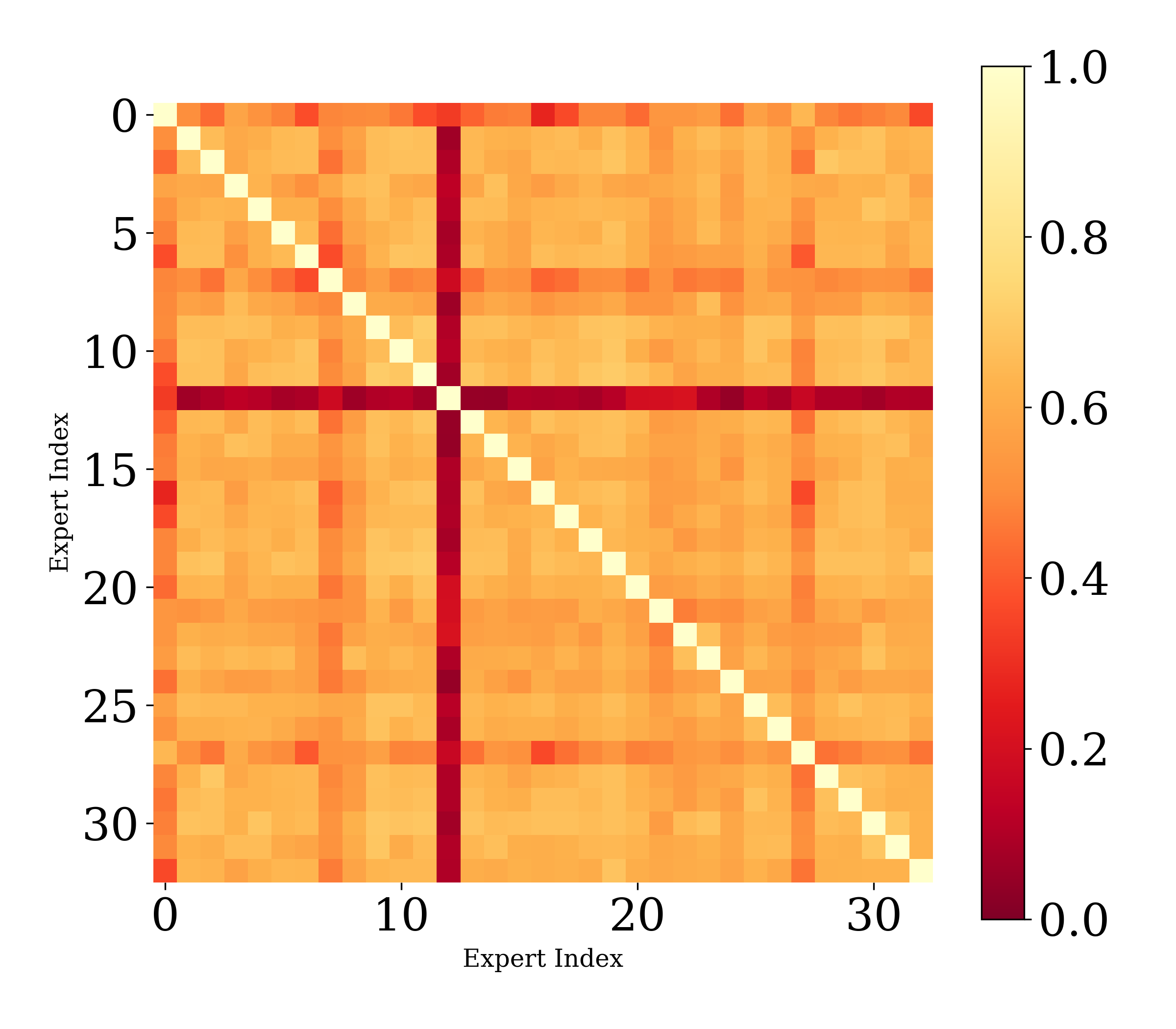}
    }
    ~ % space
    \subfigure[]{
        \includegraphics[width=.22\linewidth]{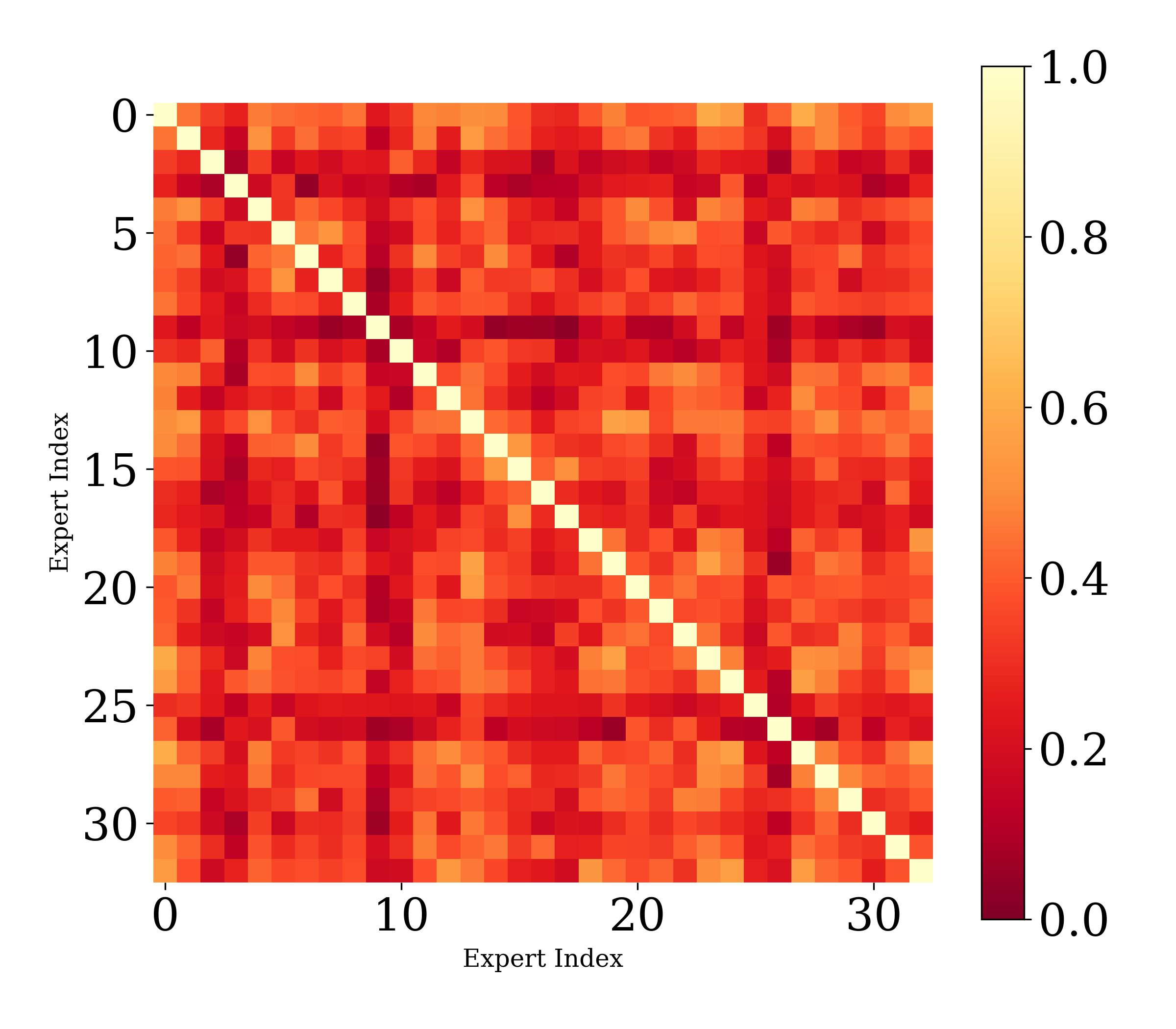}
    }
    ~ % space
    \subfigure[]{
        \includegraphics[width=.22\linewidth]{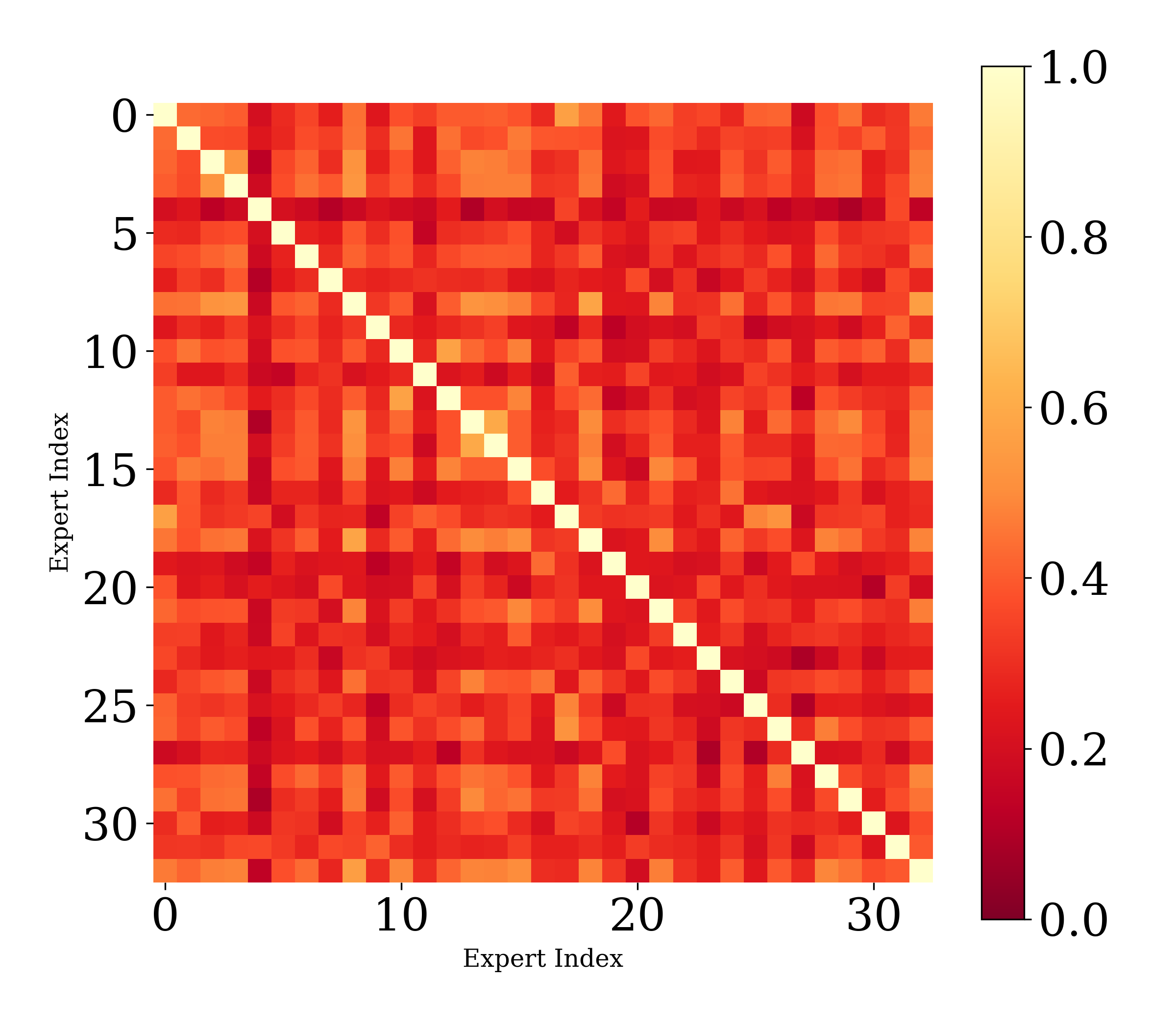}
    }
    ~
    \subfigure[]{
        \includegraphics[width=.22\linewidth]{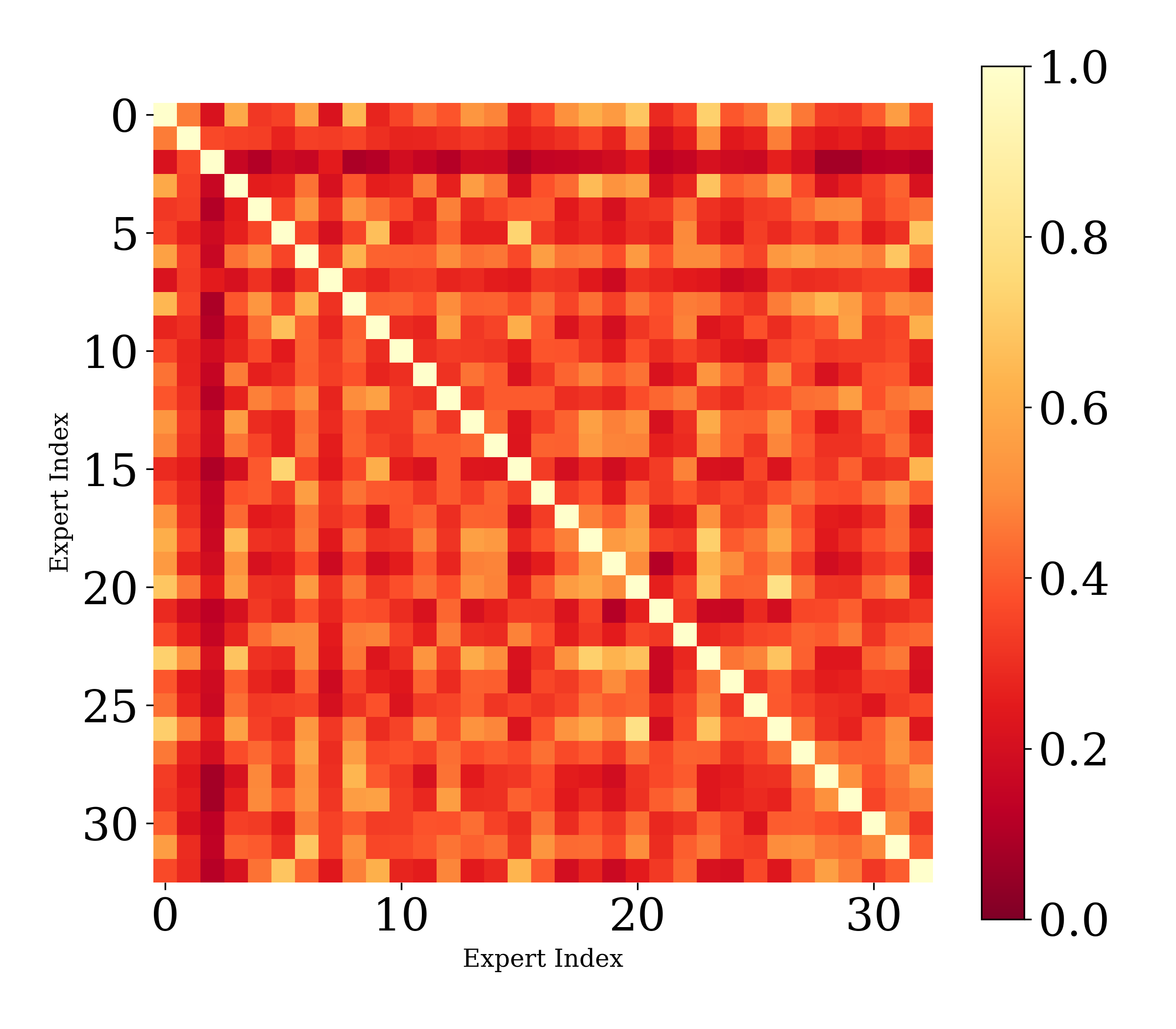}
    }
    \caption{Pairwise expert principal similarity in Qwen baseline models (pretrained from scratch in our experiments) from \textbf{(a)} layer 1, \textbf{(b)} layer 15, \textbf{(c)} layer 30, \textbf{(d)} layer 39. The overlap in expert dominant 1\% spectral subspace is consistent in all model layers.}
    \label{fig:append.all_expert_prin_sim_qwen_native}
\end{figure*}

\begin{figure*}[h]
    \centering
    \subfigure[]{
        \includegraphics[width=.22\linewidth]{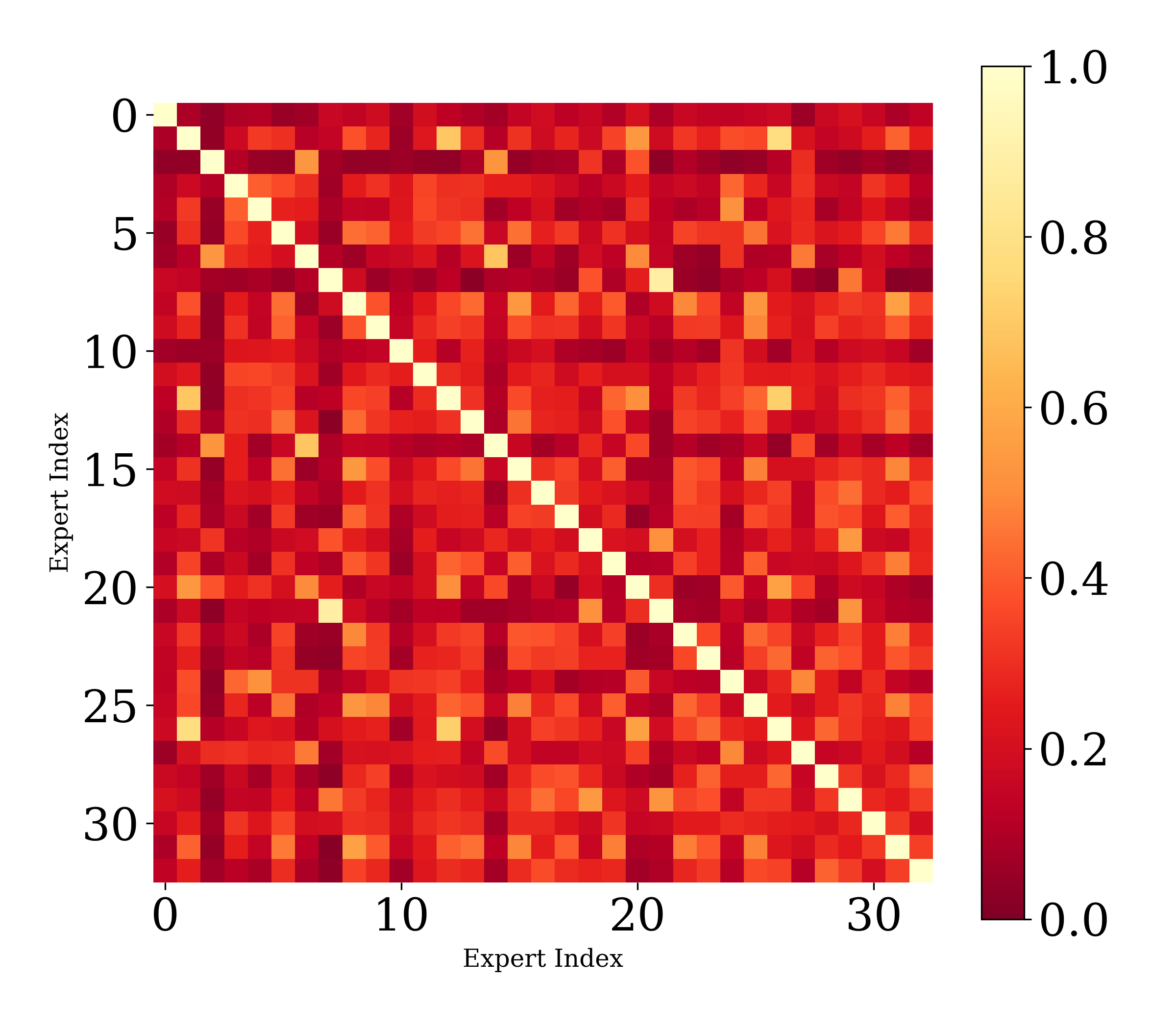}
    }
    ~ % space
    \subfigure[]{
        \includegraphics[width=.22\linewidth]{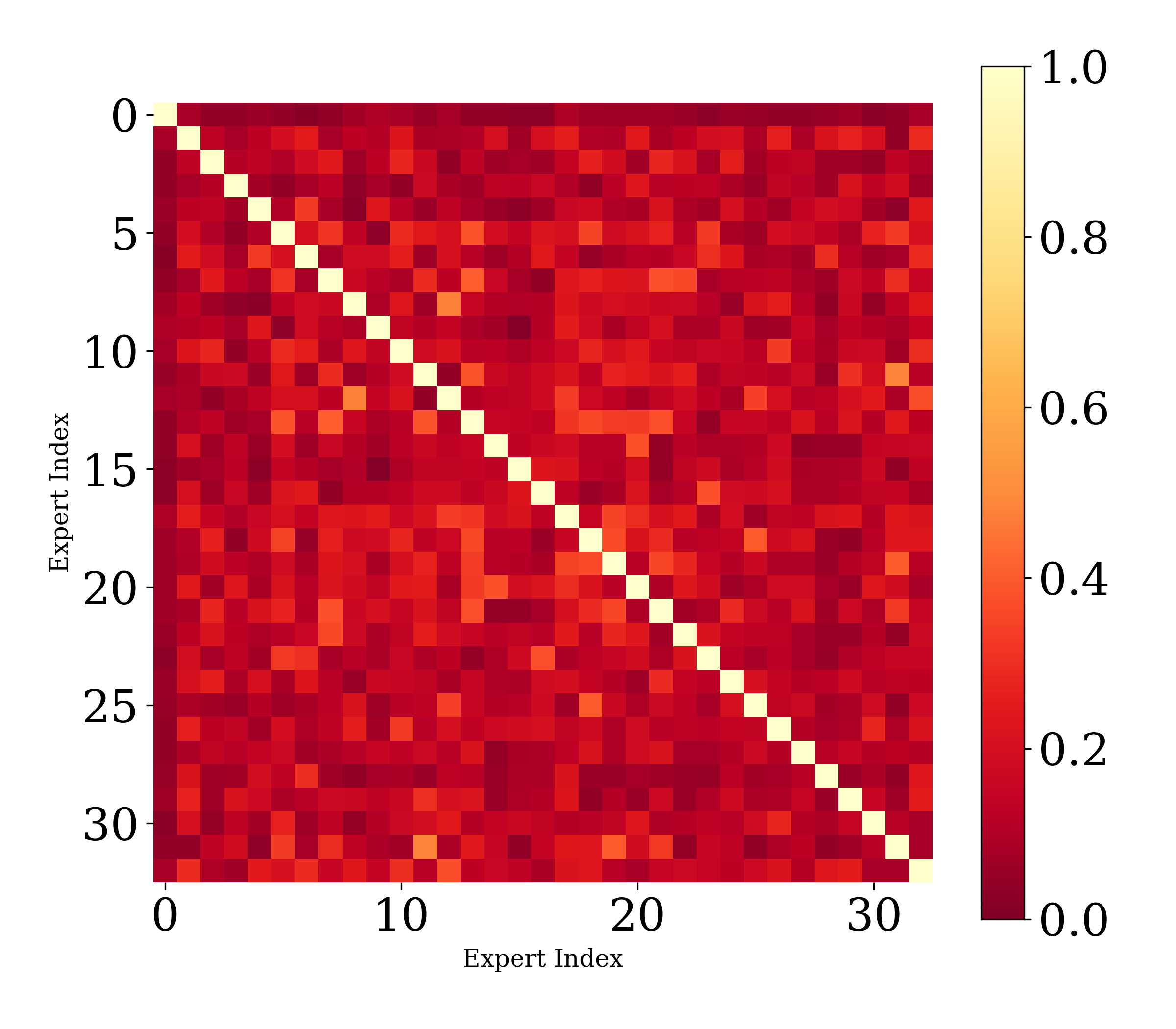}
    }
    ~ % space
    \subfigure[]{
        \includegraphics[width=.22\linewidth]{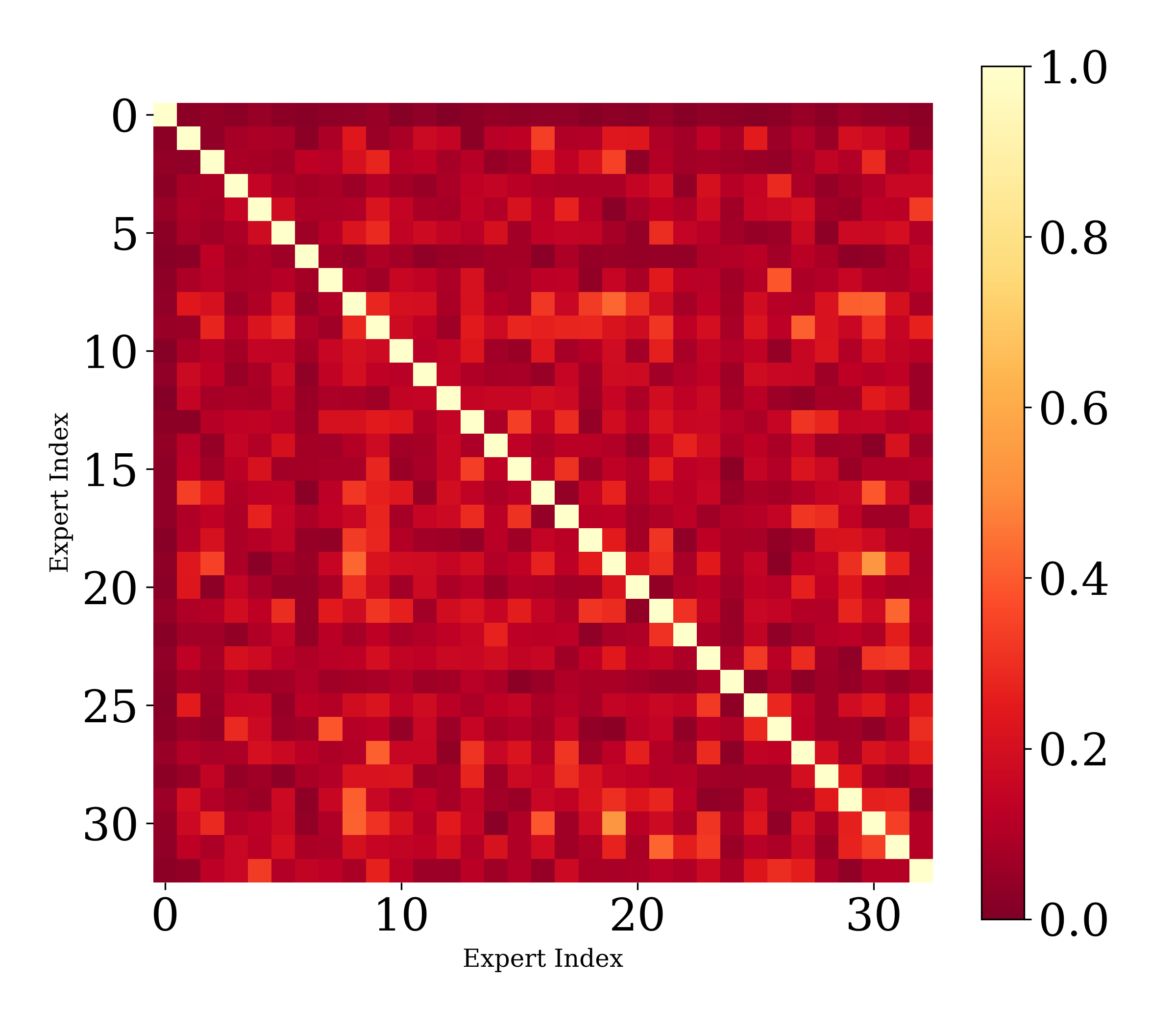}
    }
    ~
    \subfigure[]{
        \includegraphics[width=.22\linewidth]{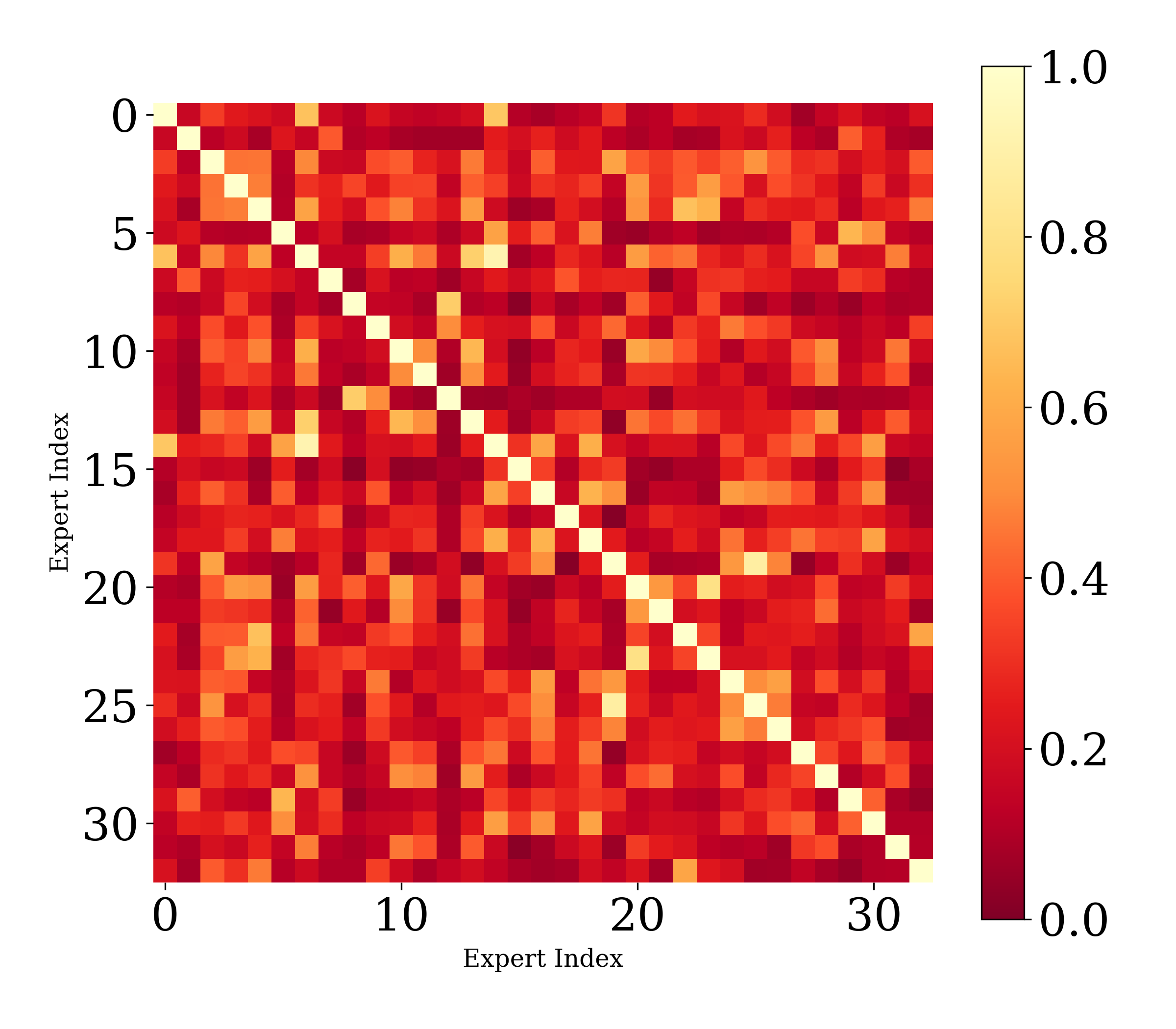}
    }
    \caption{Pairwise expert principal similarity in SD-MoE from \textbf{(a)} layer 1, \textbf{(b)} layer 15, \textbf{(c)} layer 30, \textbf{(d)} layer 39. The near-zero similarities across all expert pairs indicate effective decoupling of parameter directions, demonstrating successful expert specialization and reduced parameter redundancy.}
    \label{fig:append.all_expert_prin_sim_sdmoe}
\end{figure*}

\section{Derivation: a shared input subspace induces a shared gradient component}
\label{app:routing_to_grad}

Recall $\mathbf{G}^{(i)}=\sum_{t\in\mathcal{B}_i}\boldsymbol{\delta}_t \mathbf{x}_t^\top$.
Decompose each token input as $\mathbf{x}_t=\mathbf{P}_C\mathbf{x}_t+\mathbf{P}_{C^\perp}\mathbf{x}_t$, then
\begin{equation}
\mathbf{G}^{(i)}
=\sum_{t\in\mathcal{B}_i}\boldsymbol{\delta}_t (\mathbf{P}_C\mathbf{x}_t)^\top
+\sum_{t\in\mathcal{B}_i}\boldsymbol{\delta}_t (\mathbf{P}_{C^\perp}\mathbf{x}_t)^\top
=: \mathbf{G}_C^{(i)}+\mathbf{R}^{(i)}.
\label{eq:app_G_decomp}
\end{equation}
Moreover, $\mathbf{G}_C^{(i)}$ has no right-action on $C^\perp$:
\begin{equation}
\mathbf{G}_C^{(i)}\mathbf{P}_{C^\perp}=\mathbf{0},
\qquad\text{equivalently}\qquad
\mathbf{G}_C^{(i)}\mathbf{v}=\mathbf{0}\ \ \forall \mathbf{v}\in C^\perp,
\label{eq:app_support_on_C}
\end{equation}
since $\mathbf{P}_C\mathbf{v}=\mathbf{0}$ for all $\mathbf{v}\in C^\perp$ and thus
$
\mathbf{G}_C^{(i)}\mathbf{v}
=\sum_{t\in\mathcal{B}_i}\boldsymbol{\delta}_t (\mathbf{P}_C\mathbf{x}_t)^\top \mathbf{v}
=\sum_{t\in\mathcal{B}_i}\boldsymbol{\delta}_t \mathbf{x}_t^\top (\mathbf{P}_C\mathbf{v})
=\mathbf{0}.
$
Therefore, every expert gradient $\mathbf{G}^{(i)}$ contains a component $\mathbf{G}_C^{(i)}$ whose input-side (right-subspace) effect is supported on the shared subspace $C$.

\section{Method Details}
\label{appendix:method}

Here we detail the initialization of unique experts in the orthogonal complement subspace, and the pseudo code is given in Algorithm~\ref{alg:unique_init}.
Given a weight matrix $\mathbf{W} \in \mathbb{R}^{m \times n}$ and its rank-$r$ shared component obtained via SVD, $\mathbf{W}_c = \mathbf{U}_k \boldsymbol{\Sigma}_k \mathbf{V}_k^\top$, where $\mathbf{U}_k \in \mathbb{R}^{m \times r}$ and $\mathbf{V}_k \in \mathbb{R}^{n \times r}$ have orthonormal columns, we aim to initialize each unique expert with a matrix supported entirely in the orthogonal complement of the shared subspace.

Specifically, for the row space, we seek a random matrix $\widetilde{\mathbf{U}} \in \mathbb{R}^{m \times (m - r)}$ such that:
\begin{enumerate}
    \item $\widetilde{\mathbf{U}}^\top \widetilde{\mathbf{U}} = \mathbf{I}_{m - r}$ (columns are orthonormal);
    \item $\mathbf{U}_k^\top \widetilde{\mathbf{U}} = \mathbf{0}$ (orthogonal to the shared row subspace).
\end{enumerate}
An analogous condition holds for $\widetilde{\mathbf{V}} \in \mathbb{R}^{n \times (n - r)}$ in the column space.

To achieve this, we first sample a Gaussian random matrix $\mathbf{Z} \in \mathbb{R}^{m \times (m - r)}$. We then project it onto the orthogonal complement of $\mathrm{col}(\mathbf{U}_k)$ using the projector $\mathbf{P}_\perp = \mathbf{I} - \mathbf{U}_k \mathbf{U}_k^\top$:
\[
\mathbf{Z}_\perp = (\mathbf{I} - \mathbf{U}_k \mathbf{U}_k^\top) \mathbf{Z}.
\]
Since $\mathbf{Z}_\perp$ lies in the desired subspace but its columns are not necessarily orthogonal, we apply QR decomposition to obtain an orthonormal basis:
\[
\widetilde{\mathbf{U}} = \mathrm{QR}(\mathbf{Z}_\perp).
\]
The same procedure is applied to the column space using $\mathbf{V}_k$. Finally, the unique weight is constructed as $\mathbf{W}_u = \widetilde{\mathbf{U}} \, \widetilde{\boldsymbol{\Sigma}} \, \widetilde{\mathbf{V}}^\top$, where $\widetilde{\boldsymbol{\Sigma}}$ is a diagonal matrix populated with the long-tail singular values of $\mathbf{W}$ (i.e., $\sigma_{r+1}, \dots, \sigma_{\min(m,n)}$).

This procedure ensures that each unique expert starts in a direction fully orthogonal to the common subspace while maintaining proper scale and diversity across experts.

\begin{algorithm}
\caption{Initialization of Unique Expert in Orthogonal Complement}
\label{alg:unique_init}
\begin{algorithmic}[1]
\REQUIRE Shared bases $\mathbf{U}_k \in \mathbb{R}^{m \times r}$, $\mathbf{V}_k \in \mathbb{R}^{n \times r}$; number of experts $E$; long-tail singular values $\boldsymbol{\sigma}_{\text{tail}} \in \mathbb{R}^{\ell}$, $\ell = \min(m,n) - r$
\ENSURE Unique weights $\{\mathbf{W}_u^{(i)}\}_{i=1}^E$

\FOR{$i = 1$ to $E$}
    \STATE Sample $\mathbf{Z}_U \sim \mathcal{N}(0, 1)^{m \times (m - r)}$
    \STATE $\mathbf{Z}_U \gets \mathbf{Z}_U - \mathbf{U}_k (\mathbf{U}_k^\top \mathbf{Z}_U)$ \COMMENT{Project to $\mathrm{col}(\mathbf{U}_k)^\perp$}
    \STATE $\widetilde{\mathbf{U}}^{(i)} \gets \mathrm{QR}(\mathbf{Z}_U)$ \COMMENT{Ortho-normalize}
    
    \STATE Sample $\mathbf{Z}_V \sim \mathcal{N}(0, 1)^{n \times (n - r)}$
    \STATE $\mathbf{Z}_V \gets \mathbf{Z}_V - \mathbf{V}_k (\mathbf{V}_k^\top \mathbf{Z}_V)$ \COMMENT{Project to $\mathrm{col}(\mathbf{V}_k)^\perp$}
    \STATE $\widetilde{\mathbf{V}}^{(i)} \gets \mathrm{QR}(\mathbf{Z}_V)$
    
    \STATE Form $\widetilde{\boldsymbol{\Sigma}} = \mathrm{diag}(\boldsymbol{\sigma}_{\text{tail}})$ \COMMENT{Pad with zeros if $m \neq n$}
    \STATE $\mathbf{W}_u^{(i)} \gets \widetilde{\mathbf{U}}^{(i)} \, \widetilde{\boldsymbol{\Sigma}} \, (\widetilde{\mathbf{V}}^{(i)})^\top$
\ENDFOR

\end{algorithmic}
\end{algorithm}

\section{Detailed Model Configurations in Our Experiment}
\label{appendix:exp-settings}

The following are the model configurations used in our experiments, where the keys follow the HuggingFace Transformers\cite{wolf2020transformers} library.

\begin{table}[h]
\centering
\caption{
    Model Configurations of Qwen Models
}
\label{tab:model_config}
\begin{tabular}{p{0.2\textwidth}cc}
\toprule
\textbf{Configs} & \textbf{2B-A0.8B} & \textbf{7B-A1.5B} \\
\midrule
num\_hidden\_layers & 32 & 40 \\
hidden\_size & 1024 & 1536 \\
num\_attention\_heads & 16 & 16 \\
head\_dim & 128 & 128 \\
num\_key\_value\_heads & 8 & 8 \\
moe\_intermediate\_size & 2048 & 1024 \\
num\_experts\_per\_tok & 2 & 4 \\
num\_experts & 9(1 shared) & 33(1 shared) \\
\bottomrule
\end{tabular}
\end{table}
 
\begin{table}[h]
\centering
\caption{
    Configuration of Deepseek-2B-A0.8B
}
\begin{tabular}{p{0.25\textwidth}c}
\toprule
\textbf{Configs} & \textbf{Deepseek-2B-A0.8B} \\
\midrule
num\_hidden\_layers & 32 \\
hidden\_size & 1024 \\

num\_attention\_heads & 16 \\
num\_key\_value\_heads & 8 \\
kv\_lora\_rank & 64 \\
q\_lora\_rank & 192 \\
qk\_nope\_head\_dim & 128 \\
qk\_rope\_head\_dim & 64 \\
v\_head\_dim & 128 \\

moe\_intermediate\_size & 2048 \\
num\_experts\_per\_tok & 2 \\
num\_experts & 9(1 shared) \\
moe\_layer\_freq & 1 \\
first\_k\_dense\_replace & 0 \\
n\_group & 1 \\
topk\_group & 1 \\
routed\_scaling\_factor & 1 \\
\bottomrule
\end{tabular}
\end{table}

\section{Learning Rate Ablations on SD-MoE}
\label{app:same_lr}

As discussed in the main text, SD-MoE remains stable under learning rates more than $4\times$ larger than those tolerated by baseline MoE models. To further characterize the optimization behavior under high learning rates, we examine the evolution of the load-balance loss during training.

\begin{figure*}[t]
    \centering
    \subfigure[]{
        \includegraphics[width=.35\linewidth]{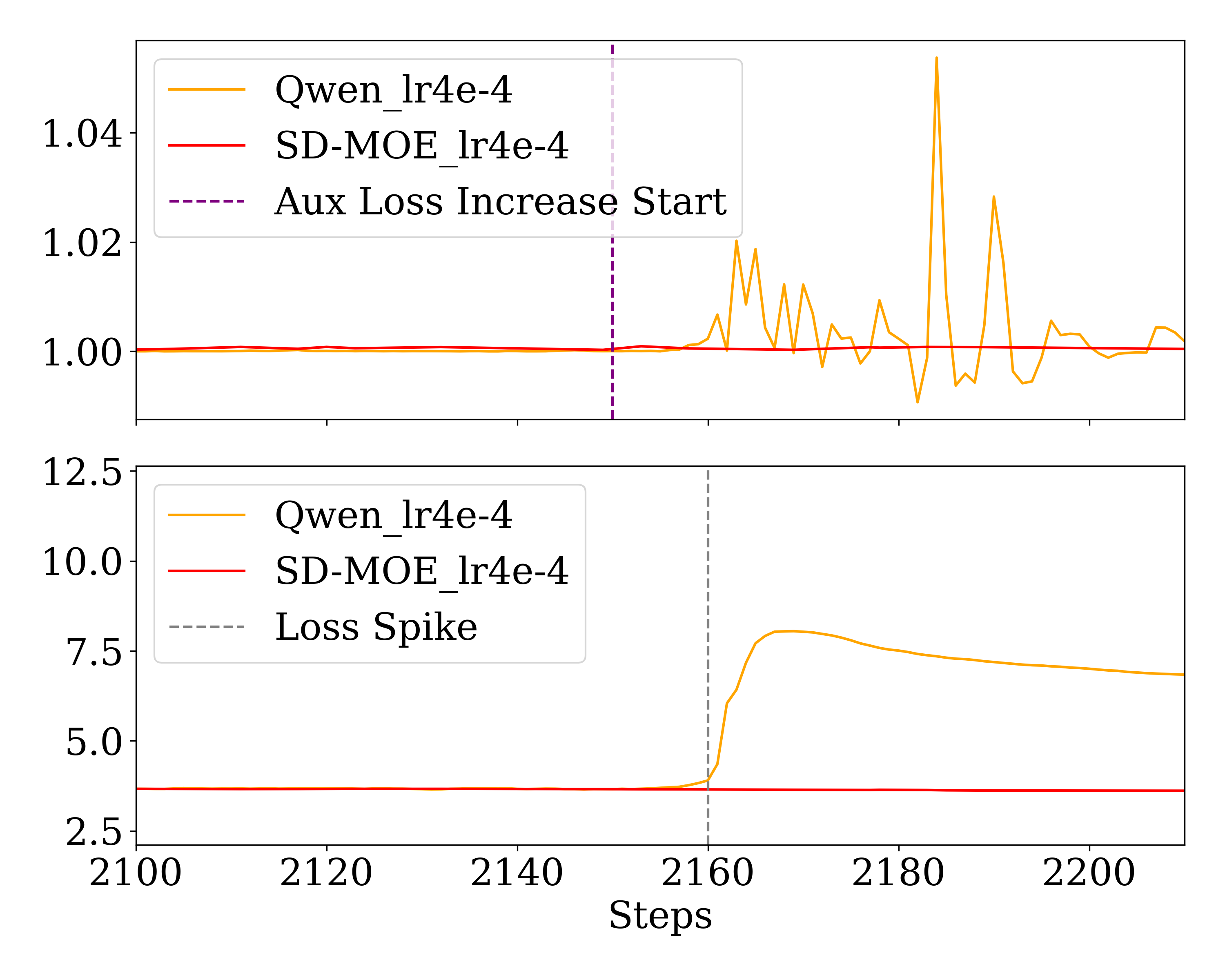}
        \label{fig:app.loss_spike}
    }
    \subfigure[]{
        \includegraphics[width=.45\linewidth]{figures/lr_loss.png}
        \label{fig:full_loss_curve}
    }
    \caption{
        \textbf{(a)} Zoomed-in view around the loss spike: the top panel shows the auxiliary loss, and the bottom panel shows the training loss. A vertical line marks the onset of rising aux loss, which precedes the sharp spike in the main loss. 
        \textbf{(b)} Full training loss curves.
    }
\label{fig:app.loss_and_lr}
\end{figure*}

Specifically, we repeat the high–learning-rate experiments and track the load-balance loss at each training step until divergence occurs in the baseline models. As shown in Figure~\ref{fig:app.loss_and_lr}, divergence in the baseline MoE is consistently preceded by a sharp increase in load-balance loss, indicating severe expert imbalance and routing collapse. In contrast, SD-MoE maintains a low and stable load-balance loss throughout training, even under aggressive learning rates.

These results provide complementary evidence that SD-MoE improves routing stability under high learning rates, consistent with the more benign spectral conditioning induced by the proposed spectral decoupling.

\section{Common Subspace Rank Ablations on SD-MoE}
\label{appendix:rank_abl}

We present ablation studies on the choice of the common subspace rank  $ k $  in SD-MoE. Table~\ref{tab:rank_ablation} summarizes the performance across different values of  $ k $ . Notably, downstream task performance peaks at  $ k = 8 $ , suggesting that a rank-8 shared subspace may be sufficient to capture the dominant alignment structure across expert parameters under the selected model configuration.

\begin{table*}[t]
\centering
\begin{tabular}{lccccc}
    \toprule
    Metric & Rank=2 & Rank=4 & Rank=8 & Rank=16 & Rank=32 \\
    \midrule
    ARC-challenge & 34.04 & 32.59 & \textbf{36.09} & 33.36 & 35.15 \\
    ARC-easy & 66.04 & 65.99 & \textbf{68.39} & 67.34 & 67.51 \\
    HellaSwag & 59.63 & \textbf{59.74} & 59.48 & 59.07 & 58.91 \\
    LAMBADA\_OPENAI & 42.79 & 42.56 & \textbf{42.97} & 42.32 & 42.77 \\
    PIQA & \textbf{74.59} & 74.54 & 74.16 & 74.05 & 74.16 \\
    RACE & 50.45 & 51.01 & 50.61 & 49.94 & \textbf{51.11} \\
    SIQA & \textbf{41.86} & 41.30 & 41.71 & 41.04 & 41.45 \\
    Winogrande & 59.75 & 59.43 & 58.33 & \textbf{61.40} & 58.41 \\
    \midrule
    Avg & 53.64 & 53.39 & \textbf{53.97} & 53.57 & 53.68 \\
    \bottomrule
\end{tabular}
\caption{
    Accuracy (\%) on 8 downstream tasks for Qwen-2B-A0.8B w. SD with varying common subspace ranks. The results show that a rank of 8 yields the best overall performance across all evaluated metrics, indicating that this might be an optimal choice for the shared subspace in our model configuration.
}
\label{tab:rank_ablation}
\end{table*}

\end{document}